\pdfoutput=1
\documentclass{article}

\PassOptionsToPackage{numbers,compress}{natbib}

\usepackage[eandd, final]{neurips_2026}

\usepackage[utf8]{inputenc} %
\usepackage[T1]{fontenc}    %
\usepackage{hyperref}       %
\usepackage{url}            %
\usepackage{booktabs}       %
\usepackage{amsfonts}       %
\usepackage{nicefrac}       %
\usepackage{microtype}      %
\usepackage{xcolor}         %
\usepackage{graphicx}
\usepackage{subcaption}     %
\graphicspath{{figs/}}

\usepackage{amsmath}
\usepackage{amssymb}

\usepackage{tabularx}
\usepackage{array}
\newcolumntype{Y}{>{\raggedright\arraybackslash}X}

\title{Towards Reliable VLM Judges: State‑Conditional Invariance and Presentation‑Aware Diagnostics}

\author{%
  {\bfseries
  Qinan Zhang\thanks{Equal Contribution.}\quad
  Xinyu Wang\footnotemark[1]\quad
  Qihang Jin\footnotemark[1]\quad
  Tianyuan Wang}\\[0.15em]
  {\bfseries
  Hongbing Luo\quad
  Pengfei Li\quad
  Wenjun Lv\quad
  Yu Kang\quad
  Jian Di\thanks{Corresponding Author.}}\\[0.25em]
  {\normalfont
  University of Science and Technology of China}\\[0.10em]
  {\normalfont\small\texttt{%
  \{qinanzhang,xinyu\_wang,qihangjin,dijian\}@mail.ustc.edu.cn%
  }}
}

\begin{document}

\maketitle

\begin{abstract}
    Vision-language models (VLMs) are increasingly used as automated judges for
    multimodal and embodied tasks, yet aggregate accuracy or consistency alone does
    not reveal whether a judge changes its verdict for the right reason. We propose
    State-Conditional Invariance (SCI), a reliability principle requiring a
    judge to remain stable under semantics-preserving changes in presentation while
    responding to causal interventions that change the task outcome. We instantiate
    SCI in \textbf{GridWM-Judge}, a simulator-grounded diagnostic benchmark built
    from deterministic MiniGrid trajectories with aligned Full,
    evidence-removal, and counterfactual variants, plus presentation and
    representation probes. Across five diagnostic tasks, GridWM-Judge tests local
    perception, action-conditioned transitions, presentation stability,
    representation exchangeability, and Full--CF causal discrimination. 
    Our diagnostic evaluations reveal heterogeneous failure modes, including sensitivity to evidence presentation and encoding, low activation on successful trajectories, and weak paired counterfactual discrimination. Perception-controlled analyses further indicate that visual grounding and objective recovery are substantial, model-dependent bottlenecks. These results motivate jointly reporting correctness, presentation stability, and causal sensitivity instead of relying on a single score.
\end{abstract}

\section{Introduction}
\label{sec:introduction}
Vision-language models (VLMs) are increasingly used not only as answer generators but also as automated judges. In text-only settings, LLM-based judges have been used to compare conversational responses and evaluate instruction-following systems
\citep{zheng2023judging,liu2023geval,zhu2023judgelm}. In multimodal settings, this paradigm has expanded to rubric-based vision-language evaluation and in-the-wild preference assessment
\citep{lee2024prometheus,chen2024mllmjudge,lu2024wildvision}, as well as multimodal reward, critic, and agent evaluation \citep{xiong2025llavacritic,li2025vlrewardbench,zhuge2024agentjudge}.
Such judges are attractive because they provide scalable alternatives to rule-based metrics and expensive human annotations. However, once a VLM judge is used to decide whether an embodied trajectory succeeds or fails, reliability becomes more subtle than aggregate accuracy: a judge may be correct for the wrong reason, stable because it defaults to one verdict, or unstable under changes that should not affect the task outcome.

\begin{figure}[t]
    \centering
    \includegraphics[width=0.99\linewidth]{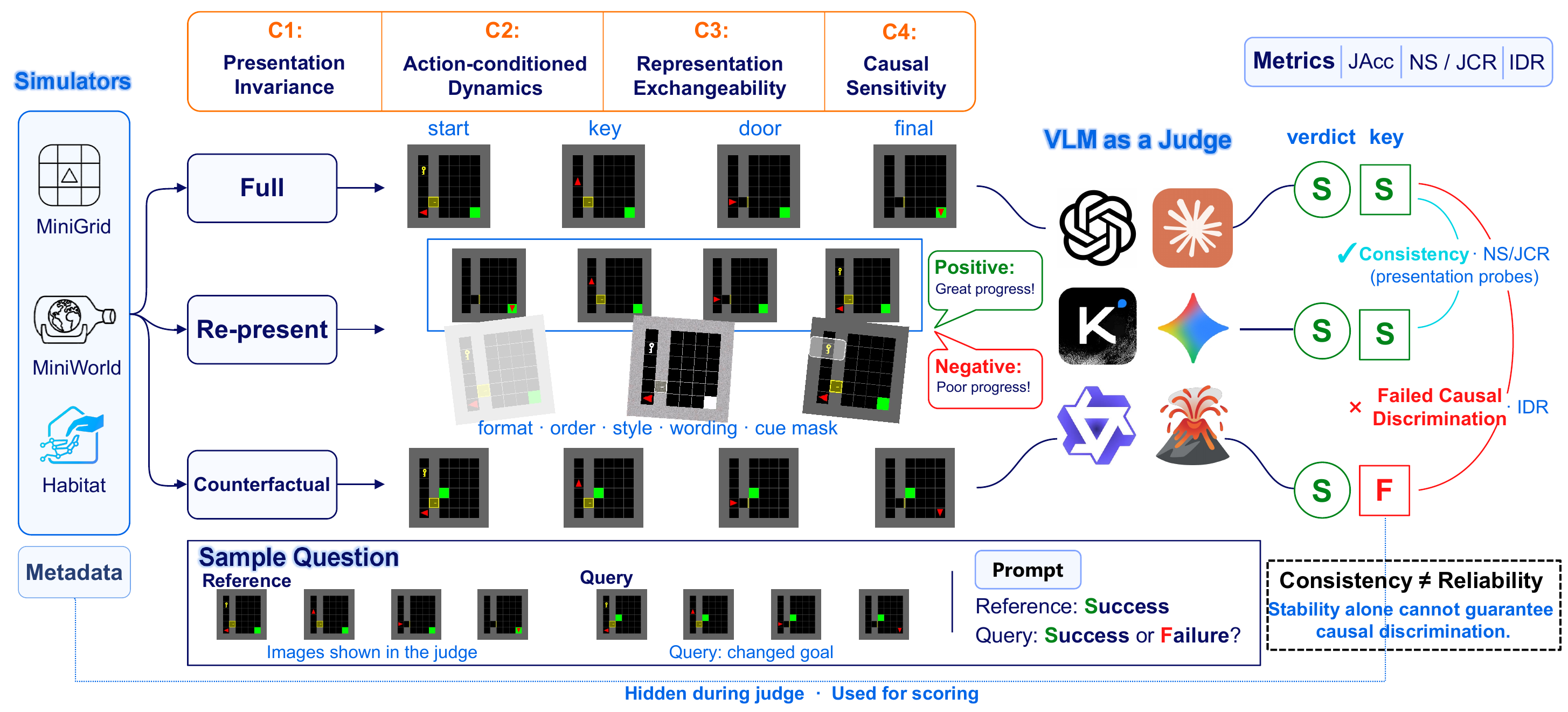}
    \caption{\textbf{Overview of GridWM-Judge.} The framework integrates simulator-based exam generation, VLM judging, and automatic scoring to assess four SCI diagnostic conditions (C1--C4). Ground-truth metadata are hidden from judges and used for scoring. The example illustrates that presentation consistency alone does not guarantee causal discrimination or reliable judging.
    }
    \label{fig:overview}
\end{figure}

This raises a simple but fundamental question: \emph{when should a vision-language judge change its mind?} Our answer is that a reliable judge should change its verdict when the underlying world changes, but not when only the presentation of the same evidence changes. For example, a successful trajectory should receive the same verdict under different layouts, explicitly indexed frame orderings, rendering styles, or prompt framing; conversely, a minimal intervention that changes the trajectory outcome from success to failure should change the verdict. We formalize this requirement as \emph{State-Conditional Invariance} (SCI): given the same simulator-grounded trajectory semantics $s$, a judge's verdict $\hat{y}$ should be conditionally independent of nuisance presentation variables
$\phi$,
\begin{equation}
    \hat{y} \perp \phi \mid s.
    \label{eq:intro_sci}
\end{equation}
SCI is therefore a selective invariance principle:
a reliable judge should be insensitive to how evidence is
presented when trajectory semantics are preserved, but
should change its verdict when the task outcome changes.
We use \emph{presentation} for temporal display order,
visual appearance, and linguistic framing within a fixed
input format; \emph{representation} refers to evidence
packaging, such as Composite versus Split images. Both are aspects of the \emph{evidence interface}:
C1 probes presentation invariance, whereas C3 tests
representation exchangeability.

Existing multimodal judge benchmarks do not fully isolate this distinction. Many evaluate agreement with human preferences, rubric-following ability, reward-model quality, or arena-style rankings
\citep{chen2024mllmjudge,lu2024wildvision,xiong2025llavacritic,li2025vlrewardbench}. These evaluations are valuable, but they often conflate two questions: whether a judge is robust to nuisance presentation changes, and whether it remains responsive to causal changes in the underlying task outcome. This distinction is especially important for embodied evaluation, where a counterfactual trajectory may look visually similar to a successful one while differing in a causally decisive event. In other words, existing benchmarks either evaluate judge agreement, preference alignment, or embodied reasoning ability, but rarely provide paired controls that jointly separate presentation invariance from causal sensitivity.

We introduce \textbf{GridWM-Judge}, a grid-world and world-model-inspired diagnostic benchmark for evaluating VLM judges under SCI.
Built on deterministic grid-world environments, GridWM-Judge uses the simulator as a Single Source of Truth (SSOT) for states, actions, events, and success predicates. For each trajectory group, it constructs aligned \textsc{Full}, evidence-removal, and counterfactual variants, and evaluates them through presentation probes and evidence representations.
This design separates presentation variants, where the world is fixed but the evidence interface changes, from causal counterfactuals, where an intervention changes the task outcome.

Figure~\ref{fig:overview} summarizes the overall diagnostic pipeline. Our main contributions can be summarized as:

\begin{itemize}
    \item We formulate VLM judge reliability as \emph{State-Conditional Invariance}, separating nuisance presentation changes from causal outcome changes.
    \item We instantiate SCI in GridWM-Judge, an executable benchmark with aligned trajectory variants, counterfactual interventions, and presentation probes.
    \item We characterize heterogeneous failure patterns, including presentation sensitivity, representation fragility, response bias, and weak causal discrimination, while using perception and action controls to qualify their interpretation.
\end{itemize}

\section{Related Work}
\label{related}
\paragraph{Model-based judges and multimodal evaluation.}
LLM-as-a-judge is now widely used for scalable evaluation of open-ended model
outputs, from comparative or rubric-based text evaluation
\citep{zheng2023judging,liu2023geval,zhu2023judgelm,li2023generative,kim2023prometheus,alpaca_eval}
to multimodal judging, preference assessment, reward modeling, and critic or
agent evaluation
\citep{lee2024prometheus,chen2024mllmjudge,lu2024wildvision,xiong2025llavacritic,li2025vlrewardbench,mmrewardbench2025,criticv2025,zhuge2024agentjudge}.
These works establish the utility of model-based judges, but primarily study
human preference alignment, rubric following, or reward quality. GridWM-Judge
instead asks whether a VLM judge changes its verdict for the right reason under
simulator-grounded presentation and causal controls.

\paragraph{Judge reliability, bias, and presentation effects.}
A growing literature studies the reliability of model judges themselves,
including fairness concerns, selection bias, verbosity bias, rubric sensitivity,
and inconsistency across prompts or candidates
\citep{wang2024large,zheng2023large,saito2023verbosity,huang2025judgebench}.
In multimodal judging, VLM evaluators may be unreliable under benign changes in
video evidence, judge composition, or input presentation
\citep{li2025judgingcharts,gao2025reliablevideojudge}. Separately, VLMs are
known to suffer from visual hallucination, image-context reasoning failures, and
object hallucination \citep{liu2023hallusionbench,li2023evaluating,wang2023evaluation}.
Long-context and image-sequence studies further show that models can depend on
where and how evidence is presented \citep{liu2024lost,wang2024mementos}. 
Recent work further identifies informativeness bias in VLM judges, where detailed responses can be preferred despite conflicting with visual evidence \citep{zou2026informativeness}.
Relatedly, Li et al.~\citep{li2025devil} investigate unimodal
spurious correlations in multimodal reward models.
Together, these findings motivate examining how judges respond
to changes in evidence presentation and underlying task semantics. 
GridWM-Judge instead uses simulator-grounded trajectories
and controlled probes to diagnose presentation and
representation sensitivity, response bias,
and failures of causal discrimination.
  
\paragraph{Embodied evaluation, world models, and counterfactual diagnostics.}
Controlled environments such as BabyAI and MiniGrid support auditable grounded
language and navigation experiments
\citep{chevalierboisvert2019babyai,chevalierboisvert2023minigrid}, while video
and embodied benchmarks such as CLEVRER and COVER evaluate causal or
counterfactual reasoning over dynamic scenes \citep{yi2020clevrer,zhou2025cover}.
World-model research studies action-conditioned prediction, planning, and
generative environment modeling
\citep{hafner2019planet,hafner2019dreamer,hansen2023,schrittwieser2020muzero,bruce2024genie,parkerholder2024genie2,yin2025trajectory,duan2025worldscore,warrier2025benchmarking}.
Recent grid-world and VLM diagnostics ask whether models solve, plan, act, or
form internal world-model-like representations
\citep{spatialworldmodels2026,spies2024transformers,goaldirectedagents2026,Gao2025};
GridWM-Judge instead evaluates whether a VLM can reliably judge what happened in
a trajectory under aligned presentation and causal counterfactuals.

\section{State-Conditional Invariance}
\label{sec:sci}

We formalize VLM judge reliability as \emph{State-Conditional Invariance} (SCI).
Our goal is not to infer whether a VLM has an internal world model, but to test
whether its success/failure verdict behaves correctly under controlled changes
to the world and to the presentation of evidence. A reliable judge should remain
invariant under semantics-preserving presentation changes, while remaining
sensitive to causal interventions that change the task outcome. This
selective-invariance view is related to invariant prediction, causal invariance,
counterfactual reasoning, and shortcut learning
\citep{arjovsky2019invariant,peters2016causal,kusner2017counterfactual,geirhos2020shortcut},
but we apply it to judge verdicts rather than to learned predictors, fairness
criteria, or agent policies.

\subsection{Setup and SCI Principle}
\label{sec:sci_setup}

Let $g \in \mathcal{G}$ denote a simulator-based trajectory group. Each
group contains an aligned state trajectory, action sequence, rendered evidence,
mission text, outcome label, and audit metadata. We write
\begin{equation}
s_g = (s_{g,0:T}, a_{g,0:T-1}, m_g), \qquad
\ell_g = h(s_g), \qquad
\hat{y}^{v,\phi}_g \in \{0,1\},
\end{equation}
where $s_g$ denotes simulator-grounded trajectory semantics, $h$ is the simulator success predicate,
$\ell_g \in \{0,1\}$ is the resulting success/failure label, $v$ is the trajectory variant, and $\phi$ is the presentation variable.
In our benchmark,
$v \in \{\mathrm{full}, \mathrm{nocue}, \mathrm{cf}\}$ denotes a successful full
trajectory, an evidence-removal diagnostic variant, and a counterfactual failure variant, respectively. 
For variant-specific rollouts, we use $s_g^v$ and
$\ell_g^v=h(s_g^v)$; Full and CF share a group identifier
and prescribed action sequence but may follow different
state trajectories. 
The variable $\phi$ describes temporal order, visual
appearance, and prompt framing within a fixed evidence
representation. We use $r$ for the Composite/Split
representation when comparing formats, suppressing this
index elsewhere for clarity.

SCI expresses nuisance invariance as conditional
independence of the verdict and presentation, given the
underlying trajectory semantics:
\begin{equation}
    \label{eq:sci}
    \hat{Y} \perp \Phi \mid S.
\end{equation}
We assess this requirement by comparing verdicts across
semantics-preserving presentations. Reliable judging
additionally requires correct decisions on successful
Full trajectories and their matched CF failures:
\begin{equation}
    \label{eq:sci_pair_cf}
    \hat{y}_{g}^{\mathrm{full},\phi}
    =
    \hat{y}_{g}^{\mathrm{full},\phi_0}
    \quad
    \text{for } \phi \in \Phi_{\mathrm{eq}}(s_g),
    \qquad
    \hat{y}_{g}^{\mathrm{full},\phi_0}=1,\ 
    \hat{y}_{g}^{\mathrm{cf},\phi_0}=0.
\end{equation}
Thus, SCI is not ordinary robustness: it asks whether invariance is appropriate.
A reliable judge should ignore nuisance presentation changes, but respond to
state changes that alter the task outcome.

\subsection{Diagnostic Conditions and Metrics}
\label{sec:sci_conditions}

GridWM-Judge instantiates SCI through four diagnostic conditions. Table~\ref{tab:sci_map}
summarizes the reliability question, benchmark instantiation, and main metrics.
The tasks are not independent leaderboards; they diagnose complementary failure
sources in the same judge.

\begin{table}[t]
    \centering
    \footnotesize
    \setlength{\tabcolsep}{3pt}
    \renewcommand{\arraystretch}{1.12}
    \caption{\textbf{SCI diagnostic conditions and their benchmark instantiations.}
    Tasks C and D use distinct evaluation protocols.
    Full metric definitions are provided in
    Appendix~\ref{app:definitions}.}
    \label{tab:sci_map}

    \begin{tabularx}{\linewidth}{@{}
        >{\raggedright\arraybackslash}p{0.20\linewidth}
        >{\raggedright\arraybackslash}p{0.265\linewidth}
        >{\raggedright\arraybackslash}X
        >{\raggedright\arraybackslash}p{0.18\linewidth}
    @{}}
        \toprule
        \textbf{Condition} &
        \textbf{Question} &
        \textbf{Instantiation} &
        \textbf{Metrics} \\
        \midrule

        \textbf{C1: Presentation Invariance} &
        Stable under semantics-preserving presentation changes? &
        Task C: temporal, visual, \newline and framing probes &
        NS, JCR, LES \\
        \addlinespace[0.3em]

        \textbf{C2: Action-conditioned Dynamics} &
        Do transition predictions \newline depend on actions? &
        Task A: next-state prediction; \newline action shuffle/mask &
        $\mathrm{Acc}_{A1}$,
        $\Delta_{\mathrm{shuffle}}$,
        $\Delta_{\mathrm{mask}}$ \\
        \addlinespace[0.3em]

        \textbf{C3: Representation Exchangeability} &
        Consistent across Composite/Split interfaces? &
        Task C: Composite/Split comparison &
        $\mathrm{VCC}_{\mathrm{rep}}$ \\
        \addlinespace[0.3em]

        \textbf{C4: Causal Sensitivity} &
        Does the judge distinguish matched Full/CF outcomes? &
        Task C: Full/CF pairs; \newline
        Task D: reference-aided Full/CF &
        $p_{\mathrm{active}}$,
        $\mathrm{IDR}_{\mathrm{raw}}$,
        $\mathrm{IDR}_{\mathrm{cond}}$ \\
        \bottomrule
    \end{tabularx}
\end{table}

We report four classes of metrics. Correctness anchors, such as
$\mathrm{JAcc}_{\mathrm{neu}}$ and $\mathrm{Acc}_{\mathrm{cf}}$, measure whether
the judge accepts successful trajectories and rejects counterfactual failures.
Presentation and representation metrics, such as NS, JCR, LES, and
$\mathrm{VCC}_{\mathrm{rep}}$, measure whether verdicts remain stable under
semantics-preserving changes. Causal-sensitivity metrics, especially
$p_{\mathrm{active}}$, $\mathrm{IDR}_{\mathrm{raw}}$, and
$\mathrm{IDR}_{\mathrm{cond}}$, measure whether the judge discriminates
successful full trajectories from matched counterfactual failures. Task-specific
controls, including Task A/B/E metrics, help distinguish perception or
action-conditioning failures from higher-level causal judgment failures.

We interpret consistency metrics only together with correctness and causal
sensitivity. A model can be stable simply because it always predicts the same
label. We therefore summarize failures as presentation leakage, representation
fragility, action-ignorant heuristics, causal misgrounding, and stability traps;
full definitions and degenerate-judge criteria are provided in
Appendix~\ref{app:definitions}.
Perception and output-format controls help interpret
these diagnostic flags.

\section{GridWM-Judge: Benchmark Construction}
\label{sec:benchmark}

\begin{figure}[t]
    \centering
    \includegraphics[width=0.99\linewidth]{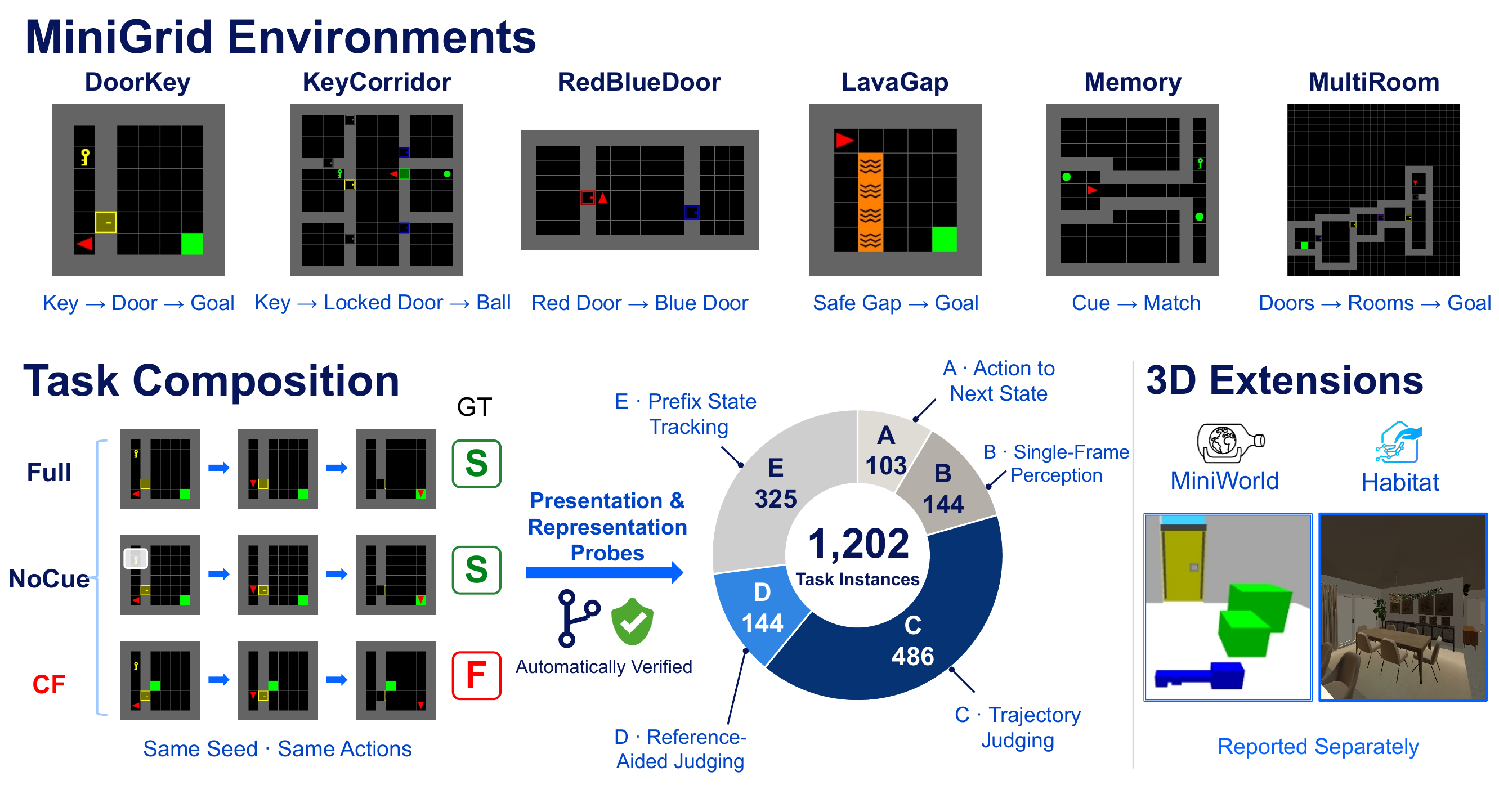}
    \caption{
    \textbf{Benchmark anatomy and dataset composition.} GridWM-Judge comprises six MiniGrid environment families and five diagnostic tasks. Simulator-generated Full, NoCue, and CF variants support controlled judging and presentation probes, while task-specific perception and transition tests provide complementary controls. The original benchmark contains 1,202 model-facing task instances; supplementary 3D extensions are evaluated separately.
    }
    \label{fig:benchmark_anatomy}
\end{figure}

GridWM-Judge instantiates the SCI framework in executable grid-world tasks. The
benchmark is designed around three requirements: (i) the simulator provides an
auditable source of truth for states, actions, events, and outcomes; 
(ii) trajectory-level evaluations support controlled
presentation changes and causal interventions;
and
(iii) the tasks diagnose complementary components of judge reliability rather
than forming a single leaderboard. 
Figure~\ref{fig:benchmark_anatomy} summarizes the six
MiniGrid environments, aligned trajectory variants,
task-specific dataset composition, and separately
reported 3D extensions.

\subsection{Environment Suite}
\label{sec:minigrid_suite}

Our original Tier-0 suite comprises six deterministic MiniGrid environment families \citep{chevalierboisvert2023minigrid}.
They cover object interaction, ordered door opening, hazard navigation, memory, and multi-room navigation, while remaining compact enough to provide symbolic states, rendered observations, discrete actions, and unambiguous success predicates. MiniGrid is not intended to approximate open-ended embodied benchmarks \citep{fan2022minedojo}; instead, it lets us audit whether a VLM judge changes its verdict when the world changes or merely when the same evidence is presented differently. 
Detailed environment descriptions and specific dataset
generation procedures are provided in
Appendix~\ref{app:env_data}.

\subsection{Trajectory Groups and Variants}
\label{sec:sample_generation}
For trajectory-level judging tasks, the source unit is a trajectory group
$g\in\mathcal{G}$, which contains aligned simulator states, actions, rendered
frames, mission text, outcome labels, and audit metadata. For each sampled
environment seed, we first construct a successful \textsc{Full} trajectory using
a deterministic planner. We then derive matched variants when supported by the
task: \textsc{NoCue}, which removes selected early evidence while preserving the
success outcome, and 
\textsc{CF}, a counterfactual failure generated by a
localized simulator intervention followed by rerollout
under the same action sequence. Although the actions
are fixed, simulator states and renderings may diverge
after the intervention; Full and CF outcomes are
independently verified by the simulator.
Thus, \textsc{NoCue} probes evidence dependence without
changing the success outcome, whereas \textsc{CF}
tests responses to outcome-changing interventions.
NoCue is not treated as a semantics-preserving
presentation transform.

Local perception and transition-control tasks are generated from task-specific
source units, such as environment-level states or local transitions, and are
reported as model-facing task instances. This distinction lets GridWM-Judge
combine trajectory-level Full/NoCue/CF diagnostics with local Task A/B/E controls
without treating all items as the same type of group. All generated items pass
task-appropriate validation gates before entering the benchmark, including
action-identity checks across aligned variants, \textsc{Full}-success and
\textsc{CF}-failure outcome gates, masking-window constraints, interaction
preservation checks, physics consistency checks, and symbolic state-encoding
checks when applicable. Full generation details and validator definitions are
given in Appendix~\ref{app:env_data}.

\subsection{Presentations and Diagnostic Tasks}
\label{sec:task_suite}

GridWM-Judge distinguishes two aspects of the evidence
interface. Temporal order, visual style, and linguistic
framing define presentation probes (C1), while Composite
and Split formats provide alternative representations
for testing exchangeability (C3).

The released benchmark contains five diagnostic tasks. Tasks B and E are
perception controls that test whether the model can parse local states, objects,
and structured scene information before higher-level judgment failures are
interpreted. Task C asks for success/failure judgments under presentation and
representation changes. Task A tests action-conditioned transition understanding
through next-observation prediction and action ablations. Task D is the primary
causal-sensitivity test, comparing successful \textsc{Full} trajectories with
matched \textsc{CF} failures. Full task interfaces, prompt templates, schemas,
and representative examples are provided in Appendix~\ref{app:tasks}.

\subsection{Dataset Release and 3D Extension}
\label{sec:dataset_release}

The benchmark is released as machine-readable exam files with image paths,
prompts, labels, and group metadata. Model-facing fields include only rendered
evidence, mission text, action text, and task instructions; audit-only metadata,
such as symbolic states, intervention records, labels, rewards, and validation
logs, are reserved for auditing and scoring. Full schemas, data cards, metadata,
and reproducibility instructions are provided in Appendix~\ref{app:implementation}.

\paragraph{Availability.}
The benchmark, including exam files, prompts,
validation records, and scoring resources, is available
through the \href{https://github.com/xinyuw922-pixel/Towards-Reliable-VLM-Judge}
{\textbf{code repository}} and
\href{https://huggingface.co/datasets/nahco318/Towards-Reliable-VLM-Judges}
{\textbf{dataset release}}.
Implementation details, model configurations, data schemas,
and reproducibility instructions are documented in
Appendix~\ref{app:implementation}.

\section{Diagnostic Experiments}
\label{sec:experiments}

We evaluate model-based judges against the four SCI diagnostic conditions, focusing on causal discrimination, presentation stability, and representation exchangeability.

\subsection{Experimental Setup}  
\label{sec:exp_setup}

\paragraph{Models, tasks, and representations.}
We evaluate a diverse set of proprietary and open VLM judges, including
frontier general-purpose models, lightweight flash-style models, and
representative open VLM families when complete runs are available
\citep{bai2023qwen,liu2023visual,wang2024cogvlm}. 
Each model acts as an external judge rather than an agent.
We evaluate the five MiniGrid tasks described in
Section~\ref{sec:benchmark}, with Tasks B/E providing
perception controls, Task A testing action-conditioned
transitions, Task C examining presentation and representation
effects, and Task D evaluating causal discrimination.
The key
representation comparison is between composite storyboards and split-image
inputs, which encode the same trajectory semantics but expose evidence through
different interfaces.

\paragraph{Inference, metrics, and statistics.}
All models receive task-specific instructions with the same model-facing fields:
rendered evidence, mission text when available, action text, and the required
answer format, following standard instruction-following and judge-prompting
protocols \citep{ouyang2022training,zheng2023judging,liu2023geval}. 
We use deterministic decoding when supported.
Responses are parsed using task-specific scoring rules;
invalid outputs and execution failures are tracked separately. The main Task-C summaries use the released eligible-pair protocol, whereas dedicated Task-D and expanded evaluations report conservative all-group estimates.
The main Task D metrics are $\mathrm{JAcc}_{\mathrm{neu}}$, $\mathrm{Acc}_{\mathrm{cf}}$,
$p_{\mathrm{active}}$, $\mathrm{IDR}_{\mathrm{raw}}$, and
$\mathrm{IDR}_{\mathrm{cond}}$; Task C additionally reports presentation and
representation metrics such as NS, JCR, LES, and $\mathrm{VCC}_{\mathrm{rep}}$.
We interpret consistency metrics only together with correctness and causal
sensitivity, because a constant-verdict judge can be stable without being
reliable. Full model identifiers, prompts, parsing rules and other experimental settings are provided in Appendix~\ref{app:implementation}.

\begin{table}[t]
\centering
\small
\setlength{\tabcolsep}{4.2pt}
\renewcommand{\arraystretch}{1.08}
\caption{\textbf{Task-C diagnostics under Composite (C)
and Split (S) evidence interfaces.} 
Deg. denotes the degenerate status under the corresponding representation; 
AF denotes all-fail degeneracy, SB denotes success bias,
and \textemdash \ denotes non-degenerate behavior.}
\label{tab:main_diagnostic}
\begin{tabular}{@{}lccccccc@{}}
\toprule
\textbf{Model}
& \textbf{$\mathrm{JAcc}_{\mathrm{neu}}$}
& \multicolumn{2}{c}{\textbf{$p_{\mathrm{active}}$}}
& \multicolumn{2}{c}{\textbf{$\mathrm{IDR}_{\mathrm{raw}}$}}
& \textbf{$\mathrm{VCC}_{\mathrm{rep}}$}
& \textbf{Deg. (C/S)} \\
\cmidrule(lr){3-4}
\cmidrule(lr){5-6}
& \textbf{C/S} & \textbf{C} & \textbf{S} & \textbf{C} & \textbf{S} & & \\
\midrule
\textit{Claude Sonnet 4.6}$^{\dagger}$
& 0.06/0.11 & 0.06 & 0.12 & 0.06 & 0.12 & 0.924 & AF/\textemdash \\
\textit{GPT-4o}
& 0.28/0.78 & 0.28 & \textbf{0.78} & 0.22 & \textbf{0.72} & 0.706 & \textemdash/\textemdash \\
\textit{GPT-5.4}
& 0.06/0.06 & 0.06 & 0.06 & 0.06 & 0.06 & \textbf{1.000} & AF/AF \\
\textit{Gemini 3 Flash}
& 0.72/0.67 & 0.72 & 0.67 & 0.67 & 0.67 & 0.707 & \textemdash/\textemdash \\
\textit{Gemini 2.5 Flash Lite}
& 0.50/0.67 & 0.50 & 0.67 & 0.28 & 0.56 & 0.796 & \textemdash/\textemdash \\
\textit{Gemini 2.0 Flash}$^{\dagger}$
& 0.00/0.61 & 0.00 & 0.59 & 0.00 & 0.53 & 0.544 & AF/\textemdash \\
\textit{Kimi K2.5}
& 0.39/0.72 & 0.50 & 0.72 & 0.36 & \textbf{0.72} & 0.661 & \textemdash/\textemdash \\
\textit{Qwen3-32B}$^{\ddagger}$
& 1.00/0.67 & 1.00 & 0.67 & 0.00 & 0.00 & 0.667 & SB/\textemdash \\
\textit{Qwen3-235B}
& 0.67/0.67 & 0.67 & 0.67 & 0.00 & 0.00 & 0.743 & \textemdash/\textemdash \\
\textit{Qwen3-8B}
& 0.67/0.00 & 0.67 & 0.00 & 0.00 & 0.00 & 0.407 & \textemdash/AF \\
\midrule
\multicolumn{8}{@{}l}{\footnotesize $^{\dagger}$ Composite is all-fail degenerate, while Split is non-degenerate.} \\
\multicolumn{8}{@{}l}{\footnotesize $^{\ddagger}$ Success-biased degenerate under Composite.} \\
\bottomrule
\end{tabular}
\end{table}

\begin{figure}[t]
    \centering
    \includegraphics[width=0.9\linewidth]{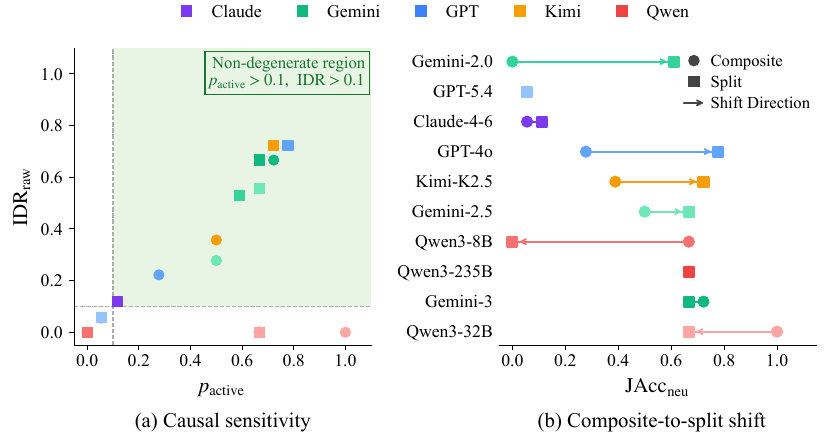}
    \caption{\textbf{Task-C diagnostic results.} 
    (a) plots Full activation against raw paired Full--CF discrimination under Composite and Split interfaces.
    (b) shows the corresponding representation shifts in Full-trajectory accuracy and discrimination. Together, the panels illustrate that high acceptance, paired discrimination, and representation consistency need not coincide.}
    \label{fig:main_diagnostics}
\end{figure}

\subsection{Analysis}
\label{sec:exp_analysis}

Table~\ref{tab:main_diagnostic} and Figure~\ref{fig:main_diagnostics} summarize
Task-C
diagnostics of activation, paired discrimination, and
representation sensitivity. We use
$(p_{\mathrm{active}}, \mathrm{IDR}_{\mathrm{raw}})$ as the primary causal-sensitivity coordinates because they jointly indicate whether a judge accepts genuine successful trajectories and rejects matched counterfactual failures.
Dedicated reference-aided
Task-D results with clustered
confidence intervals and other expanded evaluations are reported in
Appendix~\ref{app:results}.

\begin{figure}[t]
    \centering
    \includegraphics[width=0.99\linewidth]{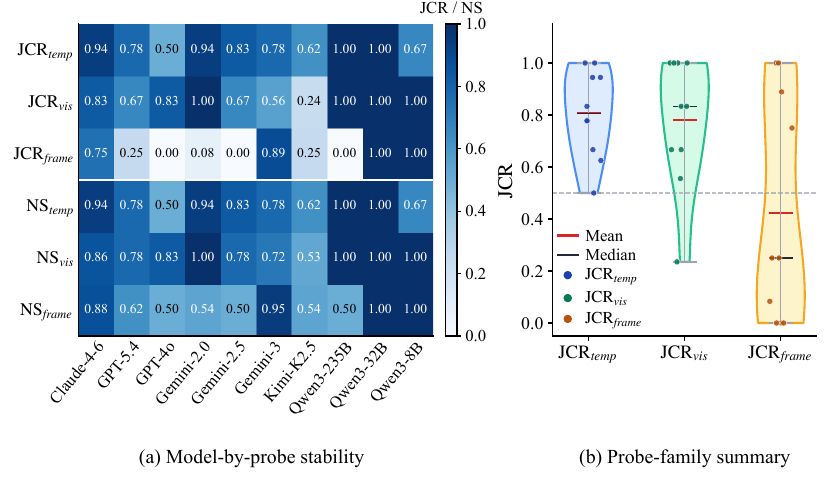}
    \caption{\textbf{Presentation stability across probe families.} 
    (a) compares NS and JCR across temporal, visual, and framing probes for each model.
    (b) summarizes the corresponding JCR distributions. 
    }
    \label{fig:presentation_robustness}
\end{figure}

\begin{table}[t]
\centering
\caption{\textbf{Composite--Split representation ablation.}
Reported values are model-wise means, with
$\Delta=\mathrm{Split}-\mathrm{Composite}$.
W/T/L denotes improvements, ties, and degradations
across models; the last column reports Wilcoxon
signed-rank $p$-values. All comparisons are not significant at $p<0.05$.}
\label{tab:rep_ablation}
\begin{tabular}{lcccccc}
\toprule
\textbf{Metric} & \textbf{Composite Mean} & \textbf{Split Mean} & \textbf{Mean $\Delta$} & \textbf{Median $\Delta$} & \textbf{W/T/L} & \textbf{$p$-value} \\
\midrule
$\mathrm{JAcc}_{\mathrm{neu}}$ $\uparrow$ & 0.433 & 0.494 & +0.061 & +0.028 & 5/2/3 & 0.617 \\
${p}_{\mathrm{active}}$ $\uparrow$ & 0.444 & 0.493 & +0.048 & +0.031 & 5/2/3 & 0.641 \\
$\mathrm{IDR}_{\mathrm{raw}}$ $\uparrow$ & 0.163 & 0.337 & +0.173 & +0.031 & 5/5/0 & 0.063 \\
$\mathrm{IDR}_{\mathrm{cond}}$ $\uparrow$ & 0.624 & 0.720 & +0.096 & +0.038 & 4/4/0 & 0.125 \\
\midrule
\multicolumn{7}{p{0.9\textwidth}}{\small 
*For $\mathrm{IDR}_{\mathrm{cond}}$, models with undefined or low-activation conditional IDR are excluded from W/T/L counting according to the reporting rule in Appendix~\ref{app:definitions}.}
\end{tabular}
\end{table}

\paragraph{Finding 1: Accuracy does not imply causal sensitivity.}
Task-C paired diagnostics show that Full acceptance and counterfactual rejection capture different aspects of judge behavior. A judge that predicts failure almost
everywhere may achieve high CF rejection but low activation, while a success-biased judge may accept Full trajectories without rejecting their matched CFs.
Dedicated reference-aided Task D confirms this distinction: GPT-5.4 achieves Full acceptance of 0.750, CF rejection of 0.500, and raw IDR of 0.333, whereas Claude Sonnet 4.6 obtains 0.250, 0.917, and 0.250, respectively (Table~\ref{tab:appD_taskd_strict} in Appendix~\ref{app:results}).
On 432 expanded Task-D groups, their raw IDRs are 0.322 and 0.088, respectively, with group-bootstrap 95\% CIs of [0.278, 0.366] and [0.062, 0.116] (Table~\ref{tab:appD_scale2} in Appendix~\ref{app:results}).
Thus, we assess causal sensitivity jointly through
$(p_{\mathrm{active}},\mathrm{IDR}_{\mathrm{raw}})$,
with conditional IDR interpreted only when activation is sufficient.

\paragraph{Finding 2: Representation changes judge behavior.}
Task-C presentation probes reveal varying stability
across temporal, visual, and framing changes
(Figure~\ref{fig:presentation_robustness}).
Composite and Split interfaces also produce different
verdicts for matched trajectories, demonstrating
sensitivity to evidence presentation and packaging.
Although Split yields a higher mean raw IDR in
Table~\ref{tab:rep_ablation}, the across-model
difference is not statistically significant
($p=0.063$). Moreover, the compared interfaces
also differed in prompt wording and run conditions,
so the effect cannot be attributed solely to image
layout. We use $\mathrm{VCC}_{\mathrm{rep}}$ to
quantify representation consistency and report
matched transition analyses in
Appendix~\ref{app:results}.
Matched temporal, noise, and style probes show both
correctness degradation and rescue, although none of
the 180 probe-level comparisons survives the specified
Holm correction.

\begin{figure}[t]
    \centering
    \includegraphics[width=0.99\linewidth]{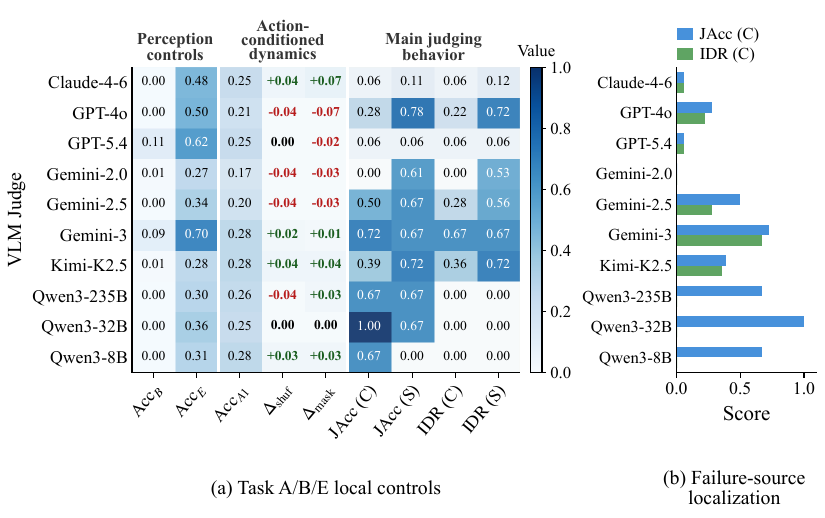}
    \caption{\textbf{Local controls and judgment outcomes.}
    (a) compares Task-B/E perception scores, Task-A transition accuracy and action-ablation effects, and Task-C judging metrics across models.
    (b) contrasts Task-C Full acceptance and raw intervention discrimination. These complementary diagnostics help interpret, but do not uniquely identify, failure sources.}
    \label{fig:local_diagnostics}
\end{figure}

\paragraph{Finding 3: Stable judgments can mask failure;
local controls reveal additional bottlenecks.}
High NS or JCR does not imply reliability: a model can remain stable while failing to discriminate
matched Full/CF outcomes because it defaults to a constant verdict or shortcut rule~\citep{geirhos2020shortcut}. Figure~\ref{fig:local_diagnostics}
and Table~\ref{tab:local_controls} provide complementary
Task-B/E perception and Task-A transition controls.
For example, GPT-5.4 achieves Task-B Acc@0.90 of 0.111
despite an all-row mean component score of 0.616,
showing that high-threshold accuracy can obscure partial
scene recovery. Task-A ablation effects are generally
small or mixed, providing limited evidence about
action-conditioned reasoning on their own.
In a matched 98-group Oracle-State experiment,
GPT-5.4's raw IDR increases from 0.469 with Pixel
evidence to 0.765 with State evidence, highlighting
visual grounding and objective recovery as potential
bottlenecks. Together, these controls help distinguish
plausible failure sources without uniquely identifying
their causes (Appendix~\ref{app:results}).

\begin{table}[t]
\centering
\caption{\textbf{Local diagnostic controls.}
Task B reports Acc@0.90, the fraction of items meeting
a 0.90 threshold on the normalized weighted scene-component
score; Task E measures object/state recognition, and
Task A measures next-observation prediction.
Action-ablation differences are computed as baseline
accuracy minus ablated accuracy.}
\label{tab:local_controls}
\begin{tabular}{lccccc}
\toprule
\textbf{Model} & \textbf{B\_Acc} & \textbf{E\_Acc} & \textbf{Acc\_A1} & \textbf{$\Delta_{\text{shuf}}$} & \textbf{$\Delta_{\text{mask}}$} \\ 
\midrule
Claude Sonnet 4.6 & 0.000 & 0.477 & 0.252 & +0.039 & +0.068 \\
GPT-4o & 0.000 & 0.495 & 0.214 & -0.039 & -0.068 \\
GPT-5.4 & 0.111 & 0.615 & 0.252 & +0.000 & -0.019 \\
Gemini 2.0 Flash & 0.007 & 0.274 & 0.175 & -0.038 & -0.025 \\
Gemini 2.5 Flash Lite & 0.000 & 0.345 & 0.204 & -0.039 & -0.029 \\
Gemini 3 Flash & 0.090 & 0.702 & 0.282 & +0.019 & +0.010 \\
Kimi K2.5 & 0.007 & 0.280 & 0.282 & +0.039 & +0.039 \\
Qwen3-235B & 0.000 & 0.305 & 0.262 & -0.039 & +0.029 \\
Qwen3-32B & 0.000 & 0.357 & 0.252 & +0.000 & +0.000 \\
Qwen3-8B & 0.000 & 0.311 & 0.282 & +0.029 & +0.029 \\
\midrule
\end{tabular}
\end{table}

\paragraph{Summary.}
Across diagnostic tasks, correctness, stability, and
causal discrimination reveal complementary failure modes.
Expanded MiniGrid and 3D evaluations
further support interpreting judge behavior
through these dimensions rather than a single aggregate score.
(Appendices~\ref{app:results}
and~\ref{app:3d_extensions}).

\section{Conclusion}
\label{sec:conclusion}

We introduced \textbf{GridWM-Judge}, a simulator-grounded
benchmark for auditing model-based judges through
\emph{State-Conditional Invariance} (SCI).
By combining matched trajectory variants, controlled
evidence-interface probes, and action/perception controls,
the benchmark distinguishes appropriate verdict stability
from responsiveness to outcome-changing interventions.
Across evaluated judges, apparent consistency and
successful-trajectory acceptance do not guarantee
reliable paired causal discrimination; additional controls
reveal model-dependent grounding and response-format
bottlenecks. Expanded MiniGrid evaluations and
supplementary MiniWorld/Habitat experiments extend these
diagnostics to more 2D and 3D settings.
Our findings motivate evaluating judge reliability through
correctness, presentation stability, and causal sensitivity
jointly, rather than relying on a single score.

\begin{ack}
\noindent\textbf{Funding.}
This work was supported in part by 
the Anhui Provincial Innovation Breakthrough Plan
(Grant No. 202523j08050018),
the Anhui Innovation Platform Construction -- High-Level New R\&D
Institution Project (Grant No. 202524vb12010005),
the Centrally Guided Local Science and Technology Development
Special Fund of Anhui Province (Grant No. 202407a12020009),
and the National Key R\&D Program of China
(Grant No. 2022YFB3206904).

\noindent\textbf{Competing Interests.}
The authors declare no competing interests.
\end{ack}

{
\small
\bibliographystyle{unsrtnat}
\bibliography{references}
}

\appendix

\section*{Appendix Overview}
\label{app:overview}
This appendix provides additional details supporting
the main paper. Appendix~\ref{app:definitions} defines
SCI conditions, metrics, statistical procedures, and
failure modes. Appendix~\ref{app:env_data} describes
MiniGrid environments, dataset construction, validation,
and composition. Appendix~\ref{app:tasks} presents task
interfaces, prompts, and representative examples.
Appendix~\ref{app:results} reports supplementary
MiniGrid results, expanded evaluations, ablations,
and failure analyses. Appendix~\ref{app:3d_extensions}
presents MiniWorld and Habitat 3D evaluations.
Appendix~\ref{app:implementation} documents implementation,
data release, and reproducibility protocols.
Appendix~\ref{app:scope_impact} discusses limitations,
societal impact, and future directions.

\section{Formal Definitions and Metrics}
\label{app:definitions}
This appendix expands the State-Conditional Invariance (SCI) framework
summarized in Section~\ref{sec:sci}.  We define the simulator-grounded
trajectory semantics, presentation variables, trajectory variants, and
representation formats used by GridWM-Judge; formalize the four diagnostic
conditions C1--C4; and provide metric and statistical definitions for
the results reported in the main text. We also define the failure modes and degenerate-judge criteria used throughout the analysis, including presentation leakage, representation fragility, action-ignorant heuristics, causal misgrounding, and stability traps. 

\subsection{State and Presentation Variables}
\label{app:state_presentation_variables}

\paragraph{Trajectory groups.}
Let $g \in \mathcal{G}$ denote a trajectory group generated from the simulator.
Each group contains an aligned action sequence, state trajectory, rendered
evidence, mission or task text, outcome labels, and audit metadata. We write
\begin{equation}
s_g = (s_{g,0:T}, a_{g,0:T-1}, m_g)
\end{equation}
for the simulator-grounded trajectory semantics, where $s_{g,0:T}$ is the
symbolic simulator-state sequence, $a_{g,0:T-1}$ is the fixed action sequence,
and $m_g$ is the mission text when provided. The binary outcome label is computed
by the simulator success predicate,
\begin{equation}
\ell_g = h(s_g), \qquad \ell_g \in \{0,1\}.
\end{equation}
By convention, $\ell_g=1$ denotes success and $\ell_g=0$ denotes failure. The
rendered images shown to a model are generated from simulator states but are not
treated as privileged state in the model-facing prompt.

\paragraph{Trajectory variants.}
For variant-specific simulator rollouts, we write
$s_g^v$ and $\ell_g^v=h(s_g^v)$. For each sampled group, GridWM-Judge constructs aligned variants
$v\in\{\mathrm{full},\mathrm{nocue},\mathrm{cf}\}$ when supported by the
corresponding task. The \textsc{Full} variant is a successful rollout, so
$\ell_g^{\mathrm{full}}=1$. The \textsc{NoCue} variant removes selected early
visual evidence while preserving the success outcome,
$\ell_g^{\mathrm{nocue}}=1$; it is an evidence-removal diagnostic, not a
semantics-preserving presentation transform. The \textsc{CF} variant is a
counterfactual failure generated by a minimal simulator intervention followed by rerollout under the unchanged action sequence, with $\ell_g^{\mathrm{cf}}=0$. Full and NoCue share
the successful outcome but not necessarily the same
rendered evidence; CF can follow a different state
trajectory after the declared intervention while retaining
the prescribed action sequence.

\paragraph{Presentations and representations.}
A VLM judge receives a presentation of trajectory evidence and produces a
binary verdict
\begin{equation}
    \label{eq:app-verdict}
    \hat{y}_{g}^{v,\phi} \in \{0,1\},
\end{equation}
where $v$ denotes the trajectory variant.
The presentation variable $\phi$ describes temporal display
order, visual appearance, and linguistic framing within a
fixed evidence representation. 
We could denote the canonical
presentation by
\begin{equation}
    \label{eq:app-canonical-presentation}
    \phi_0=(\mathrm{temp/orig},\mathrm{vis/clean},\mathrm{frame/neu}).
\end{equation}
We use $r$ to denote the
Composite/Split evidence format when testing representation
exchangeability. Oracle-State and Pixel+State inputs are
treated separately as evidence conditions that alter the
information available to the judge, rather than equivalent
representations for C3.

\paragraph{State-Conditional Invariance.}
The core reliability principle is State-Conditional Invariance (SCI):
conditioned on the same underlying trajectory semantics $s$, the judge's
verdict should be independent of nuisance presentation variables $\phi$,
\begin{equation}
    \label{eq:app-sci}
    \hat{Y}\perp \Phi\mid S.
\end{equation}
As an operational sample-level check, for a semantics-preserving presentation transform
$\phi\in\Phi_{\mathrm{eq}}(s_g)$, a reliable judge should satisfy
\begin{equation}
    \label{eq:app-sci-pairwise}
    \hat{y}_{g}^{\mathrm{full},\phi}
    =
    \hat{y}_{g}^{\mathrm{full},\phi_0}.
\end{equation}
SCI is therefore not ordinary robustness. It does not require the verdict to be
unchanged under all input changes. Instead, it asks whether invariance is
appropriate: the verdict should remain stable under presentation changes that
preserve $s_g$, but should change under causal interventions that alter the
outcome.

\subsection{Four Diagnostic Conditions}
\label{sec:sci_conditions_App}
We decompose judge reliability into four diagnostic conditions, each instantiated
by one or more tasks in GridWM-Judge.

\paragraph{C1: Presentation Invariance.}
For the same underlying trajectory semantics $s_g$, the judge should be
invariant to nuisance presentation transforms:
\begin{equation}
\forall v \in \mathcal{V}_{\mathrm{fixed}},\quad
\forall \phi \in \Phi_{\mathrm{eq}}(s_g^v),\quad
\hat{y}^{v,\phi}_g =
\hat{y}^{v,\phi_0}_g .
\end{equation}

In GridWM-Judge, $\Phi_{\mathrm{eq}}$ includes temporal presentation transforms
(e.g., original versus reversed display order), visual transforms (e.g., clean,
noisy, or style-adjusted renderings), and prompt-framing transforms (e.g.,
neutral, positive, or negative preambles). These transforms are designed to
preserve the trajectory semantics and the success label. C1 is mainly tested by
Task C through the NS, JCR, Flip, and LES metrics defined below. In the main analysis, C1 is evaluated within a fixed trajectory condition; Full--CF
differences are never counted as presentation invariance because they change the
underlying outcome.

\paragraph{C2: Action-Conditioned Dynamics.}
Judge-relevant reasoning should depend on action-conditioned state transitions,
rather than on terminal-frame heuristics or superficial visual matching. This
condition is conceptually aligned with action-conditioned world-model learning
and model-based control \citep{hafner2019planet,hafner2019dreamer,hansen2023,
schrittwieser2020muzero}. Let $f$
denote the implicit transition model needed to predict future states. A minimal
requirement is
\begin{equation}
    \label{eq:app-c2}
    f^{(k)}(s_t,a_{t:t+k-1})\approx s_{t+k},\quad k\geq 1.
\end{equation}
Task A tests the one-step case by asking the model to select the correct next
observation $o_{t+1}$ from four candidates given the current observation $o_t$
and action $a_t$. 
Action-shuffling and action-masking ablations provide
behavioral evidence of whether predictions depend on
the supplied action.

\paragraph{C3: Representation Exchangeability.}
The same trajectory semantics may be presented through different but equivalent
evidence formats, such as a composite storyboard or split-image representation. This concern is related to evidence-ordering and
image-sequence reasoning effects observed in long-context and multimodal
settings \citep{liu2024lost,wang2024mementos}. A reliable judge should not be bound to a single
interface:
\begin{equation}
\hat{y}^{v,r,\phi}_g = \hat{y}^{v,r',\phi}_g,
\qquad \text{for the same } (g,v,\phi),
\end{equation}
whenever these representations encode the same trajectory semantics. In the
main experiments, C3 is tested primarily by comparing composite and split-image
presentations; Structured-state and Pixel+State inputs are instead used as
perception-controlled evidence conditions: they alter the
information available to the judge and are not treated as
purely exchangeable visual layouts.

\paragraph{C4: Causal Sensitivity.}
A judge should be insensitive to nuisance transforms but sensitive to causal
interventions. For each group, the counterfactual variant changes the simulator
state or initial condition so that the same action sequence leads to failure:
\begin{equation}
    \label{eq:app-c4-intervention}
    \begin{aligned}
    s_g^{\mathrm{full}} &\rightarrow s_g^{\mathrm{cf}},\\
    \ell_g^{\mathrm{full}} &= 1,\qquad \ell_g^{\mathrm{cf}}=0.
    \end{aligned}
\end{equation}
A reliable judge should therefore satisfy
\begin{equation}
    \label{eq:app-c4-verdict}
    \hat{y}_{g}^{\mathrm{full},\phi_0}=1
    \quad\text{and}\quad
    \hat{y}_{g}^{\mathrm{cf},\phi_0}=0.
\end{equation}
C4 is mainly tested by Task D, which directly compares full successful
trajectories with counterfactual failures. Task C also contains paired Full/CF diagnostic rows; however, its Composite/Split protocol is distinct from the reference-aided Task-D contract.

\paragraph{Task mapping and NoCue.}
The tasks are not independent leaderboards; they diagnose different components
of SCI. Task C tests presentation invariance and representation sensitivity
(C1/C3), Task D tests causal sensitivity (C4), Task A tests action-conditioned
transition understanding (C2), and Tasks B/E provide perception and state
recognition controls. The NoCue variant probes whether the judge depends on
early visual cues, but it is not counted as either a semantics-preserving
presentation transform for C1 or a causal intervention for C4.

\subsection{Metrics Formulation}
\label{app:metrics_formulation}

We define all metrics as empirical estimates over their applicable item sets.
Let $\mathbf{1}[\cdot]$ be the indicator function. Unless otherwise stated,
$G=|\mathcal{G}|$ denotes the number of trajectory groups in the relevant task,
variant set, and representation condition.

\paragraph{Correctness anchors.}
The main correctness anchor is the neutral full-trajectory judgment accuracy:
\begin{equation}
    \label{eq:app-jacc-neu}
    \mathrm{JAcc}_{\mathrm{neu}}
    =
    \frac{1}{G}
    \sum_{g\in\mathcal{G}}
    \mathbf{1}\!\left[
    \hat{y}_{g}^{\mathrm{full},\phi_0}=1
    \right],
\end{equation}
where all full trajectories are successful by construction. We also report
variant-wise accuracies:
\begin{equation}
    \label{eq:app-acc-full}
    \mathrm{Acc}_{\mathrm{full}}
    =
    \frac{1}{G}
    \sum_{g}
    \mathbf{1}\!\left[
    \hat{y}_{g}^{\mathrm{full},\phi_0}=1
    \right],
\end{equation}
\begin{equation}
    \label{eq:app-acc-nocue}
    \mathrm{Acc}_{\mathrm{nocue}}
    =
    \frac{1}{G}
    \sum_{g}
    \mathbf{1}\!\left[
    \hat{y}_{g}^{\mathrm{nocue},\phi_0}=1
    \right],
\end{equation}
and
\begin{equation}
    \label{eq:app-acc-cf}
    \mathrm{Acc}_{\mathrm{cf}}
    =
    \frac{1}{G}
    \sum_{g}
    \mathbf{1}\!\left[
    \hat{y}_{g}^{\mathrm{cf},\phi_0}=0
    \right].
\end{equation}
Here $\mathrm{Acc}_{\mathrm{nocue}}$ is diagnostic because the NoCue trajectory
preserves the success outcome while removing selected early evidence. We also
track the success prediction rate
\begin{equation}
    \label{eq:app-succ-c}
    \mathrm{SuccC}
    =
    \frac{1}{|\mathcal{I}|}
    \sum_{i\in\mathcal{I}}
    \mathbf{1}\!\left[\hat{y}_i=1\right],
\end{equation}
over the relevant binary judgment set $\mathcal{I}$, as a label-bias indicator.

\paragraph{Presentation invariance: NS and JCR.}
For a probe family $F\in\{\mathrm{temp},\mathrm{vis},\mathrm{frame}\}$, let
$\Phi_F(g)$ be the set of non-baseline presentations in that family for group
$g$, and let $\phi_{F,0}$ be the family-specific baseline. The Nuisance
Stability score is
\begin{equation}
    \label{eq:app-ns}
    \mathrm{NS}_{F}
    =
    \frac{1}{|\mathcal{D}_F|}
    \sum_{(g,\phi)\in\mathcal{D}_F}
    \mathbf{1}\!\left[
    \hat{y}_{g}^{\mathrm{full},\phi_{F,0}}
    =
    \hat{y}_{g}^{\mathrm{full},\phi}
    \right],
\end{equation}
where $\mathcal{D}_F=\{(g,\phi):g\in\mathcal{G},\phi\in\Phi_F(g)\}$. The baseline itself is not counted as a trivial self-comparison.

We also compute the Judge Consistency Rate (JCR). Let
$\Phi_F^{\mathrm{all}}(g)=\{\phi_{g,F,1},\ldots,\phi_{g,F,M_F}\}$ contain all
probe instances in family $F$, including the family baseline when applicable.
Then
\begin{equation}
    \label{eq:app-jcr}
    \mathrm{JCR}_{F}
    =
    \frac{1}{G}
    \sum_{g\in\mathcal{G}}
    \mathbf{1}\!\left[
    \hat{y}_{g}^{\mathrm{full},\phi_{g,F,1}}
    =
    \cdots
    =
    \hat{y}_{g}^{\mathrm{full},\phi_{g,F,M_F}}
    \right].
\end{equation}
The corresponding flip rate is
\begin{equation}
    \label{eq:app-flip}
    \mathrm{Flip}_{F}=1-\mathrm{JCR}_{F}.
\end{equation}
High NS or JCR is not sufficient for reliability, because a constant-verdict
judge can be perfectly stable.

\paragraph{Presentation leakage: LES.}
To measure whether a presentation family systematically shifts verdicts, we
estimate a leakage effect size. For each family $F$, we fit a mixed-effects
logistic model over pooled binary predictions:
\begin{equation}
    \label{eq:app-les-model}
    \begin{aligned}
    \mathrm{logit}\,\Pr(\hat{Y}_{g,\phi}=1)
    &=
    \beta_0+
    \sum_{k=1}^{K_F}\beta_{F,k}\,\mathbf{1}[\phi_F=k]
    +u_g,\\
    u_g&\sim\mathcal{N}(0,\sigma_g^2).
    \end{aligned}
\end{equation}
where $u_g$ is a group-level random intercept and the dummy variables encode
the presentation identity within family $F$. We define
\begin{equation}
    \label{eq:app-les}
    \mathrm{LES}_{F}=\max_k|\beta_{F,k}|.
\end{equation}
We also report the odds ratio $\exp(\beta_{F,k})$ and the standardized effect
size
\begin{equation}
    \label{eq:app-d-logit}
    d_{\mathrm{logit},F}=\frac{\mathrm{LES}_{F}}{\pi/\sqrt{3}}.
\end{equation}
A large and statistically significant $\mathrm{LES}_{F}$ indicates presentation
leakage: the verdict depends systematically on how the same trajectory evidence
is presented.

\paragraph{Representation exchangeability.}
For two equivalent representations $r$ and $r'$, such as composite and split
storyboards, we define the Verdict Consistency Coefficient as
\begin{equation}
    \label{eq:app-vcc}
    \mathrm{VCC}_{r,r'}
    =
    \frac{1}{|\mathcal{P}_{r,r'}|}
    \sum_{g\in\mathcal{P}_{r,r'}}
    \mathbf{1}\!\left[
    \hat{y}_{g}^{r}=\hat{y}_{g}^{r'}
    \right],
\end{equation}
where $\mathcal{P}_{r,r'}$ is the set of groups evaluated under both
representations. When all groups are paired, this reduces to an average over matched groups. In the main tables, this is reported as
$\mathrm{VCC}_{\mathrm{rep}}$ for the composite--split comparison. We can
further condition this metric on the baseline representation's predicted
verdict:
\begin{equation}
    \label{eq:app-vcc-comp-success}
    \mathrm{VCC}_{\mathrm{rep}\mid \mathrm{comp}=1}
    =
    \Pr(\hat{y}^{\mathrm{split}}=\hat{y}^{\mathrm{comp}}
    \mid \hat{y}^{\mathrm{comp}}=1),
\end{equation}
and analogously for $\hat{y}^{\mathrm{comp}}=0$. 
When oracle-state evidence is available, we additionally
measure verdict agreement between oracle-state and pixel
conditions as $\mathrm{VCC}_{\mathrm{oracle}}$.
\begin{equation}
    \label{eq:app-vcc-oracle}
    \mathrm{VCC}_{\mathrm{oracle}}
    =
    \Pr(\hat{y}^{\mathrm{oracle}}=\hat{y}^{\mathrm{pixel}}).
\end{equation}
Unlike $\mathrm{VCC}_{\mathrm{rep}}$, this is an
evidence-condition diagnostic rather than a pure C3
exchangeability measure, because the available information
differs between conditions.

\paragraph{Causal sensitivity: IDR.}
C4 is measured by whether the judge discriminates successful full trajectories
from their counterfactual failures. Define
\begin{equation}
    \label{eq:app-active-and-correct-cf}
    \begin{aligned}
    A_g&=\mathbf{1}\!\left[
    \hat{y}_{g}^{\mathrm{full},\phi_0}=1
    \right],\\
    C_g&=\mathbf{1}\!\left[
    \hat{y}_{g}^{\mathrm{cf},\phi_0}=0
    \right].
    \end{aligned}
\end{equation}
The raw Intervention Discrimination Rate is
\begin{equation}
    \label{eq:app-idr-raw}
    \mathrm{IDR}_{\mathrm{raw}}
    =
    \frac{1}{G}
    \sum_{g\in\mathcal{G}}A_gC_g.
\end{equation}
The number of active groups is
\begin{equation}
    \label{eq:app-g-active}
    \begin{aligned}
    G_{\mathrm{active}}&=\sum_{g\in\mathcal{G}}A_g,\\
    p_{\mathrm{active}}&=\frac{G_{\mathrm{active}}}{G}.
    \end{aligned}
\end{equation}
Since a model that almost never predicts success on full trajectories cannot
meaningfully be tested for full-to-counterfactual discrimination, we also report
the conditional IDR:
\begin{equation}
    \label{eq:app-idr-cond}
    \mathrm{IDR}_{\mathrm{cond}}
    =
    \frac{\sum_g A_gC_g}{G_{\mathrm{active}}},
    \quad\text{if }G_{\mathrm{active}}>0.
\end{equation}
Equivalently,
\begin{equation}
    \label{eq:app-idr-cond-equivalent}
    \mathrm{IDR}_{\mathrm{cond}}
    =
    \frac{\mathrm{IDR}_{\mathrm{raw}}}{p_{\mathrm{active}}}.
\end{equation}
We additionally track the reverse error
\begin{equation}
    \label{eq:app-idr-rev}
    \mathrm{IDR}_{\mathrm{rev}}
    =
    \Pr(\hat{y}^{\mathrm{full},\phi_0}=0
    \wedge
    \hat{y}^{\mathrm{cf},\phi_0}=1),
\end{equation}
which should be close to zero for a sensible judge.

\paragraph{Task-specific eligibility.}
For the Task-C summaries in
Table~\ref{tab:main_diagnostic},
$\mathrm{JAcc}_{\mathrm{neu}}$ is computed over
all 18 Full groups per representation, treating
unparseable Full verdicts as incorrect.
In contrast, $p_{\mathrm{active}}$ and
$\mathrm{IDR}_{\mathrm{raw}}$ use the subset
of groups with parseable verdicts for both
Full and CF. For example, Kimi K2.5 under
Composite yields $7/18$ for JAcc,
$7/14$ for activation, and $5/14$ for raw IDR.
Dedicated Task-D and Scale-2 analyses instead
use all complete groups, counting invalid outputs
as incorrect.

\paragraph{Task-specific diagnostic metrics.}
Task A reports one-step transition accuracy:
\begin{equation}
    \label{eq:app-acc-a1}
    \mathrm{Acc}_{A1}
    =
    \frac{1}{N_A}
    \sum_{i=1}^{N_A}
    \mathbf{1}\!\left[
    \hat{c}_i=c_i^\star
    \right],
\end{equation}
where $c_i^\star$ is the correct next-observation choice among four candidates.
When multi-step variants are available, we analogously report
$\mathrm{Acc}_{Ak}$ for $k$-step prediction. To test whether the model uses the
action condition, we compute action-ablation drops:
\begin{equation}
    \label{eq:app-action-ablation}
    \begin{aligned}
    \Delta_{\mathrm{shuffle}}
    &=
    \mathrm{Acc}_{A1}^{\mathrm{orig}}
    -
    \mathrm{Acc}_{A1}^{\mathrm{shuffle}},\\
    \Delta_{\mathrm{mask}}
    &=
    \mathrm{Acc}_{A1}^{\mathrm{orig}}
    -
    \mathrm{Acc}_{A1}^{\mathrm{mask}}.
    \end{aligned}
\end{equation}
If these drops are small and statistically insignificant, it provides evidence that the model is more likely to rely on an action-ignorant heuristic for this diagnostic.

Task B measures structured perception from a single frame. Let
$S_B(i)\in[0,1]$ be the component-level score for item $i$ under the canonical
schema. We report
\begin{equation}
    \label{eq:app-task-b}
    \begin{aligned}
    \mathrm{B\_MeanScore}
    &=
    \frac{1}{N_B}\sum_{i=1}^{N_B}S_B(i),\\
    \mathrm{B\_Acc}
    &=
    \frac{1}{N_B}\sum_{i=1}^{N_B}\mathbf{1}[S_B(i)\geq 0.90].
    \end{aligned}
\end{equation}
Here $S_B(i)$ is the normalized weighted component
score, with invalid or unavailable outputs assigned
zero. B\_Acc@0.90 is the high-threshold
component accuracy, not JSON exact-match accuracy.
Task E analogously reports object/state classification accuracy and mean score:
\begin{equation}
    \label{eq:app-task-e}
    \begin{aligned}
    \mathrm{E\_Acc}
    &=
    \frac{1}{N_E}\sum_{i=1}^{N_E}\mathbf{1}[\hat{z}_i=z_i^\star],\\
    \mathrm{E\_MeanScore}
    &=
    \frac{1}{N_E}\sum_{i=1}^{N_E}S_E(i).
    \end{aligned}
\end{equation}
These perception-oriented diagnostics help separate failures caused by visual
parsing or interface mismatch from failures in causal judgment.

\paragraph{Metric summary.}
Table~\ref{tab:app-metric_summary} lists the metrics used in the paper and the
primary diagnostic question each addresses.

\begin{table}[t]
    \centering
    \scriptsize
    \caption{\textbf{Summary of SCI metrics.} All metrics are empirical estimates over
    their applicable task-specific item sets.}
    \label{tab:app-metric_summary}
    \begin{tabular}{p{0.19\linewidth} p{0.26\linewidth} p{0.13\linewidth} p{0.32\linewidth}}
        \toprule
        Metric & Formula sketch & Task(s) & Interpretation \\
        \midrule
        $\mathrm{JAcc}_{\mathrm{neu}}$ &
        $\frac{1}{G}\sum_g\mathbf{1}[\hat{y}^{\mathrm{full}}_g=1]$ & C/D &
        Neutral full-trajectory correctness anchor \\
        $\mathrm{Acc}_{\mathrm{full}}$ &
        $\frac{1}{G}\sum_g\mathbf{1}[\hat{y}^{\mathrm{full}}_g=1]$ & C/D &
        Variant-wise full success acceptance \\
        $\mathrm{Acc}_{\mathrm{nocue}}$ &
        $\frac{1}{G}\sum_g\mathbf{1}[\hat{y}^{\mathrm{nocue}}_g=1]$ & C/D &
        NoCue success accuracy; evidence-removal diagnostic \\
        $\mathrm{Acc}_{\mathrm{cf}}$ &
        $\frac{1}{G}\sum_g\mathbf{1}[\hat{y}^{\mathrm{cf}}_g=0]$ & D; C if CF probes are included &
        Counterfactual-failure rejection accuracy \\
        $\mathrm{SuccC}$ &
        $\frac{1}{|\mathcal{I}|}\sum_i\mathbf{1}[\hat{y}_i=1]$ & C/D &
        Overall success-verdict rate; label-bias indicator \\
        $p_{\mathrm{active}}$ &
        $G_{\mathrm{active}}/G$ & C/D &
        Rate at which the judge accepts genuine full successes \\
        $\mathrm{IDR}_{\mathrm{raw}}$ &
        $\frac{1}{G}\sum_g A_gC_g$ & C/D &
        Joint Full--CF discrimination without conditioning away inactivity \\
        $\mathrm{IDR}_{\mathrm{cond}}$ &
        $\sum_gA_gC_g/G_{\mathrm{active}}$ & C/D &
        CF rejection conditional on accepting the matched full trajectory \\
        $\mathrm{IDR}_{\mathrm{rev}}$ &
        $\Pr(\hat{y}^{\mathrm{full}}=0\wedge\hat{y}^{\mathrm{cf}}=1)$ & D &
        Reverse Full--CF ordering error \\
        $\mathrm{NS}_F$ &
        Pairwise agreement with family baseline & C &
        Stability under a presentation family $F$ \\
        $\mathrm{JCR}_F$ &
        All-presentations-same indicator averaged over groups & C &
        Per-group consistency across all probes in $F$ \\
        $\mathrm{Flip}_F$ &
        $1-\mathrm{JCR}_F$ & C &
        Presentation-induced verdict flips \\
        $\mathrm{LES}_F$ &
        $\max_k|\beta_{F,k}|$ from logistic model & C &
        Systematic presentation leakage effect size \\
        $d_{\mathrm{logit},F}$ &
        $\mathrm{LES}_F/(\pi/\sqrt{3})$ & C &
        Standardized logit-scale leakage effect size \\
        $\mathrm{VCC}_{\mathrm{rep}}$ &
        $\frac{1}{|\mathcal{P}|}\sum_g\mathbf{1}[\hat{y}^{\mathrm{comp}}_g=\hat{y}^{\mathrm{split}}_g]$ & C &
        Composite--split representation exchangeability \\
        $\mathrm{VCC}_{\mathrm{oracle}}$ &
        $\Pr(\hat{y}^{\mathrm{oracle}}=\hat{y}^{\mathrm{pixel}})$ & Optional C/D &
        Agreement between oracle state-table and visual evidence \\
        $\mathrm{Acc}_{A1}$ &
        $\frac{1}{N_A}\sum_i\mathbf{1}[\hat{c}_i=c_i^\star]$ & A &
        One-step action-conditioned transition accuracy \\
        $\mathrm{Acc}_{Ak}$ &
        Analogous $k$-step choice accuracy & Optional A &
        Multi-step transition accuracy when available \\
        $\Delta_{\mathrm{shuffle}},\Delta_{\mathrm{mask}}$ &
        Accuracy drops from original Task A & A &
        Whether action text affects next-state prediction \\
        $\mathrm{B\_MeanScore},\mathrm{B\_Acc}$ &
        Mean weighted component score; accuracy at threshold
        $0.90$ & B &
        Structured scene-perception control \\
        $\mathrm{E\_Acc},\mathrm{E\_MeanScore}$ &
        Exact-match accuracy; mean component score & E &
        Object/state tracking control \\
        \bottomrule
    \end{tabular}
\end{table}

\subsection{Failure-Mode Taxonomy and Degenerate-Judge Criteria}
\label{app:failure_modes}

The metrics above are intended for diagnosis rather than for constructing a
single leaderboard. We therefore interpret each metric through a failure-mode
taxonomy. A model may exhibit multiple failure modes at once, and degenerate
judges are flagged before drawing conclusions from stability or intervention
metrics.

\begin{itemize}
    \item \textbf{Presentation leakage} (C1). The verdict changes under
    semantics-preserving temporal, visual, or framing transforms. This is
    reflected by low $\mathrm{NS}_F$ or $\mathrm{JCR}_F$, high
    $\mathrm{Flip}_F$, or a non-negligible leakage effect size
    $\mathrm{LES}_F$. It indicates that the judge is sensitive to how the same
    trajectory evidence is presented.

    \item \textbf{Representation fragility} (C3). Equivalent evidence formats,
    such as composite storyboards and split-image sequences, yield inconsistent
    verdicts. This is reflected by low $\mathrm{VCC}_{\mathrm{rep}}$ or, when
    available, low $\mathrm{VCC}_{\mathrm{oracle}}$. It indicates that the
    judge is bound to a particular evidence interface rather than to the
    underlying trajectory semantics.

    \item \textbf{Action-ignorant heuristic} (C2). Task-A performance does not
    decrease after action shuffling or action masking. This is reflected by
    small $\Delta_{\mathrm{shuffle}}$ or $\Delta_{\mathrm{mask}}$ despite
    nontrivial $\mathrm{Acc}_{A1}$. It suggests that the model relies on visual
    matching or terminal-frame shortcuts rather than action-conditioned
    transition reasoning.

    \item \textbf{Causal misgrounding} (C4). The model fails to distinguish a
    successful FULL trajectory from its matched counterfactual failure. This is
    reflected by low $\mathrm{IDR}_{\mathrm{raw}}$ or low/unstable
    $\mathrm{IDR}_{\mathrm{cond}}$ after checking $p_{\mathrm{active}}$ and
    $G_{\mathrm{active}}$. It indicates that the judge does not reliably
    respond to an intervention that changes the simulator outcome.

    \item \textbf{Stability trap} (cross-cutting). The model has high
    consistency under presentation probes but low correctness or low causal
    sensitivity. In this regime, high NS or JCR reflects a constant verdict or
    shortcut decision rule rather than reliable judging.
\end{itemize}

We additionally flag degenerate judges before interpreting SCI metrics:
\begin{itemize}
    \item \textbf{All-fail degenerate}: $p_{\mathrm{active}}<0.1$. The model
    rarely accepts genuine successful full trajectories, so high
    $\mathrm{Acc}_{\mathrm{cf}}$ or high stability is not evidence of
    reliability.

    \item \textbf{Success-biased degenerate}: $p_{\mathrm{active}}>0.9$ and
    $\Pr(\hat{y}^{\mathrm{cf}}=1)>0.9$ (equivalently,
    $\mathrm{Acc}_{\mathrm{cf}}<0.1$). The model accepts both full and CF
    trajectories, so high full-trajectory accuracy does not imply causal
    sensitivity.

    \item \textbf{Low-activation unstable}: $0<G_{\mathrm{active}}<10$.
    $\mathrm{IDR}_{\mathrm{cond}}$ may be reported descriptively, but it should
    not be treated as strong evidence because the active set is too small.
\end{itemize}

These labels are interpretive diagnostics, not additional tasks. They are used
to connect the four SCI conditions to the qualitative failure cases reported in
Appendix~\ref{app:results}.

\subsection{Statistical Testing and Uncertainty Estimation}
\label{app:statistical_testing}

\paragraph{Bootstrap confidence intervals.}
For reported proportions and paired metrics, including
$\mathrm{JAcc}_{\mathrm{neu}}$, $\mathrm{Acc}_{\mathrm{cf}}$,
$p_{\mathrm{active}}$, $\mathrm{IDR}_{\mathrm{raw}}$,
$\mathrm{IDR}_{\mathrm{cond}}$, NS, JCR, and VCC, we use percentile bootstrap
intervals. Because multiple items may share the same underlying trajectory
group, the default procedure is a clustered bootstrap: resample trajectory
groups with replacement, include all items associated with the resampled groups,
recompute the metric, and report the 2.5th and 97.5th percentiles over the
bootstrap estimates. This preserves within-group dependence induced by shared
states, actions, and variants.

\paragraph{Paired significance tests.}
For paired binary comparisons, such as composite versus split predictions on the
same group or baseline versus ablated Task-A predictions on the same item, we
use McNemar's exact test when the number of discordant pairs is small and the
continuity-corrected McNemar test otherwise. If $n_{01}$ and $n_{10}$ are the
two discordant counts, the large-sample statistic is
\begin{equation}
    \label{eq:app-mcnemar}
    \chi^2_{\mathrm{McNemar}}
    =
    \frac{(|n_{01}-n_{10}|-1)^2}{n_{01}+n_{10}}.
\end{equation}
Across-model Composite/Split comparisons in
Table~\ref{tab:rep_ablation} use the Wilcoxon
signed-rank test on paired model-level summary
metrics. Item-level paired binary comparisons
are analyzed separately using McNemar tests.

\paragraph{Low-activation regimes.}
$\mathrm{IDR}_{\mathrm{cond}}$ can be misleading when the active set is small.
We therefore apply the following reporting rule:
\begin{enumerate}
    \item If $G_{\mathrm{active}}=0$, $\mathrm{IDR}_{\mathrm{cond}}$ is undefined.
    \item If $0<G_{\mathrm{active}}<10$, report $\mathrm{IDR}_{\mathrm{cond}}$ only as a descriptive value, mark it as unstable, and do not use it as evidence of causal sensitivity.
    \item If $G_{\mathrm{active}}\geq 10$, report $\mathrm{IDR}_{\mathrm{cond}}$ with a clustered bootstrap confidence interval, but still interpret it together with $p_{\mathrm{active}}$ and $\mathrm{IDR}_{\mathrm{raw}}$.
\end{enumerate}
In all cases, the main C4 diagnostic plot should show both $p_{\mathrm{active}}$
and $\mathrm{IDR}_{\mathrm{raw}}$.

\paragraph{Degenerate-judge criteria.}
Degenerate-judge criteria are defined in
Section~\ref{app:failure_modes}.  In the statistical workflow below, these
criteria are applied before interpreting conditional causal-sensitivity metrics
or high stability scores.

\paragraph{Minimal workflow.}
The statistical workflow used for C4-style diagnostics is:
\begin{enumerate}
    \item compute $A_g$, $C_g$, $p_{\mathrm{active}}$, $\mathrm{IDR}_{\mathrm{raw}}$, and $\mathrm{IDR}_{\mathrm{cond}}$;
    \item bootstrap group-level confidence intervals for the reported rates;
    \item flag low-activation or degenerate regimes before interpreting $\mathrm{IDR}_{\mathrm{cond}}$;
    \item use paired tests for representation or ablation comparisons;
    \item interpret stability metrics only jointly with correctness and causal sensitivity.
\end{enumerate}

\section{Environment Details and Dataset Generation}
\label{app:env_data}

This appendix expands Section~\ref{sec:benchmark} by documenting the MiniGrid
Tier-0 environments, the construction of aligned trajectory variants, the
validation gates used before exam generation, and the dataset composition. The
purpose of this appendix is to make the simulator-grounded source of truth
inspectable: for each sampled seed we retain the symbolic state sequence, action
sequence, rendered observations, mission text, outcome labels, and audit
metadata, while exposing only the model-facing evidence to the VLM judge. Full
JSON/JSONL schemas, repository paths, decoding parameters, and response parsing
rules are provided separately in Appendix~\ref{app:implementation}.

\subsection{MiniGrid Environment Cards}
\label{app:minigrid_cards}

GridWM-Judge uses six deterministic MiniGrid environments as the 2D Tier-0
suite. They cover object interaction, ordered door opening, hazard navigation,
memory-dependent choice, and multi-room traversal. Each environment provides a
compact symbolic state, an agent-centric rendered observation, a discrete action
space, and an unambiguous success predicate. Table~\ref{tab:app_env_cards}
summarizes the environment-level causal structure used by the benchmark.

\begin{table}[b]
    \centering
    \caption{\textbf{MiniGrid environment cards.} The table lists the causal evidence
    that the judge should use, the evidence removed in \textsf{NoCue}, and the
    minimal counterfactual intervention used to create matched failures.}
    \label{tab:app_env_cards}
    \scriptsize
    \setlength{\tabcolsep}{3pt}
    \renewcommand{\arraystretch}{1.08}
    \begin{tabular}{p{0.13\linewidth}p{0.25\linewidth}p{0.25\linewidth}p{0.25\linewidth}}
        \toprule
        Environment & Causal chain and success condition
        & \textsf{NoCue} evidence removal
        & \textsf{CF} intervention \\
        \midrule
        DoorKey
        & Pick up key $\rightarrow$ unlock/open matching door $\rightarrow$ reach goal. Success requires positive reward and terminal goal entry.
        & Mask early frames in which the key is visible before pickup.
        & Relocate or remove the goal at $t=0$ so that the unchanged action sequence no longer reaches the true goal. \\
        KeyCorridor
        & Pick up color-matched key $\rightarrow$ return to locked door $\rightarrow$ unlock door $\rightarrow$ reach the mission target object.
        & Mask the door-matched key in the pre-pickup window.
        & Remove the mission target object before rollout, so the same action sequence cannot satisfy the pickup objective. \\
        RedBlueDoor
        & Open red door $\rightarrow$ open blue door $\rightarrow$ pass through the ordered-door sequence. Opening blue before red is invalid.
        & Mask the red door before the first red-door interaction; blue-door evidence is retained.
        & Replace or block the blue-door passage before rollout, preserving the early red-door subtask but preventing final success. \\
        LavaGap
        & Identify the single safe gap in the lava barrier $\rightarrow$ cross through the gap $\rightarrow$ reach goal without stepping on lava.
        & Mask visible lava evidence in early pre-crossing frames.
        & Relocate or remove the goal outside the planned path while keeping the safe-crossing behavior physically valid. \\
        Memory
        & Observe cue object $\rightarrow$ traverse corridor $\rightarrow$ choose the matching terminal object.
        & Mask the cue object before corridor entry.
        & Swap the correct and distractor terminal objects before rollout, so the same remembered choice becomes wrong. \\
        MultiRoom
        & Traverse a sequence of rooms, opening intervening doors, then enter the goal room and reach the goal.
        & Mask early door/goal evidence before entering the goal-containing room.
        & Remove the final goal object before rollout, preserving room traversal but preventing terminal success. \\
        \bottomrule
    \end{tabular}
\end{table}

\begin{figure}[t]
    \centering
    \includegraphics[width=0.95\linewidth]{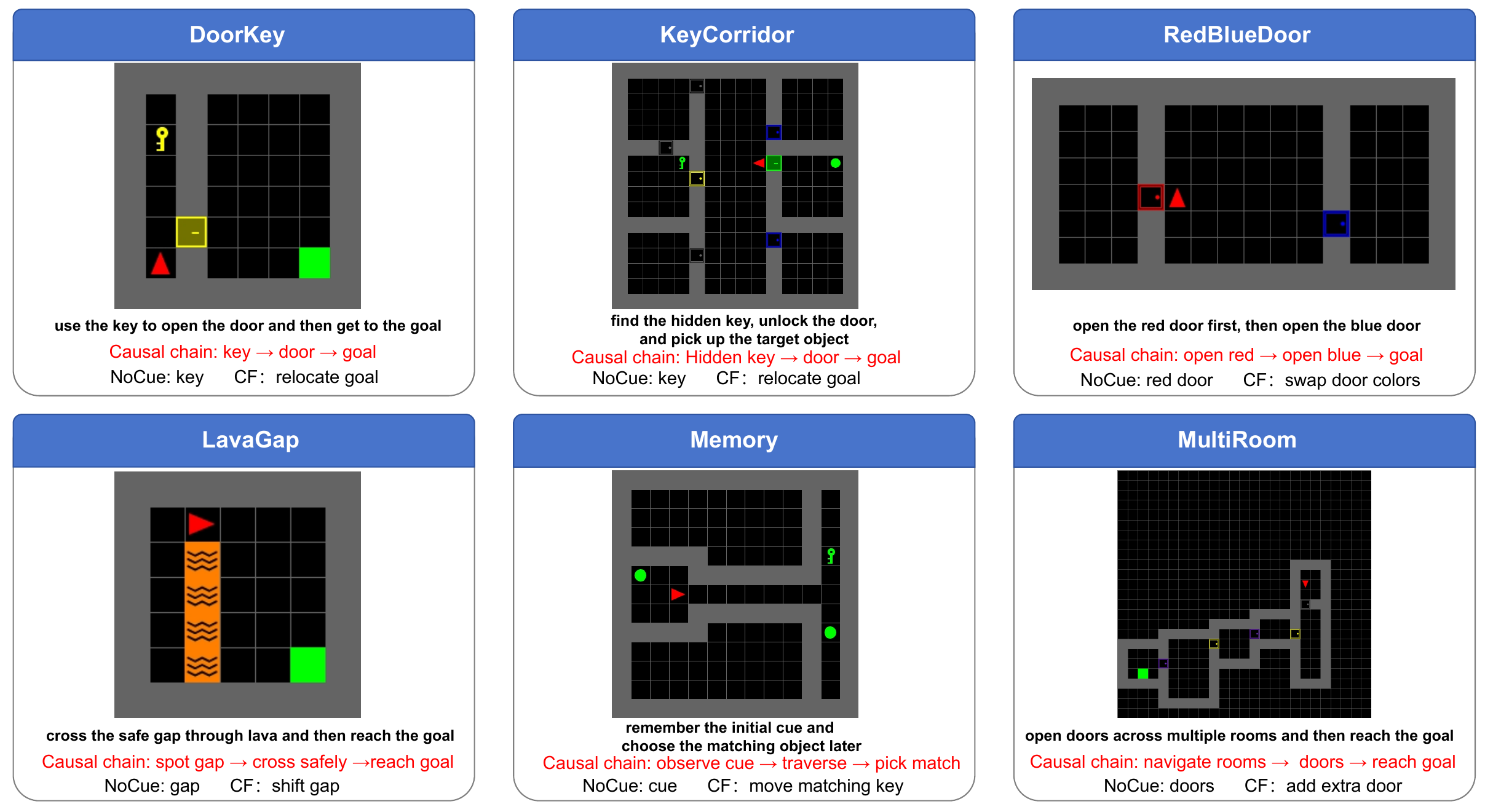}
    \caption{\textbf{Representative MiniGrid environment cards.} Each panel shows one canonical seed, the mission text, and a short causal-chain annotation.}
    \label{fig:app_env_cards}
\end{figure}

Figure~\ref{fig:app_env_cards} visualizes representative seeds and causal chains for the six MiniGrid environments summarized in Table~\ref{tab:app_env_cards}. For each environment, object identities and causal events are inferred from the
simulator state rather than from rendered pixels alone. DoorKey records whether
the agent is carrying the key and the door state; KeyCorridor resolves the
mission target and the color-matched key/door pair; RedBlueDoor checks the
red-before-blue ordering; LavaGap identifies the barrier and the unique safe
gap; Memory identifies the cue object and the matching terminal object; and
MultiRoom tracks door toggles, room transitions, and entry into the goal room.
These symbolic event traces are used only for generation and validation, not as
model-facing inputs.

\subsection{Dataset Generation Pipeline}
\label{app:generation_pipeline}

\begin{figure}[t]
    \centering
    \includegraphics[width=0.95\linewidth]{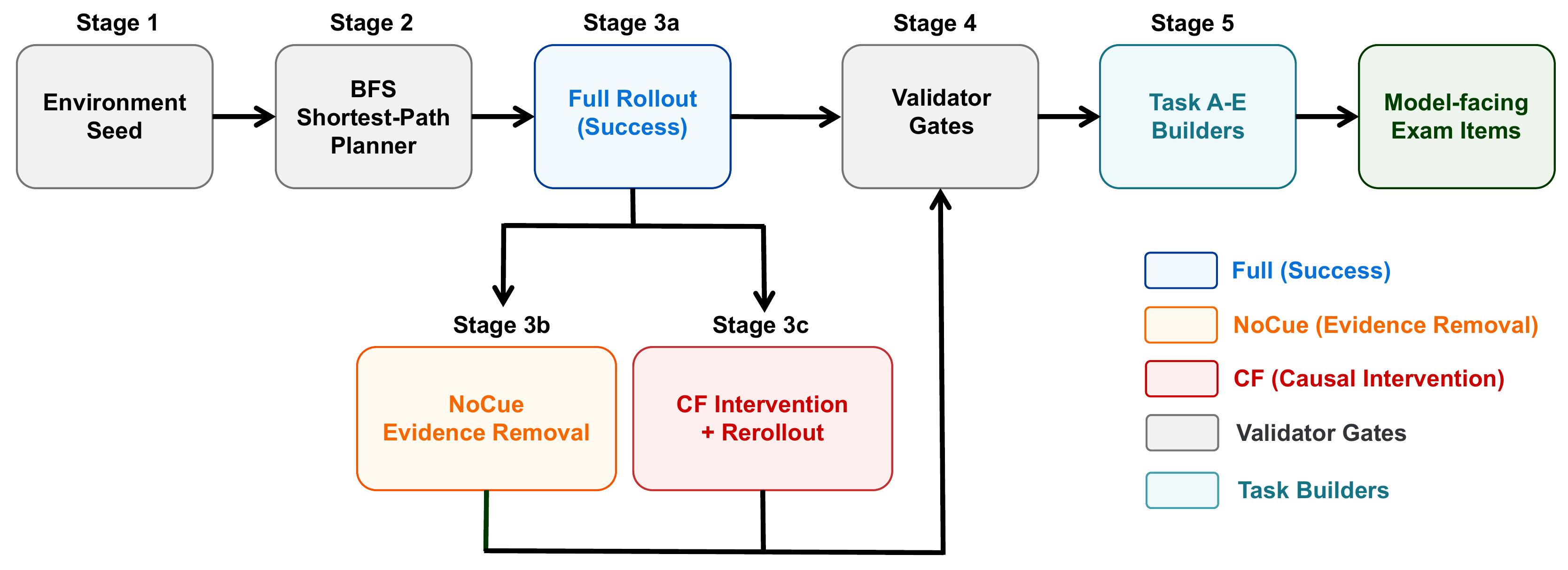}
    \caption{\textbf{Dataset generation pipeline from environment seed to validated
    exam item.} Each seed first produces a successful \textsf{Full} rollout; the
    aligned \textsf{NoCue} and \textsf{CF} variants are then constructed,
    validated, and converted into task-specific exam items. The model-facing
    prompts contain only rendered evidence and task instructions, while symbolic
    states, labels, and intervention metadata remain audit-only.}
    \label{fig:app_generation_pipeline}
\end{figure}

The atomic unit for Tasks C and D is a trajectory group
$g=(s_{0:T},a_{0:T-1},m,\ell,\eta)$, containing a simulator state sequence,
action sequence, mission text, outcome label, and audit metadata. A trajectory
group contains a successful \textsf{Full} rollout, an evidence-removal
\textsf{NoCue} variant, and a counterfactual \textsf{CF} variant. Tasks A, B,
and E are derived from the same validated simulator rollouts but expose more
localized transition, perception, or prefix-checkpoint interfaces, as described
in Appendix~\ref{app:tasks}. Figure~\ref{fig:app_generation_pipeline} shows the full generation pipeline from
environment seed to validated model-facing exam items.

\paragraph{Stage 1: successful full rollout.}
For each environment seed, we reset the simulator with a deterministic layout
and use an environment-specific shortest-path planner to construct a successful
rollout. The planner operates over symbolic states and inserts required
interaction actions, such as pickup and toggle. Rollouts with truncation,
missing critical events, or ambiguous terminal outcomes are rejected before any
exam item is produced.

\begin{figure}[t]
    \centering
    \includegraphics[width=0.95\linewidth]{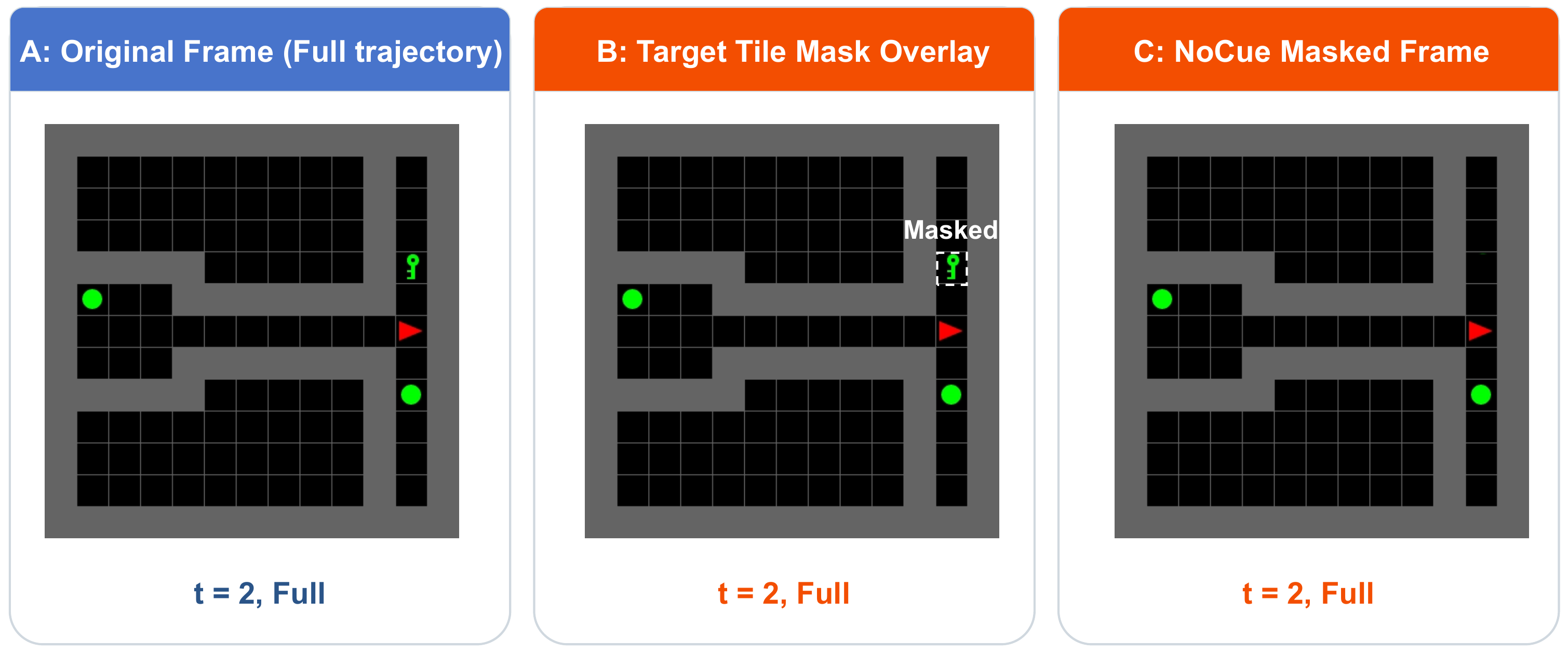}
    \caption{\textbf{Example \textsf{NoCue} audit.} The panels show the original frame, the target-tile mask overlay, and the resulting \textsf{NoCue} masked frame used to remove selected early evidence while preserving the rollout outcome.}
    \label{fig:app_nocue_audit}
\end{figure}

\paragraph{Stage 2: evidence-removal variant.}
The \textsf{NoCue} variant keeps the same action sequence and the same simulator
outcome as \textsf{Full}, but masks selected early rendered tiles that reveal the
causal evidence. The masked frames must occur strictly before the first
interaction with the target evidence. Thus \textsf{NoCue} is an
outcome-preserving evidence-removal diagnostic: it tests whether the judge
relies on early causal evidence, but it is not counted as a
semantics-preserving presentation transform for C1 or as a causal intervention
for C4. Table~\ref{tab:app_nocue_params} lists the environment-specific evidence targets, mask budgets, and alignment thresholds used for \textsc{NoCue} construction. Figure~\ref{fig:app_nocue_audit} illustrates a representative \textsc{NoCue} audit,
showing how the target evidence is masked while preserving the rollout outcome.

\begin{table}[b]
    \centering
    \caption{\textbf{Per-environment \textsf{NoCue} targets and masking constraints.}
    The mask budget is intentionally conservative so that masking removes the
    target evidence without changing the overall trajectory presentation.}
    \label{tab:app_nocue_params}
    \scriptsize
    \setlength{\tabcolsep}{4pt}
    \renewcommand{\arraystretch}{1.08}
    \begin{tabular}{p{0.20\linewidth}p{0.33\linewidth}p{0.18\linewidth}p{0.18\linewidth}}
        \toprule
        Environment & Masked target evidence & Max masked frames & Alignment threshold \\
        \midrule
        DoorKey & Key before pickup & 5 & $0.70$ \\
        KeyCorridor & Color-matched key before pickup & 5 & $0.70$ \\
        RedBlueDoor & Red door before first red-door toggle & 5 & $0.70$ \\
        LavaGap & Lava evidence before safe crossing & 4 & $0.70$ \\
        Memory & Cue object before corridor entry & 10 & $0.70$ \\
        MultiRoom & Door/goal evidence before goal-room entry & 12 & $0.70$ \\
        \bottomrule
    \end{tabular}
\end{table}

\paragraph{Stage 3: counterfactual failure variant.}
The \textsf{CF} variant applies a minimal simulator intervention before action
execution and then rerolls the unchanged action sequence. Depending on the
environment, the intervention removes, relocates, swaps, or blocks the causal
target. The validation contract requires the counterfactual rollout to be a
physically valid execution of the same action sequence and to fail for the
intended causal reason. Exact positional identity with \textsf{Full} is
required for \textsf{NoCue}; for \textsf{CF}, state divergence is allowed only
when it is induced by the declared intervention under the unchanged actions.

\paragraph{Stage 4: rendering and exam assembly.}
Validated rollouts are rendered into the presentation formats used by the task
interfaces. Task C uses event-anchored sparse storyboards and presentation
probes such as temporal order and visual style changes. Task D uses matched
\textsf{Full}/\textsf{CF} trajectory evidence for causal-sensitivity tests.
Task E uses prefix-only checkpoint evidence and deliberately hides terminal
frames. The detailed prompt templates, split-image/composite layouts, and
example questions are provided in Appendix~\ref{app:tasks}.

\subsection{Validator Gates and Quality Control}
\label{app:validator_gates}

All generated groups pass validation before they enter the benchmark. The
validators prevent data-generation artifacts from being mistaken for VLM-judge
failures. Table~\ref{tab:app_validator_gates} summarizes the gates used in the
release pipeline.

\begin{table}[t]
    \centering
    \caption{\textbf{Validator gates.} The gates jointly enforce action alignment,
    outcome correctness, physically valid rerollout, and correct isolation of
    model-facing evidence from audit-only state.}
    \label{tab:app_validator_gates}
    \scriptsize
    \setlength{\tabcolsep}{3pt}
    \renewcommand{\arraystretch}{1.08}
    \begin{tabular}{p{0.10\linewidth}p{0.20\linewidth}p{0.52\linewidth}p{0.10\linewidth}}
        \toprule
        Gate & Purpose & Rule & Scope \\
        \midrule
        D1 & Action identity
        & The action IDs are identical across matched variants in the same group.
        & all variants \\
        D2 & Length consistency
        & Frames and states align with the action sequence: $|s_{0:T}|=|a_{0:T-1}|+1$.
        & all variants \\
        D3 & Input isolation
        & Model-facing inputs contain only rendered evidence, mission/action text, and task instructions; labels, rewards, states, and intervention metadata remain audit-only.
        & all tasks \\
        D4 & Outcome contract
        & \textsf{Full} succeeds, \textsf{NoCue} preserves the \textsf{Full} outcome, and \textsf{CF} fails without truncation artifacts.
        & C/D \\
        D5 & No interaction masking
        & \textsf{NoCue} never masks the frame in which the agent is facing or interacting with the target evidence.
        & NoCue \\
        D6 & Physics validity
        & \textsf{NoCue} preserves the exact state trajectory; \textsf{CF} is a valid rerollout under the same action sequence, with divergence allowed only as a consequence of the declared intervention.
        & all variants \\
        E1 & Causal-chain completeness
        & Environment-specific critical events occur in the required order, e.g., key before door, red before blue, gap crossing without lava contact, cue before terminal choice.
        & all envs \\
        E2 & State encoding validity
        & Symbolic states use valid MiniGrid object, color, and state indices and remain aligned with rendered frames.
        & all variants \\
        N1 & Mask coverage/alignment
        & Masked pixels overlap the intended target tile with mean IoU at least $0.70$ and satisfy the global mask budget.
        & NoCue \\
        CF1 & Intervention audit
        & The counterfactual target is removed, relocated, swapped, or blocked according to the environment card, and the failure is attributable to that intervention.
        & CF \\
        \bottomrule
    \end{tabular}
\end{table}

Environment-specific semantic checks instantiate E1. DoorKey requires key pickup
before door opening; KeyCorridor requires the color-matched key before the
locked-door interaction; RedBlueDoor requires the red door to be opened before
the blue door; LavaGap requires crossing through the safe gap and never stepping
on lava; Memory requires cue observation before the terminal choice; and
MultiRoom requires the expected sequence of door toggles and goal-room entry.

\subsection{Dataset Composition}
\label{app:dataset_composition}
We report the benchmark in terms of exam rows rather than only raw trajectories,
because the same validated trajectory group can instantiate multiple variants
and presentation conditions. Table~\ref{tab:app_dataset_composition} summarizes
the paper-facing MiniGrid benchmark used in the main experiments. The table
describes the diagnostic role of each task, the source environments, the number
of model-facing exam rows, and the variant or probe structure used to construct
those rows.

\begin{table}[t]
    \centering
    \scriptsize
    \setlength{\tabcolsep}{3pt}
    \renewcommand{\arraystretch}{1.10}
    \caption{\textbf{Dataset composition for the MiniGrid benchmark.}
    Counts are reported at the exam-row level, since a single validated
    trajectory group may produce multiple model-facing items through variants,
    presentation probes, or task-specific interfaces.}
    \label{tab:app_dataset_composition}
    \begin{tabularx}{\linewidth}{@{}
        c
        >{\raggedright\arraybackslash}p{0.25\linewidth}
        >{\raggedright\arraybackslash}p{0.20\linewidth}
        c
        >{\raggedright\arraybackslash}X
    @{}}
        \toprule
        \textbf{Task} & \textbf{Diagnostic role} & \textbf{Environments} & \textbf{Exam rows} & \textbf{Variant/probe structure} \\
        \midrule
        A &
        Action-conditioned next-observation prediction &
        Six core environments &
        103 &
        Local transitions from 18 source trajectories; \newline 103 baseline questions \\
        \addlinespace[0.25em]

        B &
        Single-frame perception and interface control &
        Six core environments &
        144 &
        $6\times12\times2=144$: 12 source groups per environment, two frames per group; \newline structured scene parsing \\
        \addlinespace[0.25em]

        C &
        Presentation invariance and representation sensitivity &
        KeyCorridor, MultiRoom, RedBlueDoor &
        486 &
        18 groups: $18\times3\times2\times3=324$ temporal/visual items, plus $18\times3\times3=162$ framing items \\
        \addlinespace[0.25em]

        D &
        Causal sensitivity under matched intervention &
        Six core environments &
        144 &
        $6\times2\times3\times4=144$: two query groups per environment, three variants, four presentation conditions; successful Full references \\
        \addlinespace[0.25em]

        E &
        Prefix/checkpoint perception control &
        UnlockPickup, BlockedUnlockPickup &
        325 &
        48 source trajectories (24 per environment); 136 identity and 189 checkpoint-status questions \\
        \bottomrule
    \end{tabularx}
\end{table}

The five tasks serve complementary diagnostic roles rather than forming a
single undifferentiated leaderboard. Task A probes local action-conditioned
transition understanding; Task B and Task E provide perception and state
recognition controls; Task C instantiates presentation and representation probes;
and Task D provides the matched intervention setting for causal-sensitivity
analysis. For Task D, the primary C4 comparison is the paired
\textsf{Full}/\textsf{CF} judgment, while retained \textsf{NoCue} rows are used
only as auxiliary evidence-removal diagnostics. Model-specific run information,
aggregate exclusions, and per-model results are reported separately in the
experimental appendices.

\subsection{Data Format and Access Boundary}
\label{app:data_boundary}

Each exam item separates the evidence shown to the VLM judge from the simulator
metadata used for auditing and scoring. The model-facing payload consists of
the rendered image or image sequence, the mission or action text when required
by the task, the task instruction, and the requested answer format. Audit-only
fields include the ground-truth label, success flag, reward, termination status,
symbolic state trace, action IDs, random seed, \textsf{NoCue} metadata, and
\textsf{CF} intervention metadata. This separation is a core part of the
benchmark design: the VLM must judge from the same visual/textual evidence that
appears in the prompt, while the simulator state remains available for
reproducible validation.

A simplified schematic of an exam item is shown below:
\begin{verbatim}
{
  "uid": "C.keycorridor.s000001.full.orig.clean",
  "task": "C",
  "env_task": "keycorridor",
  "group_id": "s000001",
  "variant": "full",
  "presentation": {
    "temporal": "orig",
    "visual": "clean",
    "rep": "composite"
  },
  "model_input": {
    "image": "...",
    "prompt": "..."
  },
  "audit_only": {
    "label": 1,
    "success": true,
    "state_seq": "...",
    "cf_meta": null
  }
}
\end{verbatim}

The exact field names, file locations, dataset card, license, model identifiers,
API parameters, and response parsing rules are documented in
Appendix~\ref{app:implementation}. Keeping those implementation details in
Appendix~\ref{app:implementation} avoids duplicating schema definitions across
appendices and keeps Appendix~\ref{app:env_data} focused on the
data-generation logic that supports the main SCI claims.

\section{Task Interfaces and Prompt Templates}
\label{app:tasks}

\begin{figure}[t]
    \centering
    \IfFileExists{figs/appC_task_interface_gallery.pdf}{
        \includegraphics[width=\linewidth]{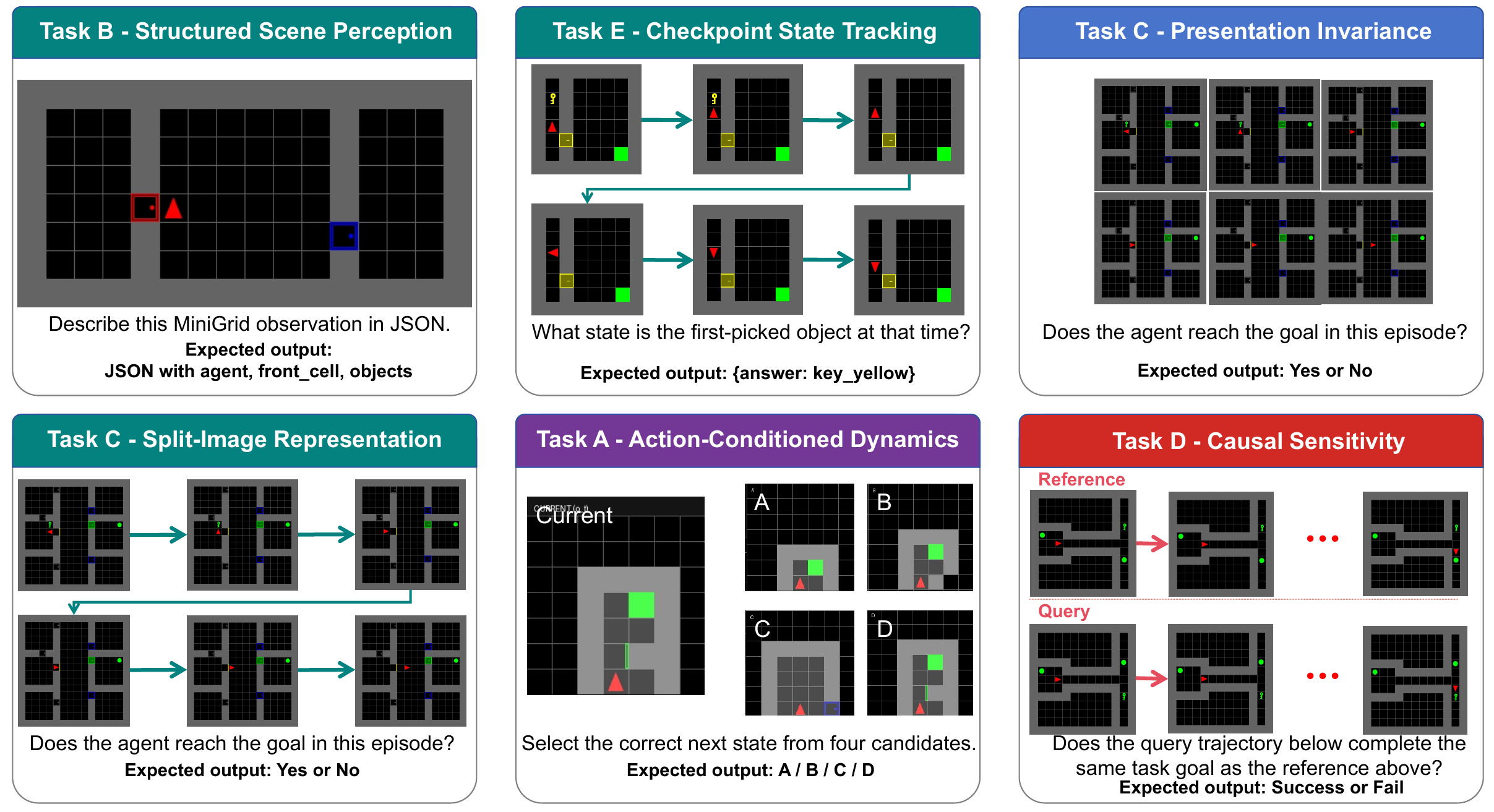}
    }{
        \fbox{
        \begin{minipage}[c][0.28\textheight][c]{0.96\linewidth}
            \centering
            \textbf{Placeholder for \texttt{figs/appC\_task\_interface\_gallery.pdf}}\\[0.5em]
            Recommended panels: Task B single-frame perception; Task E prefix-only
            checkpoint query; Task C composite storyboard; Task C split-image
            representation of the same trajectory; Task A current observation,
            action, and four candidate next observations; Task D matched
            \textsf{Full}/\textsf{CF} trajectory judgment pair.
        \end{minipage}
        }
    }
    \caption{\textbf{Representative model-facing interfaces for the five GridWM-Judge
    tasks.} Task B uses a single-frame structured perception query; Task E uses a
    prefix-only checkpoint storyboard; Task C presents the same trajectory through
    composite or split-image evidence formats; Task A asks for the correct
    action-conditioned next observation; and Task D compares matched
    \textsf{Full}/\textsf{CF} trajectory judgments. Audit-only simulator states,
    labels, rewards, and intervention metadata are not shown to the model.}
    \label{fig:appC_task_interface_gallery}
\end{figure}

GridWM-Judge contains five diagnostic task interfaces. This appendix specifies
what each model-facing item shows to the VLM judge, what answer format is
expected, and how the gold answer is obtained from simulator-grounded metadata.
Appendix~\ref{app:definitions} gives the mathematical definitions of the
metrics, Appendix~\ref{app:env_data} describes trajectory generation and
validation, and Appendix~\ref{app:implementation} gives the full
machine-readable schemas, decoding settings, and unified output parsing rules.
Throughout this appendix, \emph{model-facing} fields denote information provided
to the VLM judge, whereas labels, rewards, symbolic states, intervention records,
and audit metadata are used only for scoring and validation. Table~\ref{tab:appC_task_summary} summarizes the model-facing evidence, required
answer format, and diagnostic role for the five task interfaces. Figure~\ref{fig:appC_task_interface_gallery} gives representative model-facing layouts for
the five tasks, illustrating the difference between local controls, presentation
probes, transition prediction, and causal trajectory judgment.

\begin{table}[b]
    \centering
    \footnotesize
    \setlength{\tabcolsep}{3pt}
    \renewcommand{\arraystretch}{1.10}
    \caption{\textbf{Summary of the five task interfaces.} Appendix C describes the
    model-facing input/output contract; Appendix A defines the metrics computed
    from these outputs.}
    \label{tab:appC_task_summary}
    \begin{tabularx}{\linewidth}{@{}
        c
        >{\raggedright\arraybackslash}p{0.30\linewidth}
        >{\raggedright\arraybackslash}p{0.27\linewidth}
        >{\raggedright\arraybackslash}X
    @{}}
        \toprule
        \textbf{Task} & \textbf{Model-facing evidence} & \textbf{Required answer} & \textbf{Diagnostic role} \\
        \midrule
        B &
        Single MiniGrid frame with a generic scene-description instruction &
        Structured JSON scene description or component labels &
        Perception and interface control \\
        \addlinespace[0.25em]

        E &
        Prefix-only sparse storyboard before the terminal state &
        JSON label for the first picked object or checkpoint status &
        Object/state tracking control \\
        \addlinespace[0.25em]

        C &
        Sparse trajectory storyboard with presentation, framing, and representation probes &
        Binary success/failure verdict &
        Presentation invariance; representation exchangeability \\
        \addlinespace[0.25em]

        A &
        Current observation, action condition, and four candidate next observations &
        Multiple-choice next-state label &
        Action-conditioned transition diagnostic \\
        \addlinespace[0.25em]

        D &
        Full-trajectory evidence for matched \textsc{Full} and \textsc{CF} variants &
        Binary success/failure verdict &
        Causal sensitivity under intervention \\
        \bottomrule
    \end{tabularx}
\end{table}

\subsection{Task B and E: Perception Controls}
\label{app:tasks:perception}

Tasks B and E are local controls. They are not intended to establish SCI by
themselves; instead, they test whether the model can parse the visual interface
before higher-level failures in Tasks C and D are interpreted as presentation or
causal-reasoning failures.

\paragraph{Task B: structured scene perception.}
Task B presents a single egocentric MiniGrid observation and asks the model to
produce a structured description of the local state. The item includes a generic
instruction explaining the requested fields, but it does not include the mission,
action sequence, reward, terminal flag, or symbolic state.

\begin{quote}
\small
\textbf{Prompt template.}
You are given one MiniGrid observation. Describe the visible local state using
JSON only. Include the agent direction, whether the agent is carrying an object,
the front cell, and visible objects. Do not infer hidden objects.
\end{quote}

The canonical output is a JSON object with fields such as \texttt{agent},
\texttt{front\_cell}, and \texttt{objects}. A representative schema-level
example is:
\begin{verbatim}
{
  "agent": {
    "dir": "right",
    "carrying": null
  },
  "front_cell": {
    "type": "key",
    "color": "yellow",
    "state": null
  },
  "objects": [
    {"type": "key", "color": "yellow", "relative_pos": "front"},
    {"type": "door", "color": "yellow", "state": "locked"}
  ]
}
\end{verbatim}

Gold answers are derived from simulator-grounded
annotations corresponding to the rendered frame.
Task B reports a normalized weighted component score
and Acc@0.90, the fraction of items whose score reaches 0.90. Invalid or unavailable outputs receive
zero in the conservative all-row analysis.

\paragraph{Task E: checkpoint-style long-range state tracking.}
Task E presents a prefix-only sparse storyboard from frame $0$ through a
checkpoint $t_c$, where $t_c$ is strictly before the terminal frame. The model
must answer an object-identity or object-status question without using a terminal
shortcut. The item does not expose symbolic state, reward, success labels, or
future frames after $t_c$.

\begin{quote}
\small
\textbf{Identity-query template.}
Below is a MiniGrid trajectory storyboard showing the trajectory up to the last
shown frame only. Earlier in the shown trajectory, what was the first object the
agent picked up? Answer using JSON only:
\texttt{\{"answer":"<label>"\}}.
\end{quote}

\begin{quote}
\small
\textbf{Status-query template.}
Below is a MiniGrid trajectory storyboard showing the trajectory up to the last
shown frame only. By the last shown frame, what is the status of the first object
the agent picked up earlier? Answer using JSON only:
\texttt{\{"answer":"<label>"\}}.
\end{quote}

The identity label space contains object--color labels such as
\texttt{key\_yellow}, \texttt{key\_green}, \texttt{key\_purple},
\texttt{ball\_yellow}, \texttt{ball\_green}, and related colors appearing in the
canonical artifact. The status label space contains \texttt{holding},
\texttt{dropped}, and \texttt{reacquired\_and\_holding}. Gold labels are derived
from the simulator's object-tracking events before or at $t_c$: the first pickup
event determines the tracked object, and the object's checkpoint inventory/grid
status determines the status label.

\subsection{Task C: Presentation Invariance Tests}
\label{app:tasks:taskc}

\begin{table}[b]
    \centering
    \small
    \setlength{\tabcolsep}{3pt}
    \renewcommand{\arraystretch}{1.08}
    \caption{\textbf{Task C probe axes and their interpretation.} Presentation and
    representation probes are evaluated within a fixed trajectory condition.
    When \textsf{CF} rows are present, \textsf{Full}--\textsf{CF} differences
    are treated as causal/outcome changes rather than semantics-preserving
    presentation changes.}
    \label{tab:appC_taskc_axes}
    \begin{tabular}{p{0.20\linewidth}p{0.32\linewidth}p{0.36\linewidth}}
        \toprule
        Axis & Values & Interpretation \\
        \midrule
        Trajectory condition
        & \textsf{Full}, \textsf{NoCue}, \textsf{CF}
        & Changes the evidence condition or the underlying outcome; not all levels are semantics-preserving. \\
        Temporal order
        & original, reversed
        & C1 presentation probe over frame ordering. \\
        Visual style
        & clean, noisy, style-transformed
        & C1 presentation probe over low-level rendering perturbations. \\
        Prompt framing
        & neutral, positive, negative
        & C1 presentation probe over wording or prior-framing changes, without changing the task goal or evidence. \\
        Representation
        & composite storyboard, split images
        & C3 equivalent-evidence-format probe for representation exchangeability. \\
        \bottomrule
    \end{tabular}
\end{table}

Task C is the main interface for presentation-aware binary judging. It asks
whether a trajectory satisfies the task goal, while varying how the same
trajectory evidence is presented. In the SCI terminology, temporal order, visual
style, and prompt framing instantiate C1 presentation probes, while composite
versus split-image evidence formats instantiate the C3 representation
exchangeability probe. These changes should not alter the verdict when the
underlying simulator semantics are unchanged.

\paragraph{Input and answer contract.}
The model sees a sparse storyboard built from event-anchored keyframes and a
short success/failure question. The mission text is included when the item uses
an environment-level goal. The action sequence, symbolic states, terminal flag,
reward, intervention metadata, and gold label are not shown.

\begin{quote}
\small
\textbf{Neutral prompt template.}
Below is a storyboard from one MiniGrid trajectory.\par
Task goal: \texttt{\{mission\_string\}}\par
Did the trajectory satisfy this goal? Answer with ONLY: Yes or No.
\end{quote}

The output is mapped to a binary verdict, with \texttt{Yes}/\texttt{Success}
corresponding to success and \texttt{No}/\texttt{Fail} corresponding to failure.
The exact parser and invalid-response handling are specified in
Appendix~\ref{app:implementation}.

\begin{figure}[t]
    \centering
    \IfFileExists{figs/appC_taskc_representation.pdf}{
        \includegraphics[width=0.85\linewidth]{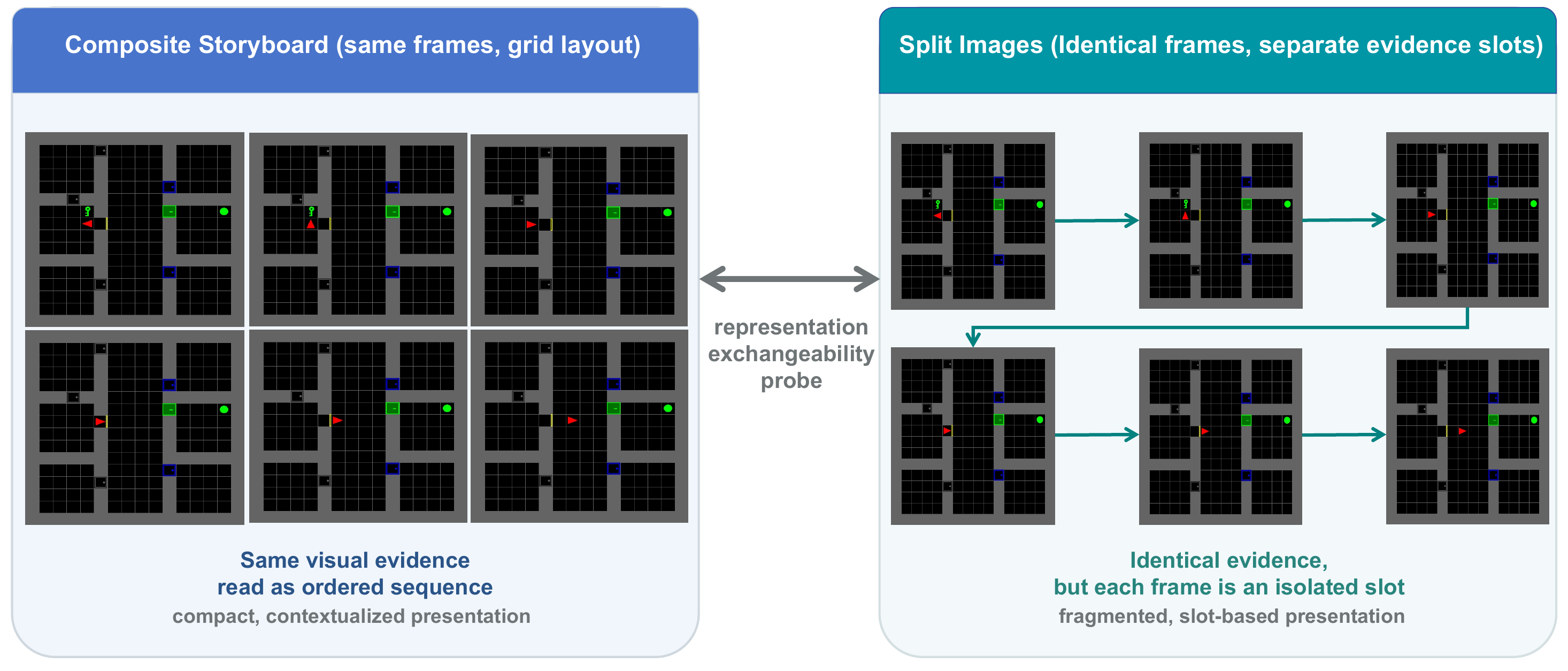}
    }{
        \fbox{
        \begin{minipage}[c][0.22\textheight][c]{0.9\linewidth}
            \centering
            \textbf{Placeholder for \texttt{figs/appC\_taskc\_representation.pdf}}\\[0.5em]
            Two panels side by side: (left) a composite storyboard that arranges
            selected keyframes into a compact grid; (right) the same keyframes
            presented as individually numbered split images. Both panels show
            the same DoorKey trajectory with identical evidence, differing only
            in evidence layout. Annotation: ``same semantics, different evidence
            interface''.
        \end{minipage}
        }
    }
    \caption{\textbf{Composite storyboard versus split-image representation for Task C.}
    The left panel arranges selected event-anchored keyframes into a compact storyboard
    (Composite format). The right panel presents the same keyframes as individually
    numbered images (Split-image format). Both formats convey identical trajectory
    semantics evidence. }
    \label{fig:appC_composite_vs_split}
\end{figure}

\paragraph{Probe axes.}
Table~\ref{tab:appC_taskc_axes} summarizes the Task C trajectory, presentation, and representation axes. Task C items are organized by trajectory condition and presentation condition.
The trajectory condition may be \textsf{Full}, \textsf{NoCue}, or \textsf{CF};
only the presentation and representation axes are used as
semantics-preserving invariance probes within a fixed trajectory condition.
\textsf{NoCue} removes early visual evidence while preserving the successful
outcome, and \textsf{CF} changes the underlying outcome through a simulator
intervention. Therefore, full--counterfactual changes should be interpreted as
causal sensitivity rather than presentation invariance.

\paragraph{Storyboard and representation formats.}
Figure~\ref{fig:appC_composite_vs_split} illustrates the composite storyboard and split-image representations used for the C3 exchangeability probe.
For composite inputs, the selected keyframes are arranged into a compact grid and
read left-to-right, top-to-bottom unless the temporal probe reverses the order.
For split-image inputs, the same selected frames are supplied as separate images slots.
Exact prompts and run conditions are documented with the associated evaluation artifacts.
The selected frames are event-complete:
they prioritize causally relevant events such as first object visibility, pickup,
door toggling, hazard crossing, goal visibility, and terminal evidence. Matched
conditions for the same trajectory group should use the same frame indices
whenever the corresponding frames exist, so that differences in verdicts can be
attributed to the intended presentation or causal variable rather than accidental
frame selection.

\paragraph{Visual and framing variants.}
The \texttt{clean} condition uses the original rendering. The \texttt{noisy}
condition applies mild deterministic image noise that should not alter the
MiniGrid state. The \texttt{style} condition applies deterministic color,
contrast, brightness, or sharpness adjustments. Prompt-framing variants should
change only the wording around the same evidence and goal; they must not add
hints, labels, rewards, or causal explanations. For example, positive framing can
say that the trajectory may have succeeded, whereas negative framing can say that
it may have failed, but both must ask the same binary question over the same
visual evidence.

\paragraph{Framing prompt templates.}
All framing variants use the same images, mission text, and answer format; only
the preamble changes. The neutral framing is:
\begin{quote}\small
Below is a storyboard from one MiniGrid trajectory.\\
Task goal: \texttt{\{mission\_string\}}\\
Did the trajectory satisfy this goal? Answer with ONLY: Yes or No.
\end{quote}
The positive framing is:
\begin{quote}\small
The trajectory may show successful progress toward the task goal. Please judge
only from the visual evidence.\\
Task goal: \texttt{\{mission\_string\}}\\
Did the trajectory satisfy this goal? Answer with ONLY: Yes or No.
\end{quote}
The negative framing is:
\begin{quote}\small
The trajectory may contain a failure to complete the task goal. Please judge
only from the visual evidence.\\
Task goal: \texttt{\{mission\_string\}}\\
Did the trajectory satisfy this goal? Answer with ONLY: Yes or No.
\end{quote}

\subsection{Task A: Action-Conditioned Dynamics}
\label{app:tasks:taska}

\begin{figure}[b]
    \centering
    \IfFileExists{figs/appC_taskA_layout.pdf}{
        \includegraphics[width=0.9\linewidth]{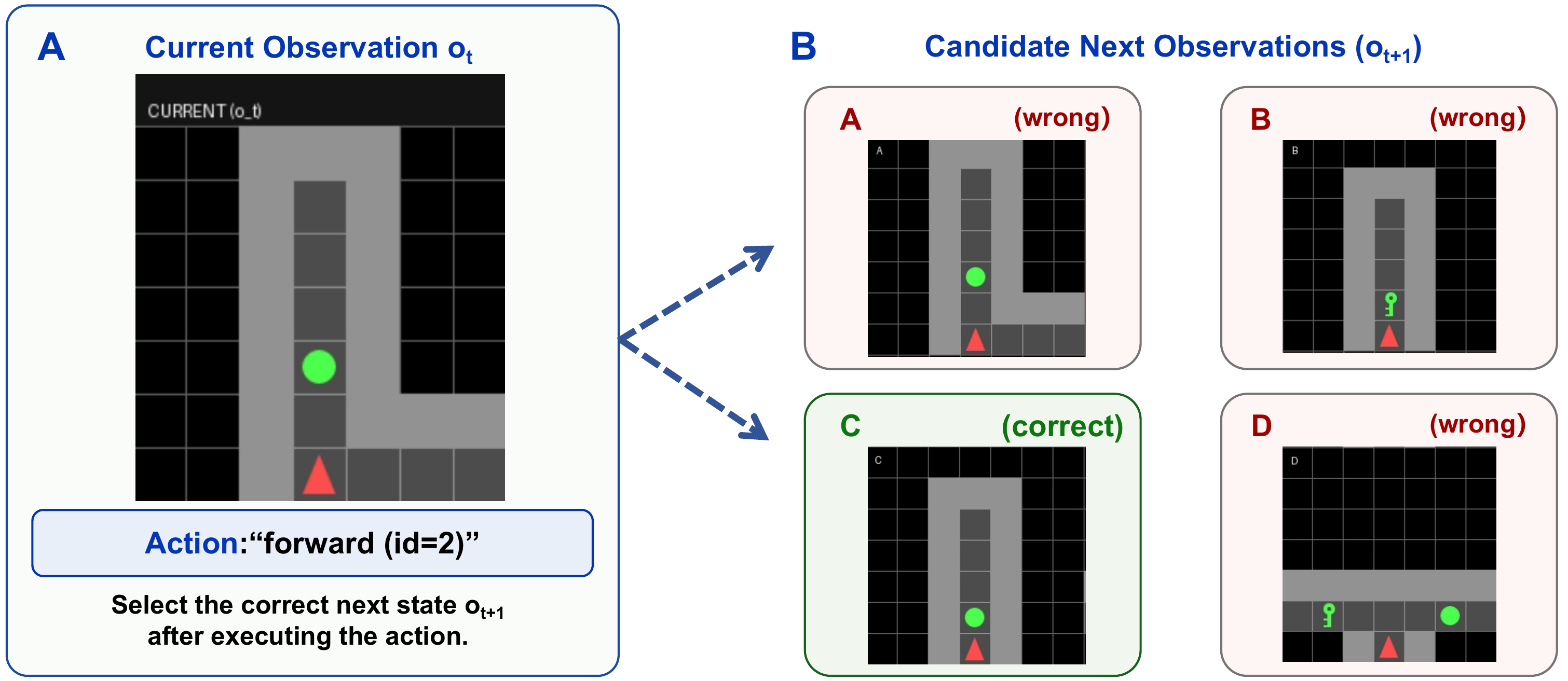}
    }{
        \fbox{
        \begin{minipage}[c][0.22\textheight][c]{0.9\linewidth}
            \centering
            \textbf{Placeholder for \texttt{figs/appC\_taskA\_layout.pdf}}\\[0.5em]
            Show a complete Task A item layout: (left) current observation $o_t$
            with the action text ``Action: right''; (right) four candidate next
            observations labeled A, B, C, D, with one marked as the correct
            transition. Use dashed lines to highlight the correct forward
            transition and to label two hard negative candidates
            (e.g., ``temporal neighbor'' and ``cross-trajectory negative'').
        \end{minipage}
        }
    }
    \caption{\textbf{Layout of a Task A multiple-choice item.} The current observation
    $o_t$ and the action text are shown on the left. Four candidate next
    observations (A--D) are rendered on the right. One candidate is the true
    next observation $o_{t+1}$; the others are visually similar hard negatives.
    Dashed annotations indicate which candidate is the correct transition and
    which negatives share the same trajectory or similar visual configurations.
    The gold answer is the letter of the correct candidate.
    Correct/wrong annotations are shown only for illustration and are not included in model-facing inputs.
    }
    \label{fig:appC_taskA_layout}
\end{figure}

Task A tests whether the VLM judge uses the action condition to identify the
correct next observation. Unlike Tasks C and D, Task A is a local transition
interface rather than a trajectory-level success/failure judgment.

\paragraph{Input and prompt.}
The item shows a current observation $o_t$, an action condition, and four
candidate next observations labeled A--D. The candidates are rendered from
MiniGrid states and are arranged as a multiple-choice panel. One candidate is the
true next observation $o_{t+1}$ under the action; the other candidates are
visually similar negatives. Figure~\ref{fig:appC_taskA_layout} shows the multiple-choice layout used for Task A;
the correct/wrong annotations are included only for illustration and are never
shown to the model.

\begin{quote}
\small
\textbf{Prompt template.}
Action: \texttt{\{action\_string\}}.\par
On the left is the current state. On the right are four candidate next states
(A, B, C, D). Select the correct next state after executing the action. Answer
with ONLY one letter: A, B, C, or D.
\end{quote}

\paragraph{Candidate construction.}
Negative candidates are selected to avoid trivial recognition. Same-trajectory
negatives are preferred, because they share the same environment, objects, and
visual style but correspond to different timesteps. If needed, cross-trajectory
negatives with similar local configurations are used as fallback. Candidate order
is randomized with the gold candidate uniformly placed among A--D.

\paragraph{Action ablations.}
Task A supports three action conditions. In the baseline condition, the true
action is shown. In the shuffle condition, the action field is replaced by a
mismatched action sampled from the action label space. In the mask condition, the
action field is replaced with an unknown-action token. The ablated interfaces
should not expose an action identifier that trivially recovers the original
action. 
Large accuracy drops indicate sensitivity to the
supplied action condition. Small or mixed differences
are compatible with action-insensitive behavior but
do not establish it alone.

\paragraph{Ablated action templates.}
The baseline condition displays the true action as
\texttt{Action: \{action\_string\}}. In the shuffle condition, the same field is
replaced by a different action sampled from the MiniGrid action set:
\texttt{Action: \{shuffled\_action\_string\}}. In the mask condition, the action
field is replaced by \texttt{Action: unknown}. The image candidates and answer
format remain unchanged across these conditions.

\subsection{Task D: Causal Sensitivity Tests}
\label{app:tasks:taskd}

Task D is the primary interface for causal sensitivity. It compares matched
successful \textsf{Full} trajectories against \textsf{CF} trajectories generated
by a minimal simulator intervention and rerollout under the unchanged action
sequence. The intended question is whether the judge changes its verdict when the
underlying task outcome changes.

\paragraph{Input and prompt.}
The dedicated reference-aided Task-D protocol contains (1) a successful Reference rollout that illustrates the intended task objective and (2) a separate Query rollout whose outcome must be judged. The Query may correspond to a Full, NoCue, or CF variant; Full/CF are the primary
paired conditions for discrimination. The judge must infer the specific objective from the Reference and return Success or Fail for the Query; audit-only gold labels, rewards, simulator success flags, and intervention metadata are excluded from the input.

\begin{quote}
\small
\textbf{Prompt template.}
Below are two storyboard trajectories from the same MiniGrid environment.
Reference trajectory:
This trajectory is a known successful example of completing the task in this environment.
Query trajectory:
Determine whether this trajectory also completes the same task goal as the reference trajectory.
Answer with ONLY: Success or Fail.
\end{quote}

\paragraph{Gold answer and scoring.}
The gold answer is \texttt{Success} for the matched successful \textsf{Full}
trajectory and \texttt{Fail} for the corresponding \textsf{CF} trajectory. The
main C4 metrics are computed over paired \textsf{Full}--\textsf{CF} groups and
are defined in Appendix~\ref{app:definitions}. If a \textsf{NoCue} version is
included in the Task D artifact, it should be treated as an evidence-removal
diagnostic rather than as the causal-intervention pair used for C4.

\begin{figure}[b]
    \centering
    \IfFileExists{figs/appC_taskD_paired_example.pdf}{
        \includegraphics[width=0.9\linewidth]{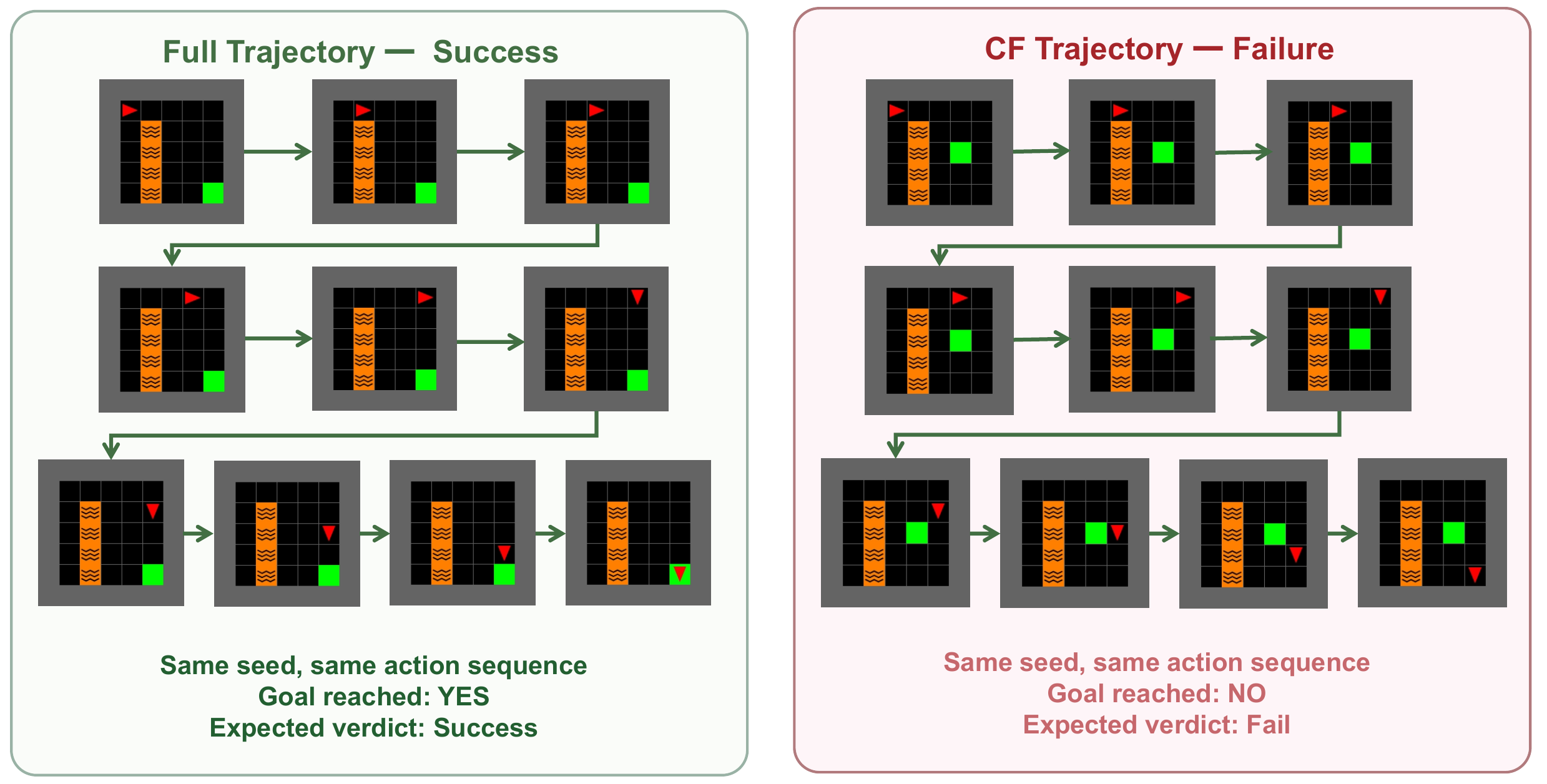}
    }{
        \fbox{
        \begin{minipage}[c][0.22\textheight][c]{0.9\linewidth}
            \centering
            \textbf{Placeholder for \texttt{figs/appC\_taskD\_paired\_example.pdf}}\\[0.5em]
            Show a DoorKey Task D paired case: (left) the \textsf{Full} trajectory
            montage with key pickup, door unlock, and goal reached; label the
            gold verdict ``Success''. (right) The matched \textsf{CF} montage
            showing the same action sequence and visual format, but with the
            goal relocated by a minimal intervention; label the gold verdict
            ``Fail''. Annotation: ``same seed, same action sequence,
            different outcome''.
        \end{minipage}
        }
    }
    \caption{\textbf{Illustration of a matched \textsf{Full}/\textsf{CF} pair for Task D
    (DoorKey example).} Both panels show the same action sequence in the same
    visual format. In the \textsf{Full} trajectory (left), the goal is reachable
    and the agent completes the task; the gold verdict is \texttt{Success}. In
    the matched \textsf{CF} trajectory (right), a minimal intervention relocates
    the goal before the episode starts; the same action sequence now fails. The
    gold verdict is \texttt{Fail}. A reliable judge should accept the
    \textsf{Full} and reject the \textsf{CF}.}
    \label{fig:appC_taskD_paired_example}
\end{figure}

\paragraph{Representative paired case.}
Figure~\ref{fig:appC_taskD_paired_example} illustrates a representative DoorKey Full/CF pair. The pair keeps the same action sequence. In the \textsf{Full}
trajectory, the agent picks up the key, unlocks the door, and reaches the goal.
In the matched \textsf{CF} trajectory, the initial intervention relocates or
removes the goal so that the same action sequence no longer satisfies the task.
A reliable judge should accept the \textsf{Full} montage and reject the matched
\textsf{CF} montage, while not changing its verdict merely because the same
montage is shown in a different but semantically equivalent representation.

\section{Supplementary Results and Analysis}
\label{app:results}

This appendix provides detailed MiniGrid evaluations
supporting the SCI diagnostics in the main paper.
We report model-level and environment-level results,
reference-aided causal discrimination, presentation
and representation sensitivity, perception and
output-compliance controls, and matched prompt ablations.
Expanded evaluations further examine the stability of
these findings with larger trajectory sets and
clustered uncertainty estimates. Throughout, Task C
and Task D retain their distinct judging protocols,
and statistical comparisons follow the eligibility
and rules defined in Appendix~\ref{app:definitions}.
Qualitative cases and additional model-family results
complement the quantitative analysis.

\subsection{Main MiniGrid Diagnostics}
\label{app:results:aggregated}
The main Task-C results are summarized in
Table~\ref{tab:main_diagnostic}. We provide
additional diagnostics of representation consistency
and reference-aided causal discrimination below,
followed by environment-level and controlled
experimental analyses.

\paragraph{Representation consistency.}
\label{app:results:rep}
The main paper reports that composite and split-image evidence formats can change judge behavior even when the underlying trajectory semantics are fixed. Table~\ref{tab:vcc_metrics} provides a compact representation-consistency summary using $\mathrm{VCC}$. Higher values indicate that the judge gives the same verdict across equivalent evidence formats; 1.0 would indicate perfect representation exchangeability.

\begin{table}[b]
    \caption{\textbf{Task-C representation consistency.} $VCC_{\mathrm{rep},C}^{\mathrm{pooled}}$ measures verdict agreement between matched Composite and Split inputs, pooled across Full and CF trajectory conditions. NoCue is excluded from this pooled comparison.}
    \label{tab:vcc_metrics}
    \centering
    \footnotesize
    \setlength{\tabcolsep}{4pt}
    \renewcommand{\arraystretch}{1.08}
    \begin{tabular}{l c}
        \toprule
        Model & $VCC_{\mathrm{rep},C}^{\mathrm{pooled}}$ \\
        \midrule
        Claude Sonnet 4.6      & 0.69  \\
        Gemini 2.0 Flash       & 0.66  \\
        Gemini 2.5 Flash Lite  & 0.66  \\
        Gemini 3 Flash & 0.73 \\
        GPT-4o                 & 0.71 \\
        GPT-5.4                & 0.66 \\
        Kimi K2.5              & 0.76 \\
        Qwen3-8B               & 0.66 \\
        Qwen3-32B              & 0.66 \\
        Qwen3-235B        & 0.66 \\
        \midrule
        Mean                   & 0.68 \\
        \bottomrule
    \end{tabular}
\end{table}

The mean $\mathrm{VCC}$ values are around 0.68--0.70, indicating that
representation consistency is imperfect across the model suite. These results
support the need to report representation exchangeability separately from both
plain accuracy and causal-sensitivity metrics.

\paragraph{Reference-Aided Task-D Results.}
\label{app:taskd_repair}
\label{app:results:causal}
The dedicated Task-D evaluation provides a successful Reference and a Composite-format Query, which the judge labels \texttt{Success} or \texttt{Fail}.  
Table~
\ref{tab:appD_taskd_strict} reports paired causal
discrimination on 12 complete Full/CF Query groups
per model, using strict verdict parsing and an
all-group denominator. Invalid or truncated outputs
are counted as incorrect. The resulting activation
and CF-rejection profiles differ substantially:
GPT-5.4 obtains $p_{\mathrm{active}}=0.750$ and
$\mathrm{IDR}_{\mathrm{raw}}=0.333$, whereas
Claude Sonnet 4.6 obtains $0.250$ and $0.250$,
respectively. These estimates also motivate the
larger-scale analyses reported below.

\begin{table}[t]
\centering
\caption{\textbf{Reference-aided Task-D discrimination (strict/all-groups).}
Strict all-group estimates over 12 matched Full/CF Query groups per model. Confidence intervals are reported for raw IDR using group-level and environment-level bootstrap resampling. }
\label{tab:appD_taskd_strict}
\footnotesize\setlength{\tabcolsep}{3pt}
\begin{tabular}{lcccc}
\toprule
Model & $p_{\mathrm{active}}$ & $\mathrm{IDR}_{\mathrm{raw}}$ & Group 95\% CI & Env. 95\% CI \\
\midrule
Claude Sonnet 4.6 & 0.250 & 0.250 & [0.000,0.500] & [0.000,0.583] \\
Gemini 2.0 Flash & 0.667 & 0.583 & [0.333,0.833] & [0.250,0.917] \\
Gemini 2.5 Flash Lite & 0.750 & 0.167 & [0.000,0.417] & [0.000,0.333] \\
Gemini 3 Flash & 0.833 & 0.167 & [0.000,0.417] & [0.000,0.500] \\
GPT-4o & 0.750 & 0.500 & [0.250,0.750] & [0.167,0.833] \\
GPT-5.4 & 0.750 & 0.333 & [0.083,0.583] & [0.083,0.667] \\
Kimi K2.5 & 0.167 & 0.000 & [0.000,0.000] & [0.000,0.000] \\
Qwen3-235B & 1.000 & 0.000 & [0.000,0.000] & [0.000,0.000] \\
Qwen3-32B & 0.000 & 0.000 & [0.000,0.000] & [0.000,0.000] \\
Qwen3-8B & 1.000 & 0.000 & [0.000,0.000] & [0.000,0.000] \\
\bottomrule
\end{tabular}
\end{table}

\subsection{Environment-Level Breakdown}
\label{app:results:env}
\begin{figure}[t]
    \centering
    \IfFileExists{figs/appD_env_breakdown_heatmap.pdf}{%
        \includegraphics[width=0.9\linewidth]{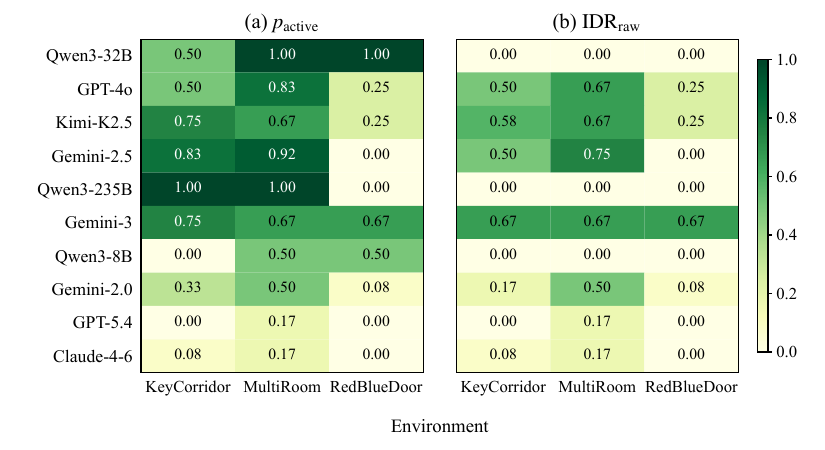}
    }{%
        \fbox{\begin{minipage}[c][0.30\textheight][c]{0.95\linewidth}
        \centering
        \textbf{Placeholder for \texttt{figs/appD\_env\_breakdown\_heatmap.pdf}}\\[0.6em]
        This figure should contain three heatmap panels for the Task~C
        environments shared by the presentation/representation probes:
        KeyCorridor, MultiRoom, and RedBlueDoor. Panel~(a) reports
        $\mathrm{JAcc}_{\mathrm{neu}}$, Panel~(b) reports
        $p_{\mathrm{active}}$, and Panel~(c) reports
        $\mathrm{IDR}_{raw}$. Rows are models and cell values are averaged
        over composite and split representations.
        \end{minipage}}
    }
\caption{\textbf{Environment-level MiniGrid breakdown for Task C.}
The heatmaps report (a) Full-trajectory activation
$p_{\mathrm{active}}$ and (b) raw paired Full--CF
discrimination $\mathrm{IDR}_{\mathrm{raw}}$.
Each environment contains six trajectory groups per
representation under neutral presentation.
Scores are averaged equally across Composite and Split,
using all groups and counting invalid verdicts as incorrect.
}

    \label{fig:appD_env_breakdown}
\end{figure}

The aggregate results above may hide environment-specific failure patterns.
Figure~\ref{fig:appD_env_breakdown} therefore reports an environment-level
breakdown for the Task~C MiniGrid settings. We restrict this plot to the three
environments that share the same Task~C presentation and representation probe
structure---KeyCorridor, MultiRoom, and RedBlueDoor---so that the environment
comparison is not confounded by task-specific coverage differences. Scores are
averaged over composite and split representations.

\subsection{Task-B Output Compliance and Failure-Source Accounting}
\label{app:taskb_errors}
Task B reports $\mathrm{Acc@0.90}$, the all-row fraction whose normalized,
weighted scene-component score reaches at least $0.90$; it is not JSON
exact-match accuracy. The submitted ten-model run contains 1,440 responses:
897 strict JSON outputs, 269 deterministically recoverable outputs,
121 other unparsable outputs, 72 explicit truncations, 81 API errors,
and no missing responses. Deterministic recovery is limited to JSON
extraction, syntax normalization, and canonical field names.
Table~\ref{tab:appD_taskb_compliance} reports
response validity and all-row performance, while
Table~\ref{tab:appD_taskb_components} breaks down
component-level recovery on parseable responses.
Together, they distinguish output compliance from
partial scene understanding.

\begin{table}[t]
\centering
\caption{\textbf{Task-B output compliance and partial semantic recovery.}
Every model has 144 submitted rows. Mean-all includes infrastructure and
format failures as zero; mean-parsed conditions on deterministically parsed
responses. The all-row Acc@0.90 values correspond to the
Task-B results reported in Table~\ref{tab:local_controls}.}
\label{tab:appD_taskb_compliance}
\scriptsize\setlength{\tabcolsep}{3pt}
\begin{tabular}{lrrrrrr}
\toprule
Model & Parsed $n$ & API & Trunc. & Other unparse. & Acc@0.90 & Mean-all / parsed \\
\midrule
Claude Sonnet 4.6 & 141 & 0 & 0 & 3 & 0.000 & 0.186 / 0.190 \\
Gemini 2.0 Flash & 63 & 81 & 0 & 0 & 0.007 & 0.099 / 0.226 \\
Gemini 2.5 Flash Lite & 140 & 0 & 0 & 4 & 0.000 & 0.157 / 0.161 \\
Gemini 3 Flash & 30 & 0 & 0 & 114 & 0.090 & 0.168 / 0.807 \\
GPT-4o & 144 & 0 & 0 & 0 & 0.000 & 0.223 / 0.223 \\
GPT-5.4 & 144 & 0 & 0 & 0 & 0.111 & 0.616 / 0.616 \\
Kimi K2.5 & 72 & 0 & 72 & 0 & 0.007 & 0.210 / 0.420 \\
Qwen3-235B & 144 & 0 & 0 & 0 & 0.000 & 0.400 / 0.400 \\
Qwen3-32B & 144 & 0 & 0 & 0 & 0.000 & 0.119 / 0.119 \\
Qwen3-8B & 144 & 0 & 0 & 0 & 0.000 & 0.119 / 0.119 \\
\bottomrule
\end{tabular}
\end{table}

\begin{table}[t]
\centering
\caption{\textbf{Task-B component scores on parsed rows only.}
The parsed support varies considerably across models and must not be
interpreted as a uniform full-sample estimate.}
\label{tab:appD_taskb_components}
\scriptsize\setlength{\tabcolsep}{3pt}
\begin{tabular}{lrrrrr}
\toprule
Model & $n_{parsed}$ & Position & Carrying & Front cell & Object Jaccard \\
\midrule
Claude Sonnet 4.6 & 141 & 0.064 & 0.709 & 0.035 & 0.182 \\
Gemini 2.0 Flash & 63 & 0.206 & 0.714 & 0.095 & 0.052 \\
Gemini 2.5 Flash Lite & 140 & 0.043 & 0.721 & 0.007 & 0.111 \\
Gemini 3 Flash & 30 & 1.000 & 0.867 & 0.533 & 0.817 \\
GPT-4o & 144 & 0.201 & 0.708 & 0.090 & 0.058 \\
GPT-5.4 & 144 & 0.903 & 0.847 & 0.444 & 0.228 \\
Kimi K2.5 & 72 & 0.569 & 0.694 & 0.208 & 0.254 \\
Qwen3-235B & 144 & 0.618 & 0.715 & 0.257 & 0.014 \\
Qwen3-32B & 144 & 0.000 & 0.715 & 0.000 & 0.000 \\
Qwen3-8B & 144 & 0.000 & 0.715 & 0.000 & 0.000 \\
\bottomrule
\end{tabular}
\end{table}

\paragraph{Separate generation-fix check.}
On the same 114 Gemini 3 Flash Task-B UIDs, a separately labeled
output-generation/routing fix improved parsing from $0.208$ to $0.993$,
the all-row mean component score from $0.168$ to $0.788$, and
Acc@0.90 from $0.090$ to $0.451$. This additional run diagnoses an output
pipeline limitation and does not silently replace the originally submitted
scores. High-threshold failures therefore cannot be equated with total
absence of partial spatial or object understanding.

\subsection{Perception-Controlled Oracle-State Hierarchy}
\label{app:oracle_hierarchy}
The Oracle-State control evaluates GPT-5.4 and Gemini 3.1 Pro on
98 shared groups in nine environments. Each group includes a successful
Reference from which the judge must recover the task type and objective
before classifying a separate Query. The \emph{Pixel}
condition supplies rendered frames; \emph{State} adds selected observable
trajectory facts; and \emph{Pixel+State} provides both. The state allowlist
excludes the exact objective, gold verdict, reward, success, and termination fields. These are differences in available evidence, not merely alternative layouts of identical pixels. 

\begin{table}[!htbp]
\centering
\caption{\textbf{Oracle-State diagnostic hierarchy (98 shared groups).}
Task, objective and verdict are the individually scored stages; joint
requires their simultaneous success. These values correspond to the same
matched three-arm experiment.}
\label{tab:appD_oracle_stages}
\footnotesize\setlength{\tabcolsep}{4pt}
\begin{tabular}{llrrrr}
\toprule
Model & Evidence & Task & Objective & Verdict & Joint \\
\midrule
GPT-5.4 & Pixel & 0.442 & 0.405 & 0.731 & 0.320 \\
GPT-5.4 & State & 0.905 & 0.905 & 0.895 & 0.820 \\
GPT-5.4 & Pixel+State & 0.854 & 0.874 & 0.881 & 0.765 \\
Gemini 3.1 Pro & Pixel & 0.704 & 0.395 & 0.755 & 0.347 \\
Gemini 3.1 Pro & State & 0.949 & 0.912 & 0.973 & 0.895 \\
Gemini 3.1 Pro & Pixel+State & 0.983 & 0.929 & 0.997 & 0.929 \\
\bottomrule
\end{tabular}
\end{table}

\begin{table}[!htbp]
\centering
\caption{\textbf{Oracle-State paired Full/CF diagnostics.}
95\% intervals apply to raw IDR and are based on complete matched groups.}
\label{tab:appD_oracle_idr}
\footnotesize\setlength{\tabcolsep}{4pt}
\begin{tabular}{llrrrr}
\toprule
Model & Evidence & $p_{active}$ & $\mathrm{IDR}_{raw}$ & Group 95\% CI & $\mathrm{IDR}_{cond}$ \\
\midrule
GPT-5.4 & Pixel & 0.745 & 0.469 & [0.367,0.571] & 0.630 \\
GPT-5.4 & State & 0.929 & 0.765 & [0.684,0.847] & 0.824 \\
GPT-5.4 & Pixel+State & 0.918 & 0.765 & [0.684,0.847] & 0.833 \\
Gemini 3.1 Pro & Pixel & 0.612 & 0.602 & [0.500,0.694] & 0.983 \\
Gemini 3.1 Pro & State & 1.000 & 0.929 & [0.878,0.980] & 0.929 \\
Gemini 3.1 Pro & Pixel+State & 1.000 & 0.990 & [0.969,1.000] & 0.990 \\
\bottomrule
\end{tabular}
\end{table}

For GPT-5.4, State increases joint task/objective/verdict accuracy from
$0.320$ to $0.820$ and raw IDR from $0.469$ to $0.765$; for Gemini,
those values increase from $0.347$ to $0.895$ and from $0.602$ to $0.929$.
The group interval for GPT's gain excludes zero, but the more conservative
environment-level interval includes zero. These contrasts implicate
visual/interface grounding and objective recovery as major, model-dependent
bottlenecks without establishing that every Pixel error originates in
causal reasoning. 
Tables~\ref{tab:appD_oracle_stages}
and~\ref{tab:appD_oracle_idr} report the
corresponding stage-wise and paired metrics.

\subsection{Matched Presentation and Representation Probes}
\label{app:probe_diagnostics}
\label{app:interface_analysis}
Presentation changes preserve simulator state, actions, and outcome labels
but may change perceptual accessibility. The submitted ten-endpoint tests
therefore report both correctness degradation and correctness rescue on
exact matched items. Holm correction is applied within the 180-cell
family (60 per probe type).

\begin{table}[!htbp]
\centering
\caption{\textbf{Historical matched presentation effects.}
Counts aggregate exactly matched comparisons across the original model
suite. None of the 180 cells survives the stated global Holm correction.}
\label{tab:appD_matched_probes}
\footnotesize\setlength{\tabcolsep}{4pt}
\begin{tabular}{lrrrrr}
\toprule
Probe & Exact pairs & Degradation & Rescue & $\Delta$Acc. & Correctness flip \\
\midrule
Noise & 1080 & 59 & 75 & +0.015 & 0.124 \\
Style & 1080 & 46 & 55 & +0.008 & 0.094 \\
Temporal reversal & 1080 & 136 & 92 & -0.041 & 0.211 \\
\bottomrule
\end{tabular}
\end{table}

\begin{table}[!htbp]
\centering
\caption{\textbf{Outcome-stratified historical Composite/Split transitions.}
Flips are changes in predicted verdict, whereas gain/loss counts refer to
correctness and may cancel in the net mean. Historical interfaces also
varied in prompt wording, packaging, and run date; these contrasts do not
isolate image resolution or tiling. The CF mean effect is exploratory and
unadjusted.}
\label{tab:appD_comp_split_transitions}
\footnotesize\setlength{\tabcolsep}{3pt}
\begin{tabular}{lrrrrr}
\toprule
Variant & Pairs & Verdict flip & Gain / Loss & $\Delta$Accuracy & 95\% CI \\
\midrule
Full & 1620 & 35.7\% & 277 / 274 & +0.002 & [-0.173,0.167] \\
NoCue & 1620 & 33.8\% & 235 / 293 & -0.036 & [-0.198,0.122] \\
CF & 1620 & 25.6\% & 328 / 84 & +0.151 & [0.028,0.272] \\
\bottomrule
\end{tabular}
\end{table}

In the original 270 finer Composite/Split comparison cells, only 10
survived the stated global Holm correction. In a separate Scale-2 probe
check comprising 101 groups, ten environment configurations and 12
Full/CF cells, no comparison survived that experiment's own global Holm
correction. These multiple-testing families are distinct and should not
be merged post hoc. 
Tables~\ref{tab:appD_matched_probes}
and~\ref{tab:appD_comp_split_transitions}
summarize the matched presentation and
representation effects.

\subsection{Matched Prompt-Structure Ablation}
\label{app:prompt_ablations}
A further Task-D study evaluates identical groups, images, and ordering
under three prompting conditions: Direct, Structured Verdict, and
Structured Rationale. The results demonstrate changes in response bias,
not a monotonic benefit from requesting more reasoning (Table~\ref{tab:appD_prompt_ablations}).

\begin{table}[!htbp]
\centering
\caption{\textbf{Paired Task-D prompt ablation.} The three arms use matched
Task-D groups; the reported intervals apply to raw IDR.}
\label{tab:appD_prompt_ablations}
\footnotesize\setlength{\tabcolsep}{3pt}
\begin{tabular}{llrrrr}
\toprule
Model & Prompt & Full accept & CF reject & $\mathrm{IDR}_{raw}$ [95\% CI] & $\mathrm{IDR}_{cond}$ \\
\midrule
GPT-5.4 & Direct & 0.827 & 0.541 & 0.367 [0.276,0.459] & 0.444 \\
GPT-5.4 & Structured verdict & 0.704 & 0.520 & 0.265 [0.184,0.357] & 0.377 \\
GPT-5.4 & Structured rationale & 0.449 & 0.878 & 0.398 [0.306,0.490] & 0.886 \\
Claude Sonnet 4.6 & Direct & 0.255 & 0.939 & 0.224 [0.143,0.306] & 0.880 \\
Claude Sonnet 4.6 & Structured verdict & 0.694 & 0.816 & 0.541 [0.439,0.643] & 0.779 \\
Claude Sonnet 4.6 & Structured rationale & 0.520 & 0.796 & 0.367 [0.276,0.459] & 0.706 \\
\bottomrule
\end{tabular}
\end{table}
For GPT-5.4, the Rationale--Direct raw IDR difference is $+0.031$,
with a paired 95\% CI from $-0.102$ to $+0.163$.
For Claude, the Rationale--Verdict difference is $-0.173$,
with a CI from $-0.286$ to $-0.071$. 

\subsection{Expanded Scale-2 Evaluation and Clustered Uncertainty}
\label{app:scale2}
The expanded scale-2 evaluations provide substantially larger
task-specific samples without pooling them with the
base MiniGrid results. Table~\ref{tab:appD_scale2_size}
summarizes the evaluation scale and environment
coverage. Task C and Task D use distinct judging
contracts, and model coverage varies by task.
Table~\ref{tab:appD_scale2} reports the corresponding
paired causal diagnostics and uncertainty estimates.

\begin{table}[htbp!]
\centering
\caption{\textbf{Expanded Evaluation scale and coverage.}
The expanded evaluation increases task-specific
sample coverage while preserving the distinct
Task-C and Task-D protocols.
Counts refer to exam rows and complete source
groups, not independent model evaluations.}
\label{tab:appD_scale2_size}
\footnotesize
\setlength{\tabcolsep}{6pt}
\renewcommand{\arraystretch}{1.08}
\begin{tabular}{lrrrrr}
\toprule
Task & Base rows & Expanded rows & Ratio
     & Source groups & Env. configs \\
\midrule
A & 103 & 1,030 & $10.0\times$ & 76 & 9 \\
C & 486 & 1,326 & $2.7\times$ & 442 & 10 \\
D & 144 & 1,296 & $9.0\times$ & 432 & 9 \\
\bottomrule
\end{tabular}
\end{table}

\begin{table}[htbp!]
\centering
\caption{\textbf{Scale-2 paired diagnostics.} C and D use different judging
contracts.  
Only evaluated model-task combinations are shown.
Confidence intervals apply to raw IDR, with intact
trajectory-group and environment-configuration
resampling reported separately.
}
\label{tab:appD_scale2}
\scriptsize\setlength{\tabcolsep}{2.8pt}
\begin{tabular}{llrccccc}
\toprule
Task & Model & Groups & $p_{\mathrm{active}}$ & CF rej. & $\mathrm{IDR}_{raw}$ & Group 95\% CI & Env. 95\% CI \\
\midrule
C & GPT-5.4 & 442 & 0.523 & 0.629 & 0.314 & [0.271,0.357] & [0.249,0.372] \\
C & Claude Sonnet 4.6 & 442 & 0.043 & 0.993 & 0.043 & [0.025,0.063] & [0.009,0.107] \\
\midrule
D & GPT-5.4 & 432 & 0.720 & 0.530 & 0.322 & [0.278,0.366] & [0.226,0.461] \\
D & Claude Sonnet 4.6 & 432 & 0.113 & 0.933 & 0.088 & [0.062,0.116] & [0.037,0.189] \\
D & Gemini 3 Flash & 432 & 0.972 & 0.637 & 0.613 & [0.567,0.660] & [0.335,0.767] \\
D & Qwen3-32B & 432 & 1.000 & 0.000 & 0.000 & [0.000,0.000] & [0.000,0.000] \\
\bottomrule
\end{tabular}
\end{table}

All reported confidence intervals use 20,000 percentile-bootstrap resamples. The primary intervals resample intact trajectory groups, preserving their Full/NoCue/CF dependence and matched models; the sensitivity analysis resamples complete environment configurations. Invalid and unparsable outputs count as incorrect in these all-group summaries.

For Task C, the conditional IDR estimates were $0.602$ for GPT-5.4 and
$1.000$ for Claude Sonnet 4.6; for Task D, they were $0.447$ and $0.776$
for the same two models, $0.631$ for Gemini 3 Flash, and $0.000$ for the
Qwen3-32B. On Task-D groups, the
paired GPT-5.4 minus Claude Sonnet 4.6 difference in raw IDR was $+0.234$,
with group-bootstrap 95\% CI $[+0.185,+0.282]$ and environment-bootstrap
95\% CI $[+0.127,+0.358]$. The larger groups sharpen the uncertainty
analysis, but the point estimates do not all remain numerically constant:
Claude's original Task-D IDR is $0.250$, versus $0.088$ on Scale-2.
The supported conclusion is persistence of heterogeneous activation,
CF rejection, and paired discrimination rather than invariant model rankings.

\subsection{Qualitative and Exploratory Results}
\label{app:results:failure_cases}

\begin{figure}[t]
    \centering
    \IfFileExists{figs/appD_failure_cases.pdf}{%
        \includegraphics[width=\linewidth]{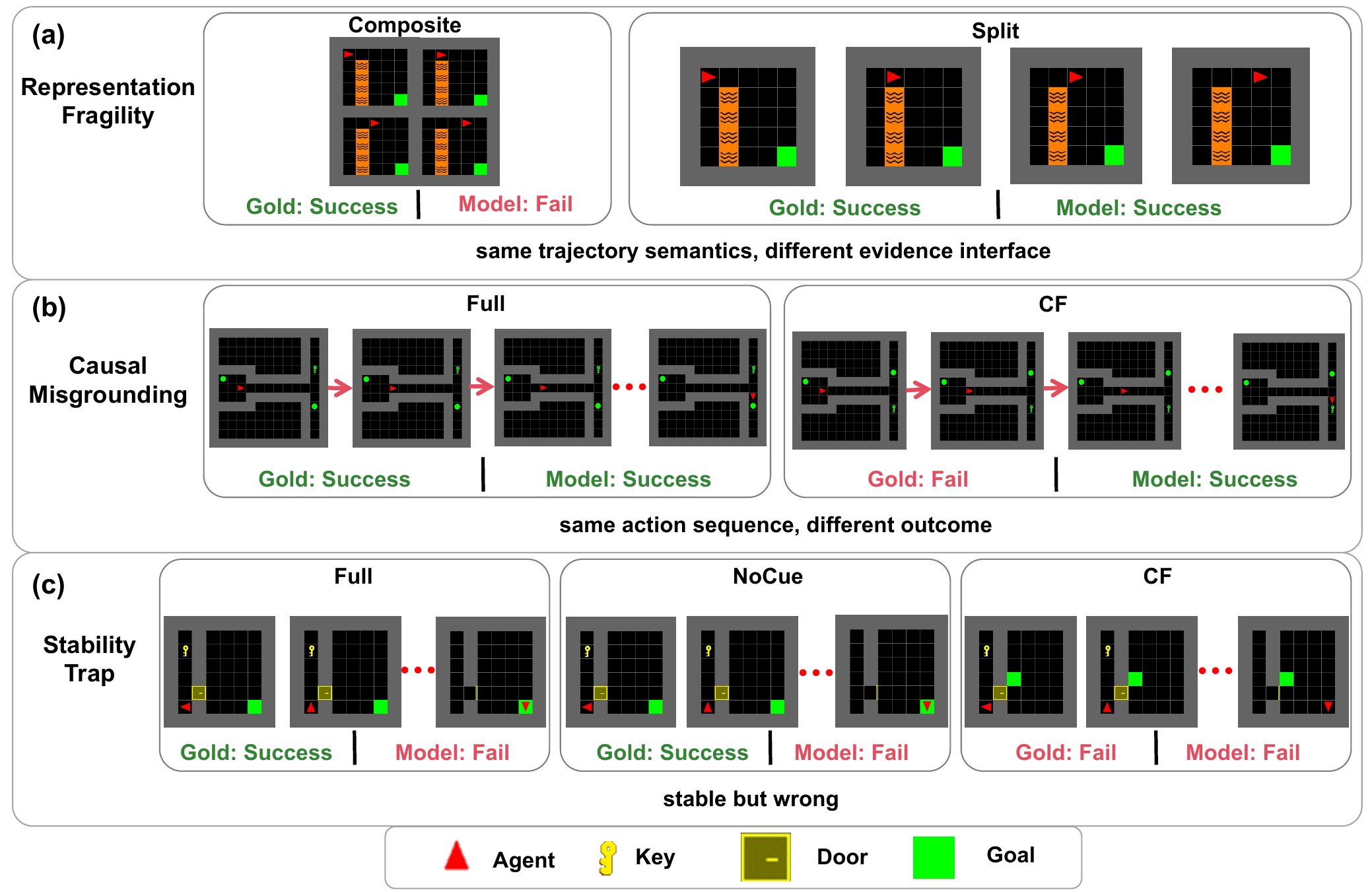}
    }{%
        \fbox{\begin{minipage}[c][0.34\textheight][c]{0.95\linewidth}
        \centering
        \textbf{Placeholder for \texttt{figs/appD\_failure\_cases.pdf}}\\[0.6em]
        Recommended layout: three rows. Row~1: representation fragility, showing
        the same trajectory in composite and split formats with different model
        verdicts. Row~2: causal misgrounding, showing a matched
        \textsf{Full}/\textsf{CF} pair where the model fails to change verdict.
        Row~3: stability trap, showing stable but incorrect all-success or
        all-fail predictions across variants or presentation probes. Each row
        should include the gold label and the model verdict.
        \end{minipage}}
    }
    \caption{\textbf{Representative MiniGrid failure cases.} Each row shows the model-facing evidence, simulator-derived gold label, and model verdict before and after the diagnostic change. Row 1 illustrates representation fragility, where the same Full trajectory is judged differently under composite and split representations. Row 2 illustrates causal misgrounding, where a matched Full/CF pair receives an incorrect or unchanged verdict. Row 3 illustrates a stability trap, where the model remains stable across evidence conditions but is wrong or degenerate. Gold labels are derived from simulator state and are not shown to the model.}
    \label{fig:appD_failure_cases}
\end{figure}

To complement the aggregate metrics, Figure~\ref{fig:appD_failure_cases}
shows representative qualitative failures selected from the final frozen
MiniGrid manifest. The purpose is not to introduce new metrics, but to make the
failure-mode taxonomy in Appendix~\ref{app:definitions} concrete.

\paragraph{Exploratory LLaVA-Family Results.}
\label{app:results:llava}
For completeness, we also evaluated LLaVA-family multimodal models under the
same MiniGrid diagnostic protocol. These runs are exploratory and are not used
for the main claims, because they extend the model suite beyond the primary
judge set discussed in the main text. Table~\ref{tab:llava_results} summarizes
the compact results.

\begin{table}[htbp!]
    \caption{\textbf{Exploratory LLaVA-family model performance on MiniGrid Tier-0.} The metrics are averaged over composite and split representations. These results are reported as auxiliary evidence rather than as part of the main model     suite.}
    \label{tab:llava_results}
    \centering
    \footnotesize
    \setlength{\tabcolsep}{4pt}
    \renewcommand{\arraystretch}{1.08}
    \begin{tabular}{l c c c c}
        \toprule
        Model & Task C Acc. & Task A Acc. & Task E Acc. & Strict Rate \\
        \midrule
        LLaVA-Critic-R1-7B & 0.33 & 0.28 & 0.11 & 0.27 \\
        LLaVA-Critic-7B & 0.65 & 0.25 & 0.00 & 0.00 \\
        LLaVA-OneVision-7B & 0.63 & 0.27 & 0.00 & 0.00 \\
        LLaVA-OneVision-Chat & 0.63 & 0.24 & 0.00 & 0.00 \\
        \midrule
        Mean & 0.56 & 0.26 & 0.03 & -- \\
        \bottomrule
    \end{tabular}
\end{table}

The exploratory LLaVA results are substantially below the main judge suite on
both correctness and intervention-discrimination metrics. The main failure
surface appears to combine visual-interface parsing difficulty with weak
causal-discrimination ability. We therefore report these results only as a
reference point for open VLM-family models, not as evidence for the primary SCI
claims.

\subsection{Summary}
\label{app:results:discussion}
The supplementary MiniGrid analyses reinforce three
complementary observations. First, judge accuracy,
verdict stability, and paired causal discrimination
capture distinct aspects of reliability, as shown by
the reference-aided and expanded evaluations.
Second, presentation and representation probes reveal
model-dependent interface sensitivity, while prompt
ablations expose substantial changes in activation
and rejection behavior. Third, perception controls,
output-compliance analyses, and Oracle-State
comparisons identify additional bottlenecks that
cannot be attributed uniquely to causal reasoning.
Together, these results support the joint use of
correctness, invariance, and causal-sensitivity
diagnostics.

\section{3D Extensions: MiniWorld and Habitat}
\label{app:3d_extensions}
We extend GridWM-Judge to two simulated 3D environments,
MiniWorld and Habitat-Sim/HSSD, to examine SCI diagnostics
under richer visual evidence. These evaluations cover
action-conditioned transitions, trajectory judgments,
presentation effects, and paired causal discrimination,
using simulator-grounded outcomes for evaluation.

\subsection{MiniWorld Environments and Diagnostic Tasks}
\begin{figure}[htbp!]
    \centering
    \IfFileExists{figs/appE_miniworld_interface_ssot.pdf}{
        \includegraphics[width=\linewidth]{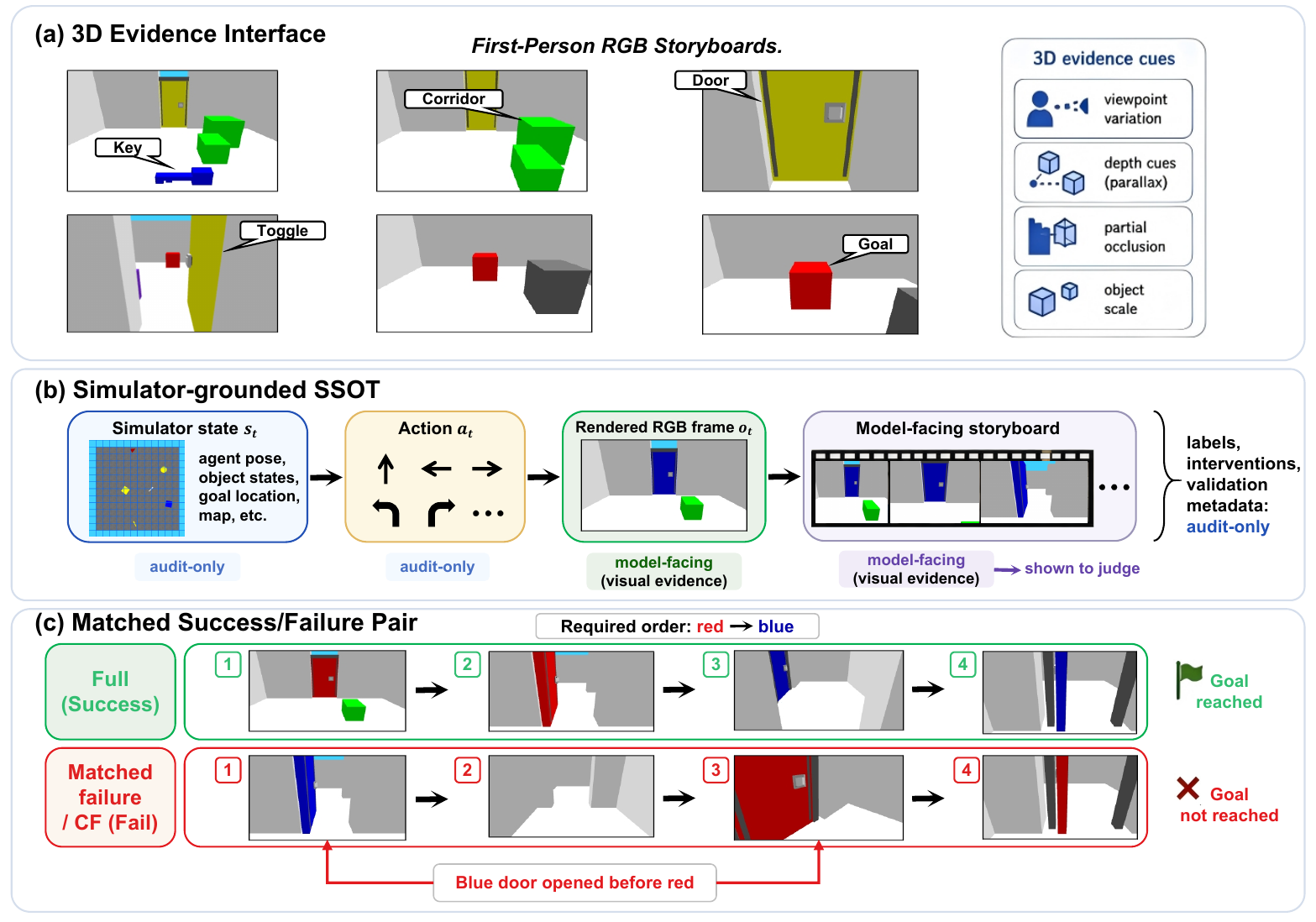}
    }{
        \fbox{
        \begin{minipage}[c][0.30\textheight][c]{0.96\linewidth}
            \centering
            \textbf{Placeholder for \texttt{figs/appE\_miniworld\_interface\_ssot.pdf}}\\[0.6em]
            Left: representative first-person MiniWorld frames from two or three tasks.\\[0.4em]
            Middle: simulator state/action trace aligned with rendered RGB evidence.\\[0.4em]
            Right: matched successful and failure trajectories with the intervention or failure point marked.
        \end{minipage}
        }
    }
    \caption{\textbf{MiniWorld evidence interface and simulator-grounded source of truth.}
    The figure shows representative first-person 3D frames, aligned simulator state/action traces, and a matched success/failure pair. Rendered frames are model-facing, while states, interventions, and labels remain audit-only metadata.}
    \label{fig:miniworld_interface_ssot}
\end{figure}

We include a small MiniWorld extension as a 3D external-validity probe for
GridWM-Judge. Unlike the Tier-0 MiniGrid suite, which is designed for tightly
audited symbolic-state control, MiniWorld renders trajectories from a
first-person 3D viewpoint. This introduces viewpoint variation, depth cues,
partial occlusion, object-scale variation, and navigation ambiguity while
preserving the same simulator-backed single-source-of-truth principle used
throughout the benchmark. Figure~\ref{fig:miniworld_interface_ssot} illustrates the MiniWorld evidence
interface, simulator metadata, and matched success/failure structure.

The MiniWorld extension examines whether SCI
diagnostics remain informative under first-person
3D visual evidence. We retain simulator-grounded
state traces, actions, observations, and outcome
labels, while exposing only task-relevant evidence
to the judge.
For each
MiniWorld rollout, we retain a simulator state trace, an action sequence,
rendered visual evidence, a mission or task description, and an outcome label.
The model-facing item contains only rendered evidence and task instructions;
states, intervention records, labels, and validation metadata remain audit-only. Table~\ref{tab:miniworld_task_summary} summarizes how the MiniWorld extension mirrors
the SCI diagnostic logic while remaining separate from the Tier-0 MiniGrid
results.

The MiniWorld extension focuses on three diagnostic questions: (i) whether the judge
can use action-conditioned 3D evidence, (ii) whether success/failure judgments change under prompt framing, and (iii) whether the judge distinguishes successful trajectories from matched failure cases. We report the resulting accuracies separately from the MiniGrid Tier-0 results.

\begin{table}[htbp!]
    \caption{\textbf{MiniWorld extension design.} The 3D extension mirrors the SCI
    diagnostic logic of MiniGrid but is reported separately because the visual
    interface and trajectory distribution differ from the Tier-0 2D suite.}
    \label{tab:miniworld_task_summary}
    \centering
    \footnotesize
    \setlength{\tabcolsep}{3pt}
    \renewcommand{\arraystretch}{1.08}
    \begin{tabularx}{\linewidth}{p{0.20\linewidth} p{0.22\linewidth} X p{0.18\linewidth}}
        \toprule
        Component & Diagnostic role & Model-facing evidence & Reported output \\
        \midrule
        MiniWorld suite
        & 3D external-validity probe
        & First-person RGB storyboards; simulator state traces retained as audit metadata
        & Env./group/item counts \\
        Task A-style
        & C2: action-conditioned \newline evidence use
        & Current 3D observation, action text, and candidate next observations or trajectory slices
        & $\mathrm{Acc}_{A}$ \\
        Task C-style
        & C1: presentation-sensitive trajectory judgment
        & Composite 3D storyboard with mission and action context
        & $\mathrm{Acc}_{C}$ \\
        Framing probe
        & C1: prompt-framing \newline sensitivity
        & Alternative framing prompt applied to the same 3D trajectory evidence
        & $\Delta_C^{\mathrm{frame}}$ \\
        Task D-style
        & C4: causal sensitivity
        & Successful trajectory evidence versus matched failure or counterfactual evidence
        & $\mathrm{Acc}_{D}$, $\Delta_D^{\mathrm{frame}}$ \\
        \bottomrule
    \end{tabularx}
\end{table}

\subsection{MiniWorld Results and Analysis}

\begin{table}[t]
    \caption{\textbf{MiniWorld diagnostic results.} Values are point estimates from the MiniWorld composite-SSOT result file. $\Delta^{\mathrm{frame}}$ denotes the change from the neutral prompt to the alternative framing probe; negative values indicate lower accuracy under framing. Because the current MiniWorld result file reports aggregate accuracies rather than paired Full--CF discrimination rates, $\mathrm{Acc}_{D}$ should be interpreted as a coarse Task-D-style failure-discrimination diagnostic rather than the full C4 IDR analysis used in MiniGrid. Support counts are provided in the released MiniWorld manifest.}
    \label{tab:miniworld_results}
    \centering
    \footnotesize
    \setlength{\tabcolsep}{3pt}
    \begin{tabular}{l c c c c c}
        \toprule
        Model
        & $\mathrm{Acc}_{A}$
        & $\mathrm{Acc}_{C}$
        & $\Delta_C^{\mathrm{frame}}$
        & $\mathrm{Acc}_{D}$
        & $\Delta_D^{\mathrm{frame}}$ \\
        \midrule
        Claude Sonnet 4.6      & 0.50 & 0.38 & $+0.12$ & 0.68 & $-0.13$ \\
        Gemini 2.0 Flash       & 0.40 & 0.43 & $-0.15$ & 0.20 & $+0.17$ \\
        Gemini 2.5 Flash Lite  & 0.48 & 0.35 & $+0.02$ & 0.50 & $-0.01$ \\
        Gemini 3 Flash & 0.55 & 0.81 & $-0.07$ & 0.78 & $+0.02$ \\
        GPT-4o                 & 0.55 & 0.75 & $-0.16$ & 0.50 & $+0.05$ \\
        GPT-5.4                & 0.50 & 0.51 & $+0.03$ & 0.66 & $-0.02$ \\
        Kimi K2.5              & 0.50 & 0.82 & $-0.14$ & 0.54 & $+0.01$ \\
        Qwen3-8B               & 0.49 & 0.67 & $-0.11$ & 0.50 & $+0.00$ \\
        Qwen3-32B              & 0.49 & 0.67 & $-0.11$ & 0.50 & $+0.00$ \\
        Qwen3-235B        & 0.51 & 0.55 & $-0.03$ & 0.50 & $+0.00$ \\
        \midrule
        Mean                   & 0.50 & 0.59 & $-0.06$ & 0.54 & $+0.01$ \\
        \bottomrule
    \end{tabular}
\end{table}

\begin{figure}[t]
    \centering
    \includegraphics[width=0.95\linewidth]{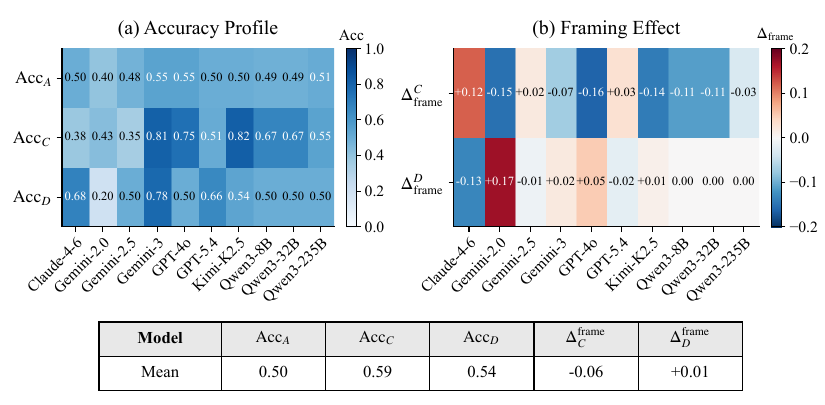}
    \caption{\textbf{MiniWorld diagnostic profile across models.}
    (a) \textbf{Accuracy Profile}: point-estimate accuracies for three
    MiniWorld diagnostic axes: action-conditioned evidence use
    ($\mathrm{Acc}_{A}$), trajectory-level success/failure judgment
    ($\mathrm{Acc}_{C}$), and Task-D-style failure discrimination
    ($\mathrm{Acc}_{D}$). Darker cells indicate higher accuracy, and the
    overlaid numbers give the corresponding point estimates.
    (b) \textbf{Framing Effect}: changes in accuracy under the finalized
    alternative framing prompt, computed relative to the corresponding neutral
    prompt as
    $\Delta_C^{\mathrm{frame}}=\mathrm{Acc}_C^{\mathrm{frame}}-\mathrm{Acc}_C$
    and
    $\Delta_D^{\mathrm{frame}}=\mathrm{Acc}_D^{\mathrm{frame}}-\mathrm{Acc}_D$.
    Negative values indicate lower accuracy under framing, while positive
    values indicate higher accuracy. The summary row reports the mean value of
    each metric across models. The figure highlights that MiniWorld model
    rankings differ across diagnostic axes and that framing sensitivity is not
    reducible to overall accuracy.}
    \label{fig:miniworld_result_profile}
\end{figure}

Table~\ref{tab:miniworld_results} reports MiniWorld results as a 3D
external-validity probe rather than as a second leaderboard. The table should be
read along three axes: action-conditioned evidence use
($\mathrm{Acc}_{A}$), trajectory-level success/failure judgment
($\mathrm{Acc}_{C}$), and Task-D-style failure discrimination
($\mathrm{Acc}_{D}$). The framing-delta columns report how much the same
evaluation changes under the finalized alternative framing prompt.

Figure~\ref{fig:miniworld_result_profile} visualizes the same MiniWorld results as a diagnostic profile, separating accuracy-style axes from framing effects across
models. The main pattern is that no single MiniWorld accuracy number explains judge
reliability. For example, \textsc{Gemini 3 Flash} is among the strongest
models on the trajectory-level and Task-D-style columns, with
$\mathrm{Acc}_{C}=0.81$ and $\mathrm{Acc}_{D}=0.78$. However, its
Task-C score changes under framing
($\Delta_C^{\mathrm{frame}}=-0.07$), showing that strong trajectory-level
accuracy does not by itself imply presentation invariance. Similarly,
\textsc{GPT-4o} obtains a high $\mathrm{Acc}_{C}=0.75$ but a lower
$\mathrm{Acc}_{D}=0.50$, indicating that success/failure judgment and matched
failure discrimination should be interpreted separately.

Model ranking also changes across diagnostic axes. \textsc{Kimi K2.5} obtains
the highest $\mathrm{Acc}_{C}$ in this MiniWorld evaluation ($0.82$), but its
Task-D-style score is lower than that of \textsc{Gemini 3 Flash}
($0.54$ vs.~$0.78$). Conversely, \textsc{claude-sonnet-4-6} performs better on
$\mathrm{Acc}_{D}$ ($0.68$) than on $\mathrm{Acc}_{C}$ ($0.38$). This separation is consistent with the SCI framing: presentation robustness,
trajectory-level correctness, and Task-D-style failure discrimination are distinct properties and should not be collapsed into a single aggregate score.

The mean MiniWorld scores remain close to the mid-range:
$\mathrm{Acc}_{A}=0.50$, $\mathrm{Acc}_{C}=0.59$, and
$\mathrm{Acc}_{D}=0.54$. This suggests that the 3D extension is not saturated by
current judges. In particular, several models remain near chance on Task-D-style
metrics, and \textsc{gemini-2.0-flash} obtains a low
$\mathrm{Acc}_{D}=0.20$. Because the current MiniWorld result file reports
aggregate accuracies rather than paired IDR-style discrimination rates, these
numbers should be interpreted as coarse diagnostic evidence rather than as the
full C4 analysis used in MiniGrid.

Compared with the 2D MiniGrid benchmark, the MiniWorld extension preserves the
same methodological logic but exposes a different failure surface. Both settings
use simulator-backed labels, aligned visual evidence, and controlled
presentation or causal variants. MiniWorld additionally introduces first-person
3D perception burdens, including camera pose, partial observability, occlusion,
depth cues, and object-scale variation. It is therefore best viewed as a stress
test of whether SCI-style diagnostics remain informative under a richer evidence
interface, not as a replacement for the fully audited Tier-0 MiniGrid suite.

\subsection{Habitat-Sim/HSSD Experimental Setup}
\label{app:hssd_setup}
We evaluate GridWM-Judge in Habitat-Sim using HSSD
synthetic articulated indoor scenes, with GPT-5.4 and
Gemini 3 Flash as the evaluated VLM judges.
Task A comprises 150 action-conditioned transition
items from 10 scenes per model.
Tasks C and D evaluate the same 100 underlying
Query groups across 18 scenes, with matched Full/CF
conditions. Task C uses trajectory-level judgments,
whereas Task D additionally provides a successful
Reference to establish the task objective.
Counterfactual failures are generated through
articulated action-effect interventions while
preserving the Query action sequence.

\subsection{Habitat Results}
\label{app:hssd_results}

\begin{table}[t]
\centering
\caption{\textbf{Habitat synthetic-3D evaluation.}
$N$ denotes transition items for Task A and matched
Query groups for Tasks C/D. For Tasks C/D, accuracy is the mean of
Full acceptance and CF rejection. Raw IDR is reported
with group-bootstrap 95\% confidence intervals.}
\label{tab:appE_hssd}

\footnotesize
\setlength{\tabcolsep}{2.5pt}
\renewcommand{\arraystretch}{1.08}

\begin{tabular}{llrrccccc}
\toprule
Model & Task & $N$ & Scenes & Acc. &
$p_{\mathrm{active}}$ & CF rej. &
$\mathrm{IDR}_{\mathrm{raw}}$ & Group 95\% CI \\
\midrule
GPT-5.4 & A & 150 & 10 & 0.447 &
-- & -- & -- & -- \\
Gemini 3 Flash & A & 150 & 10 & 0.553 &
-- & -- & -- & -- \\
\midrule
GPT-5.4 & C & 100 & 18 & 0.590 &
0.280 & 0.900 & 0.230 & [0.150,0.320] \\
Gemini 3 Flash & C & 100 & 18 & 0.720 &
0.870 & 0.570 & 0.470 & [0.370,0.570] \\
\midrule
GPT-5.4 & D & 100 & 18 & 0.550 &
0.580 & 0.520 & 0.160 & [0.090,0.240] \\
Gemini 3 Flash & D & 100 & 18 & 0.590 &
0.970 & 0.210 & 0.200 & [0.120,0.280] \\
\bottomrule
\end{tabular}
\end{table}

Table~\ref{tab:appE_hssd} reports the HSSD evaluation
results. The findings reveal distinct patterns of
Full activation, CF rejection, and paired discrimination
under synthetic 3D visual evidence.
For GPT-5.4 on Task C, Full activation is 0.280
and CF rejection reaches 0.900, but raw IDR is only
0.230. Conversely, Gemini 3 Flash achieves 0.970
Full activation on reference-aided Task D, yet its
CF rejection is 0.210 and raw IDR is 0.200.
These contrasting profiles show that aggregate
accuracy and successful-trajectory acceptance do not
fully characterize paired causal discrimination.
The results extend the central SCI diagnostic
distinction to a richer synthetic 3D setting.

\paragraph{Cross-simulator interpretation.}
\label{app:3d_limitations}
MiniWorld and Habitat-Sim/HSSD use different visual
interfaces, intervention protocols, and model subsets.
Their results are therefore analyzed separately rather
than pooled into a single ranking. Both evaluations
demonstrate the diagnostic value of distinguishing
judgment correctness, presentation sensitivity, and
causal discrimination under simulated 3D evidence.

\section{Implementation Details and Reproducibility}
\label{app:implementation}

This appendix documents the implementation, data release, schema conventions,
prompting and parsing protocol, and reproducibility information for
GridWM-Judge. The experiments comprise the base MiniGrid Task~A--E
diagnostics, the Scale-2 Task~A/C/D scale extension, the MiniWorld and
Habitat Task~A/C/D 3D environment extensions, and controlled analyses.
Scale-2 expands the trajectory bank and environment coverage. MiniWorld
and Habitat evaluate judging in 3D environments. The controlled analyses
use Oracle-State evidence to control for visual perception, structured-CoT
conditions to examine response format and reasoning, and matched
presentation probes to compare evidence representations and presentation
changes. The following sections describe the code, data, configurations,
and commands used to generate, run, and score these experiments.

\subsection{Code and Data Availability}
\label{app:impl:availability}
The public code and dataset are available at the following links:

\begin{description}
    \item[Code:] \href{https://github.com/xinyuw922-pixel/Towards-Reliable-VLM-Judge}
    {GitHub repository}.
    \item[Data:] \href{https://huggingface.co/datasets/nahco318/Towards-Reliable-VLM-Judges}
    {Hugging Face dataset}.
\end{description}

The released artifacts include the generated exam files, rendered images,
ground-truth annotations, prompt templates, validation metadata, scoring scripts, and result summaries used in the paper. The code is released under the MIT License. Project-authored dataset annotations and metadata use Apache-2.0. HSSD-derived rendered images retain CC BY-NC 4.0 and the associated attribution requirements; third-party assets are not relicensed by this release. Component-specific terms are documented in the dataset card and the Habitat \texttt{LICENSES.md} file.

\subsection{Repository Structure and Compute}
\label{app:impl:code}

The repository is organized around generation, inference, scoring, and export
scripts. The top-level structure is:

\begin{verbatim}
Towards-Reliable-VLM-Judge/
  scripts/                       # shell launchers
  tools/
    minigrid/                    # trajectory and exam generation
    miniworld/                   # MiniWorld workflows
    habitat/                     # Habitat generation and validation
    exams/                       # exam and request assembly
    inference/                   # shared inference gateway
    scoring/                     # parsers, scores and aggregation
    experiments/                 # controlled diagnostic analyses
  utils/                         # shared helpers
  environments/                  # Conda environment specifications
  requirements/                  # pinned Python dependencies
  README.md
  LICENSE
\end{verbatim}

Generated trajectories, exams, responses, and scores are written to an
external run directory selected by the user, rather than bundled with the
source repository.

This work involves no model training or fine-tuning. Dataset generation and
validation are CPU-bound and use fixed seeds. VLM inference is performed either
through external APIs or local inference for open-weight models. Local model
runs use a single NVIDIA RTX 4060 Ti 16GB GPU. On our reference generation setup, generating the 1,202 base
MiniGrid Task~A--E exam items takes approximately 3.6 minutes. The 3,652
Scale-2 Task~A/C/D items take 25.8 minutes, the 214 MiniWorld Task~A/C/D
items take 7.9 minutes, and the 550 Habitat Task~A/C/D items take 82.1 minutes.
Generation time does not scale directly with the number of exam items:
multiple presentation variants can share the same underlying trajectory.

For example, the 486 base Task~C items are derived from 18 trajectory groups
with temporal, visual, and framing variations, whereas the 1,326 Scale-2
Task~C items use 442 trajectory groups with three variants per group
(\textsf{Full}, \textsf{NoCue}, and \textsf{CF}). Scale-2 therefore requires
substantially more trajectory simulation, planning, validation, and frame
rendering, rather than merely more presentations of existing trajectories.
Its construction also includes reference matching and candidate-quality
checks. The base and Scale-2 totals cover different task sets (A--E versus
A/C/D), so their item counts alone are not comparable measures of generation
workload.

The dominant runtime and monetary cost comes from VLM inference.
Table~\ref{tab:appF_dependencies} lists the main software dependencies used for
generation, inference, rendering, and scoring in the core reproduction
environment (Python 3.12.12). Dependency versions are pinned in the released
environment specifications and lock files. Habitat uses a separate Python 3.9.23 environment
with Habitat-Sim 0.3.3, NumPy 1.26.4, and Pillow 10.4.0; its environment
specification and lock file are released separately.

\begin{table}[t]
\centering
\small
\setlength{\tabcolsep}{4pt}
\renewcommand{\arraystretch}{1.08}
\caption{Main software dependencies in the core reproduction
environment; Habitat uses a separate environment.}
\label{tab:appF_dependencies}
\begin{tabular}{l l l l}
\toprule
Dependency & Version & License & Purpose \\
\midrule
MiniGrid & 3.0.0 & Apache 2.0 & Grid-world environments \\
Gymnasium & 1.2.3 & MIT & Environment interface \\
OpenAI SDK & 2.29.0 & Apache 2.0 & API inference client \\
MiniWorld & 2.1.0 & Apache 2.0 & 3D environments \\
Pygame & 2.6.1 & LGPL & Rendering support \\
Pillow & 12.1.1 & HPND & Image processing \\
\bottomrule
\end{tabular}
\end{table}

\subsection{Data Format and Metadata}
\label{app:impl:schema}

The released benchmark uses JSONL files, with one exam item per line. The
distinction between model-facing and audit-only fields is semantic: native
exam records generally use task-specific, flat fields rather than universal
\texttt{model\_input} and \texttt{audit\_only} containers. Only the designated
prompt and ordered image inputs are sent to a VLM judge in the pixel-based experiments.
Audit-only fields contain simulator states, labels, intervention metadata,
validation logs, and other information
used only for scoring and reproducibility.

The following is a conceptual illustration of that separation, not the literal
on-disk schema:
\begin{verbatim}
{
  "uid": "C.keycorridor.g000001.full.orig.clean.comp",
  "task": "C",
  "schema_version": "gridwm.exam.v1",
  "env_task": "keycorridor",
  "group_id": "g000001",
  "variant": "full",
  "presentation": {
    "temporal": "orig",
    "visual": "clean",
    "framing": "neutral",
    "rep": "composite"
  },
  "model_input": {
    "images": ["relative/path/to/image_or_frame.png"],
    "prompt": "model-facing task instruction"
  },
  "audit_only": {
    "label": 1,
    "success": true,
    "state_seq": "relative/path/to/state_trace.json",
    "actions": ["forward", "toggle", "..."],
    "reward": 1.0,
    "nocue_meta": null,
    "cf_meta": null,
    "validator_status": "passed"
  }
}
\end{verbatim}

Task-specific records implement this logical separation. Task~A items store
the action condition and multiple-choice candidate order. Task~B items store the
structured perception target, including agent direction, carried object, front
cell, and visible objects. Task~C items store trajectory variant and
presentation-probe metadata. Task~D items store matched \textsf{Full}/\textsf{CF}
group identifiers for causal-sensitivity scoring. Task~E items store the
checkpoint index and object/status label. Actual field names and example
JSONL records are provided by the released task files. For example, a native
Task~C record uses top-level fields such as \texttt{uid}, \texttt{task},
\texttt{schema\_version}, \texttt{env\_task}, \texttt{group\_id},
\texttt{variant}, \texttt{temporal}, \texttt{visual}, \texttt{image},
\texttt{prompt}, \texttt{answer}, and \texttt{label}. Here \texttt{image} and
\texttt{prompt} specify model-facing evidence; \texttt{answer} and
\texttt{label} are retained for scoring and are not appended to the prompt.
Other tasks may assemble their prompt or image list when exporting inference
requests. The Croissant reading tables are derived views in which nested
arrays and objects are encoded as reversible JSON strings; the original exam
JSONLs retain their native structure.

For Task~B, directions follow the MiniGrid convention:
\begin{center}
\small
\begin{tabular}{c c}
\toprule
Value & Direction \\
\midrule
0 & right / east \\
1 & down / south \\
2 & left / west \\
3 & up / north \\
\bottomrule
\end{tabular}
\end{center}

Door states follow the MiniGrid object-state convention: state 0 denotes open,
state 1 denotes closed, and state 2 denotes locked. The \texttt{front\_cell}
occupancy type may be \texttt{empty}, \texttt{wall}, \texttt{door},
\texttt{goal}, \texttt{ball}, \texttt{key}, \texttt{lava}, or \texttt{box}.
The \texttt{objects} list contains only visible non-background objects
(keys, balls, doors, goals, lava, and boxes), excluding the front cell itself;
walls, floors, and empty cells are not listed as objects. Agent information is
reported separately under \texttt{agent}.

\subsection{Prompting, Decoding, and Parsing}
\label{app:impl:prompting}

All models are evaluated as external judges of model-facing evidence. In the
pixel-based experiments, the prompts contain only rendered images, mission or action text when required by
the task, task instructions, and the expected answer format. They do not contain
ground-truth labels, rewards, simulator states, validation flags, or
intervention metadata. The Oracle-State control analysis adds
simulator-derived state evidence in its State and Pixel+State conditions to
control for visual perception; Pixel remains the image-only condition. These
analysis conditions do not expose the gold verdict, reward, exact simulator
mission, or audit-only validation and intervention annotations.

The evaluated model aliases in the paper are summarized in
Table~\ref{tab:appF_models}. Exact provider endpoint strings, run timestamps,
and API parameters are recorded in the released run metadata files.

\begin{table}[t]
\centering
\small
\setlength{\tabcolsep}{4pt}
\renewcommand{\arraystretch}{1.08}
\caption{Model aliases used in the paper and corresponding inference route.
Exact endpoint strings and run metadata are included in the released
machine-readable logs.}
\label{tab:appF_models}
\begin{tabular}{l l l}
\toprule
Alias in paper & Model family / provider & Inference route \\
\midrule
Claude Sonnet 4.6 & Claude Sonnet family & API \\
GPT-4o & GPT-4o family & API \\
GPT-5.4 & GPT-5-family judge run & API \\
Gemini 3 Flash & Gemini Flash family & API \\
Gemini 3.1 Pro & Gemini Pro family (Oracle analysis) & API \\
Gemini 2.5 Flash Lite & Gemini Flash Lite family & API \\
Gemini 2.0 Flash & Gemini Flash family & API \\
Kimi K2.5 & Kimi family & API \\
Qwen3-8B & Qwen open-weight family & local/API run \\
Qwen3-32B & Qwen open-weight family & local/API run \\
Qwen3-235B & Qwen large open-weight family & local/API run \\
\bottomrule
\end{tabular}
\end{table}

The following table reports the decoding settings for the base MiniGrid
direct-answer experiments; temperature is set to zero when supported by the provider:
\begin{center}
\small
\begin{tabular}{l l}
\toprule
Parameter & Value \\
\midrule
temperature & 0.0 \\
top\_p & not set unless required by provider \\
top\_k & not set unless required by provider \\
max tokens, Task A & 512 \\
max tokens, Task B & 2048 \\
max tokens, Task C/D/E & 256 \\
\bottomrule
\end{tabular}
\end{center}

The output-token limits above are not shared by all experiments or reruns.
Controlled analyses, including the four-field Oracle-State responses, and
truncation-recovery reruns may use larger output budgets.
For example, the Gemini 3.1 Pro Oracle run uses an 8192-token limit.
Experiment-specific run metadata governs these exceptions; a response that
reaches its output limit may be truncated and is scored according to the
actual recoverability of its answer.

Invalid, empty, refused, or unparsable responses are
logged separately. In conservative all-group analyses,
such outputs are counted as incorrect. The main
Task-C paired metrics retain the released valid-pair
eligibility rules, whereas dedicated Task-D and
Scale-2 analyses use their specified all-group
protocols (Appendix~\ref{app:definitions}). The parser is  task-specific:

\paragraph{Task A.}
The parser extracts the first isolated multiple-choice label in
\texttt{\{A,B,C,D\}}. If no valid label is found, the response is marked invalid.

\paragraph{Tasks C and D.}
Binary success/failure judgments are parsed with case-insensitive keyword
matching. Success words include \texttt{success}, \texttt{succeed},
\texttt{succeeded}, \texttt{successful}, and \texttt{yes}. Failure words include
\texttt{fail}, \texttt{failed}, \texttt{failure}, and \texttt{no}. If both
classes appear, the parser applies the deterministic conflict rule implemented
in the released scoring script; otherwise invalid or ambiguous outputs are
marked incorrect.

\paragraph{Tasks B and E.}
The scorers distinguish strict, recoverable, and failed parses rather than
rejecting all formatting deviations. Task~B first extracts a JSON object and
attempts standard JSON parsing, with limited recovery for surrounding text,
code fences, and trailing commas. Task~B requires the expected
structured fields for the agent, front cell, and visible objects. Task~E expects
a JSON object with an \texttt{answer} field and also attempts recovery from
code fences or surrounding text. Unambiguously recoverable answers contribute
to the scores, with parse modes logged separately. Exact-match and component-level
scores are both logged when applicable.

\subsection{Reproduction Commands and Reporting}
\label{app:impl:reproduction}

\begin{table}[htbp!]
\centering
\small
\setlength{\tabcolsep}{4pt}
\renewcommand{\arraystretch}{1.08}
\caption{Reproducibility summary.}
\label{tab:appF_repro_summary}
\begin{tabular}{l l}
\toprule
Item & Documentation / release location \\
\midrule
Code & public GitHub repository \\
Dataset & public Hugging Face dataset \\
Exam files & JSONL files in dataset release \\
Rendered images & image folders referenced by JSONL records \\
Prompt templates & repository and dataset metadata \\
Ground-truth labels & audit-only fields in JSONL records \\
Model outputs & prediction logs and score files \\
Scoring scripts & \texttt{tools/scoring/}, via \texttt{scripts/score.sh} \\
Random seeds & workflow-specific; see commands and run metadata \\
Training or fine-tuning & none \\
Dataset license & Apache-2.0; HSSD images: CC BY-NC 4.0 \\
Code license & MIT License \\
Privacy & synthetic simulator data only; no human subjects or scraped images \\
\bottomrule
\end{tabular}
\end{table}

Table~\ref{tab:appF_repro_summary} summarizes the released artifacts and
documentation needed to reproduce the benchmark results. A typical reproduction workflow is:

\begin{verbatim}
# 1. Install dependencies
conda env create -f environments/core.yml
conda activate gridwm-repro-core
python -m pip install -r requirements/core.lock.txt
export RUN_ROOT="$HOME/gridwm-runs"

# 2. Generate validated trajectories and exam items
bash scripts/generate_minigrid.sh --profile formal \
  --output "$RUN_ROOT/base"

# 3. Run model inference
export BUNDLE="$RUN_ROOT/base/minigrid/bundle"
export REQUESTS="$BUNDLE/_requests_from_manifest"
bash scripts/infer.sh \
  --requests "$REQUESTS/requests_ABCDE_full.jsonl" \
  --protocol chat --base-url "$VLM_BASE_URL" --model "$MODEL" \
  --max-tokens 8192 --temperature 0 --workers 2 --rps 1 \
  --output "$RUN_ROOT/base-responses"

# 4. Score predictions and export metrics
bash scripts/score.sh --protocol minigrid \
  --exam-dir "$BUNDLE/exam" \
  --responses "$RUN_ROOT/base-responses/responses.jsonl" \
  --ci --output "$RUN_ROOT/base-scores"
\end{verbatim}

Set \texttt{VLM\_BASE\_URL} and \texttt{MODEL}, and configure the provider
credentials as documented in the repository README, before inference. The
inference command illustrates the shared launcher with an 8192-token
budget. For the base MiniGrid direct-answer settings, run task-specific
request files with the corresponding limits listed above.
The repository README provides wrappers for the other
environments and analysis protocols.

Seeds are workflow-specific. The generation
launcher defaults to 830000; base Task~A exam sampling uses 42. Habitat
navigation generation uses base seed 20260725, while its action-candidate
construction and exam sampling use 20260726. The structured-CoT and
Oracle-State scoring protocols use bootstrap seeds 20260724 and 20260725,
respectively; the general confidence-interval workflow defaults to
20261103. Seeds and overrides should be reported with the corresponding
generation or scoring command.

The released scoring scripts include the deterministic metric aggregation
and uncertainty-estimation procedures used to produce the tables and figures
in the paper.

\section{Scope and Impact}
\label{app:scope_impact}
This appendix discusses the scope, limitations,
and broader impacts of GridWM-Judge. We summarize
the validity boundaries of the controlled diagnostics,
consider responsible benchmark use and release,
and outline directions for extending SCI evaluation
to more diverse models and environments.

\subsection{Scope and Limitations}
\label{app:scope_limitations}
Table~\ref{tab:scope_limitations} summarizes the scope of claims supported by our work and the corresponding limitations.

\begin{table}[b]
    \centering
    \small
    \setlength{\tabcolsep}{4pt}
    \renewcommand{\arraystretch}{1.08}
    \caption{Scope of claims and corresponding limitations.}
    \label{tab:scope_limitations}
    \begin{tabular}{p{0.22\linewidth}p{0.34\linewidth}p{0.34\linewidth}}
        \toprule
        Axis & What the paper supports & What remains outside scope \\
        \midrule
        Domain
        & Controlled MiniGrid diagnostics with auditable states, actions, and success predicates.
        & Direct deployment claims for real robots, photorealistic video, or open-ended embodied tasks. \\
        Agent behavior
        & Deterministic rollouts that isolate judge-side reliability failures.
        & Noisy control, stochastic policies, suboptimal exploration, and long-horizon recovery behavior. \\
        Presentation
        & Temporal, visual, prompt-framing, and representation-format probes over matched trajectory semantics.
        & All possible UI formats, natural-language task framings, or video encodings. \\
        Models
        & A finite set of model-based judges under documented task-specific evaluation protocols.
        & Universal claims about all VLMs, future models, or heavily optimized prompting/training regimes. \\
        Metrics
        & Diagnostic evidence for correctness, presentation stability, representation exchangeability, and causal sensitivity.
        & A single deployment-readiness score or safety certificate. \\
        \bottomrule
    \end{tabular}
\end{table}

\paragraph{Controlled diagnostic scope.}
GridWM-Judge is designed as a controlled diagnostic benchmark rather than a
realistic robotics benchmark. The Tier-0 evaluation suite uses MiniGrid-style
symbolic environments because they provide auditable states, actions, object
identities, and success predicates. This control is essential for constructing
aligned \textsf{Full}, \textsf{NoCue}, and \textsf{CF} variants, and for
separating presentation changes from causal interventions. The corresponding
limitation is that MiniGrid abstracts away continuous perception, 3D geometry,
photorealistic appearance, multi-agent interaction, and real-world sensor noise.
Our results therefore diagnose reliability failures under controlled embodied
trajectories; they should not be interpreted as direct estimates of deployment
performance in real robotic systems.

\paragraph{Environment and agent-behavior scope.}
The current benchmark emphasizes short-horizon, discrete-action navigation and
interaction tasks with deterministic simulator dynamics. The trajectories are
produced by deterministic planners so that the underlying success/failure labels
and causal events can be audited exactly. This design does not cover noisy
control, partial actuation failure, stochastic policies, exploration behavior, or
suboptimal agent trajectories. Future versions should include more diverse
behavior policies and longer-horizon tasks, but the present paper intentionally
uses deterministic rollouts to isolate judge-side failures rather than
agent-side failures.

\paragraph{Model and prompting scope.}
The evaluated models constitute a finite set of
model-based judges under the task-specific prompting
and decoding protocols described in
Appendix~\ref{app:implementation}.
The benchmark
is not a comprehensive leaderboard over all proprietary and open VLMs. The
reported failures should be read as evidence that current representative judges
can violate state-conditional invariance, not as a universal impossibility
result. 
Likewise, we use standard judge prompts rather than systematically
optimizing chain-of-thought prompting, few-shot demonstrations, tool use,
self-consistency voting, or specialized fine-tuning. 
Although we include controlled prompt-structure
ablations, we do not exhaustively optimize few-shot
demonstrations, tool use, self-consistency voting,
or specialized fine-tuning.
Better prompting or
training may improve some scores, but such improvements would need to be tested
under the same causal and presentation controls.

\paragraph{Task and SCI-condition scope.}
GridWM-Judge operationalizes four diagnostic conditions: presentation
invariance, action-conditioned dynamics, representation exchangeability, and
causal sensitivity. Task~C primarily probes presentation invariance through
temporal, visual, and prompt-framing perturbations; Task~A probes
action-conditioned transition use; the composite--split comparison probes
representation exchangeability; and Task~D probes causal sensitivity through
matched \textsf{Full}/\textsf{CF} comparisons. Tasks~B and~E are local
perception and state-tracking controls that help interpret higher-level
failures. This coverage is diagnostic rather than exhaustive: a judge that
performs well on these tasks may still fail under different environments, longer
contexts, ambiguous goals, or naturalistic video evidence.

\paragraph{\textsf{NoCue} interpretation.}
\textsf{NoCue} removes selected early visual evidence while preserving the
underlying successful trajectory. It is therefore an evidence-removal diagnostic,
not a semantics-preserving presentation transformation in the same sense as
temporal order, visual style, prompt framing, or representation format. We use
\textsf{NoCue} to ask whether a judge relies on early causal cues, but we do not
treat \textsf{NoCue} stability alone as proof of presentation invariance.

\paragraph{3D extensions.}
The MiniWorld and Habitat experiments in Appendix~\ref{app:3d_extensions} provide supplementary 3D diagnostic evaluations. They should be interpreted separately from the Tier-0
MiniGrid results because the 3D environments introduce different perceptual and
rendering factors. We do not aggregate MiniWorld, Habitat and MiniGrid results into a
single headline score.

\paragraph{Threats to validity.}
We summarize three main validity threats. First, \textit{internal validity} may
be affected by parser ambiguity, API nondeterminism, or generator artifacts. We
mitigate these risks by using constrained answer formats, deterministic parsing
rules, fixed seeds where applicable, and validation gates over generated
trajectories. Second, \textit{external validity} is limited by the synthetic grid-world domain and by the finite set of evaluated models. 
The MiniWorld and Habitat extensions partially
probe this issue but do not eliminate it.
Third,
\textit{construct validity} depends on whether SCI captures the aspects of judge
reliability that matter in downstream use. We view SCI as a necessary diagnostic
principle: a reliable judge should not change its verdict merely because
evidence is reformatted, and should change when a causal intervention changes
the outcome. However, SCI is not sufficient for deployment safety, because real
settings may involve ambiguous goals, incomplete observations, distribution
shift, and human factors not represented here.

\subsection{Societal Impact and Outlook}
\label{app:societal_impact}
Table~\ref{tab:impact_mitigations} summarizes the main impact considerations
and recommended mitigations for responsible release and use.

\paragraph{Positive impacts.}
VLM judges are increasingly used to evaluate multimodal and embodied systems. A
diagnostic benchmark such as GridWM-Judge can help researchers identify when a
judge is correct for the wrong reason, unstable under benign presentation
changes, or insensitive to causal outcome changes. This may reduce overreliance
on opaque aggregate accuracy and encourage more careful reporting of judge
reliability, especially in evaluation pipelines that influence model selection,
dataset curation, or simulated safety testing.

\paragraph{Risks of overclaiming and misuse.}
The main risk is not that the benchmark directly enables harmful physical
actions, since it uses synthetic grid-world data and releases no personal,
private, or safety-critical operational data. The more realistic risk is
\textit{overclaiming}: users may treat high GridWM-Judge scores as a guarantee
that a VLM judge is safe for real-world robotics or high-stakes decision-making.
This would be inappropriate. The benchmark should be used as one diagnostic
layer, not as a deployment certificate. Another risk is benchmark gaming: once
examples and prompts are public, models or systems may be tuned to exploit the
benchmark without genuinely improving causal grounding. Future releases should
therefore include held-out diagnostic slices and versioned evaluation protocols.

\paragraph{Responsible release.}
The dataset is synthetic and simulator-generated. It does not contain
human-subject data, personal information, scraped images, or private
trajectories. The release should nevertheless include documentation of generation
scripts, prompt templates, schemas, validation logs, licenses for dependencies,
and known limitations. We recommend that any public release state that
the benchmark evaluates VLM judges under controlled MiniGrid, MiniWorld, and Habitat
environments, and that it is not intended as a standalone safety evaluation for
deployed embodied systems.

\paragraph{Environmental considerations.}
MiniGrid trajectory generation and validation are lightweight relative to
large-scale VLM inference. The main compute footprint comes from repeated model
calls over multiple tasks, variants, and presentation probes. The benchmark is
intentionally modular so that researchers can run smaller diagnostic subsets
when compute is limited. Reporting compute resources and inference settings in
Appendix~\ref{app:implementation} is important for reproducibility and for
understanding the environmental cost of evaluation.

\begin{table}[t]
    \centering
    \small
    \setlength{\tabcolsep}{4pt}
    \renewcommand{\arraystretch}{1.08}
    \caption{Impact considerations and recommended mitigations.}
    \label{tab:impact_mitigations}
    \begin{tabular}{p{0.26\linewidth}p{0.32\linewidth}p{0.32\linewidth}}
        \toprule
        Consideration & Potential concern & Mitigation in this paper / release \\
        \midrule
        Overclaiming
        & Treating SCI scores as deployment-readiness guarantees.
        & Explicitly state the controlled diagnostic scope and report limitations alongside results. \\
        Benchmark gaming
        & Tuning models to public prompts or examples without improving real reliability.
        & Document prompts and probes; plan versioned splits and held-out diagnostic slices for future releases. \\
        Misuse of judge diagnostics
        & Crafting inputs that exploit known judge weaknesses.
        & Frame results as reliability audits and avoid presenting the benchmark as an adversarial attack toolkit. \\
        Privacy and data rights
        & Releasing sensitive images or trajectories.
        & Use simulator-generated synthetic data with no human subjects or private data. \\
        Compute cost
        & Repeated VLM inference across many variants.
        & Provide modular subsets, deterministic scripts, and transparent compute reporting. \\
        \bottomrule
    \end{tabular}
\end{table}

\paragraph{Future directions.}
The most important next step is to test whether the same SCI failures persist in
richer domains, such as longer MiniWorld tasks, photorealistic simulators, real
robot videos, and embodied tasks with ambiguous or compositional goals. A second
direction is to expand the model suite, including open-weight VLMs,
video-language models, and specialized judge models. A third direction is to
evaluate mitigation strategies such as structured output constraints, calibrated
abstention, evidence localization, multi-view inputs, self-consistency, or
training on counterfactual diagnostic data. Future benchmark versions should
also include adversarially selected but simulator-valid trajectories to test
whether apparent judge reliability survives more challenging causal and
presentation shifts.

\paragraph{Conclusion on scope.}
GridWM-Judge provides reproducible diagnostics
for examining whether model-based judges respond
appropriately to presentation changes and causal
interventions. Its findings are bounded by the
evaluated domains, models, and protocols, and
should not be interpreted as deployment or
safety guarantees.

\end{document}